\PassOptionsToPackage{unicode}{hyperref}
\PassOptionsToPackage{hyphens}{url}
\PassOptionsToPackage{dvipsnames,svgnames,x11names}{xcolor}
\documentclass[
  onecolumn]{article}
\usepackage{xcolor}
\usepackage{amsmath,amssymb}
\usepackage{iftex}
\ifPDFTeX
  \usepackage[T1]{fontenc}
  \usepackage[utf8]{inputenc}
  \usepackage{textcomp} 
\else 
  \usepackage{unicode-math} 
  \defaultfontfeatures{Scale=MatchLowercase}
  \defaultfontfeatures[\rmfamily]{Ligatures=TeX,Scale=1}
\fi
\usepackage{lmodern}
\ifPDFTeX\else
\fi
\IfFileExists{upquote.sty}{\usepackage{upquote}}{}
\IfFileExists{microtype.sty}{
  \usepackage[]{microtype}
  \UseMicrotypeSet[protrusion]{basicmath} 
}{}
\makeatletter
\@ifundefined{KOMAClassName}{
  \IfFileExists{parskip.sty}{%
    \usepackage{parskip}
  }{
    \setlength{\parindent}{0pt}
    \setlength{\parskip}{6pt plus 2pt minus 1pt}}
}{
  \KOMAoptions{parskip=half}}
\makeatother
\makeatletter
\ifx\paragraph\undefined\else
  \let\oldparagraph\paragraph
  \renewcommand{\paragraph}{
    \@ifstar
      \xxxParagraphStar
      \xxxParagraphNoStar
  }
  \newcommand{\xxxParagraphStar}[1]{\oldparagraph*{#1}\mbox{}}
  \newcommand{\xxxParagraphNoStar}[1]{\oldparagraph{#1}\mbox{}}
\fi
\ifx\subparagraph\undefined\else
  \let\oldsubparagraph\subparagraph
  \renewcommand{\subparagraph}{
    \@ifstar
      \xxxSubParagraphStar
      \xxxSubParagraphNoStar
  }
  \newcommand{\xxxSubParagraphStar}[1]{\oldsubparagraph*{#1}\mbox{}}
  \newcommand{\xxxSubParagraphNoStar}[1]{\oldsubparagraph{#1}\mbox{}}
\fi
\makeatother

\usepackage{color}
\usepackage{fancyvrb}

\DefineVerbatimEnvironment{Highlighting}{Verbatim}{commandchars=\\\{\}}
\usepackage{framed}
\definecolor{shadecolor}{RGB}{241,243,245}
\newenvironment{Shaded}{\begin{snugshade}}{\end{snugshade}}

\newcommand{\BuiltInTok}[1]{\textcolor[rgb]{0.00,0.23,0.31}{#1}}

\newcommand{\CommentTok}[1]{\textcolor[rgb]{0.37,0.37,0.37}{#1}}

\newcommand{\DecValTok}[1]{\textcolor[rgb]{0.68,0.00,0.00}{#1}}

\newcommand{\FloatTok}[1]{\textcolor[rgb]{0.68,0.00,0.00}{#1}}

\newcommand{\ImportTok}[1]{\textcolor[rgb]{0.00,0.46,0.62}{#1}}

\newcommand{\NormalTok}[1]{\textcolor[rgb]{0.00,0.23,0.31}{#1}}
\newcommand{\OperatorTok}[1]{\textcolor[rgb]{0.37,0.37,0.37}{#1}}

\newcommand{\SpecialCharTok}[1]{\textcolor[rgb]{0.37,0.37,0.37}{#1}}
\newcommand{\SpecialStringTok}[1]{\textcolor[rgb]{0.13,0.47,0.30}{#1}}
\newcommand{\StringTok}[1]{\textcolor[rgb]{0.13,0.47,0.30}{#1}}

\usepackage{longtable,booktabs,array}
\usepackage{calc} 
\usepackage{etoolbox}
\makeatletter
\patchcmd\longtable{\par}{\if@noskipsec\mbox{}\fi\par}{}{}
\makeatother
\IfFileExists{footnotehyper.sty}{\usepackage{footnotehyper}}{\usepackage{footnote}}
\makesavenoteenv{longtable}
\usepackage{graphicx}
\makeatletter
\newsavebox\pandoc@box
\newcommand*\pandocbounded[1]{
  \sbox\pandoc@box{#1}%
  \Gscale@div\@tempa{\textheight}{\dimexpr\ht\pandoc@box+\dp\pandoc@box\relax}%
  \Gscale@div\@tempb{\linewidth}{\wd\pandoc@box}%
  \ifdim\@tempb\p@<\@tempa\p@\let\@tempa\@tempb\fi
  \ifdim\@tempa\p@<\p@\scalebox{\@tempa}{\usebox\pandoc@box}%
  \else\usebox{\pandoc@box}%
  \fi%
}
\def\fps@figure{htbp}
\makeatother

\NewDocumentCommand\citeproctext{}{}

\makeatletter
 \let\@cite@ofmt\@firstofone
 \def\@biblabel#1{}
 \def\@cite#1#2{{#1\if@tempswa , #2\fi}}
\makeatother
\newlength{\cslhangindent}
\newlength{\csllabelwidth}
\newenvironment{CSLReferences}[2] 
 {\begin{list}{}{%
  \setlength{\itemindent}{0pt}
  \setlength{\leftmargin}{0pt}
  \setlength{\parsep}{0pt}
  \ifodd #1
   \setlength{\leftmargin}{\cslhangindent}
   \setlength{\itemindent}{-1\cslhangindent}
  \fi
  \setlength{\itemsep}{#2\baselineskip}}}
 {\end{list}}
\usepackage{calc}

\providecommand{\tightlist}{%
  \setlength{\itemsep}{0pt}\setlength{\parskip}{0pt}}

\usepackage{arxiv}
\usepackage{orcidlink}
\usepackage{amsmath}
\makeatletter
\@ifpackageloaded{tcolorbox}{}{\usepackage[skins,breakable]{tcolorbox}}
\@ifpackageloaded{fontawesome5}{}{\usepackage{fontawesome5}}
\definecolor{quarto-callout-color}{HTML}{909090}
\definecolor{quarto-callout-note-color}{HTML}{0758E5}
\definecolor{quarto-callout-important-color}{HTML}{CC1914}
\definecolor{quarto-callout-warning-color}{HTML}{EB9113}
\definecolor{quarto-callout-tip-color}{HTML}{00A047}
\definecolor{quarto-callout-caution-color}{HTML}{FC5300}
\definecolor{quarto-callout-color-frame}{HTML}{acacac}
\definecolor{quarto-callout-note-color-frame}{HTML}{4582ec}
\definecolor{quarto-callout-important-color-frame}{HTML}{d9534f}
\definecolor{quarto-callout-warning-color-frame}{HTML}{f0ad4e}
\definecolor{quarto-callout-tip-color-frame}{HTML}{02b875}
\definecolor{quarto-callout-caution-color-frame}{HTML}{fd7e14}
\makeatother
\makeatletter
\@ifpackageloaded{caption}{}{\usepackage{caption}}
\AtBeginDocument{%
\ifdefined\contentsname
  \renewcommand*\contentsname{Table of contents}
\else
  \newcommand\contentsname{Table of contents}
\fi
\ifdefined\listfigurename
  \renewcommand*\listfigurename{List of Figures}
\else
  \newcommand\listfigurename{List of Figures}
\fi
\ifdefined\listtablename
  \renewcommand*\listtablename{List of Tables}
\else
  \newcommand\listtablename{List of Tables}
\fi
\ifdefined\figurename
  \renewcommand*\figurename{Figure}
\else
  \newcommand\figurename{Figure}
\fi
\ifdefined\tablename
  \renewcommand*\tablename{Table}
\else
  \newcommand\tablename{Table}
\fi
}
\@ifpackageloaded{float}{}{\usepackage{float}}
\floatstyle{ruled}
\@ifundefined{c@chapter}{\newfloat{codelisting}{h}{lop}}{\newfloat{codelisting}{h}{lop}[chapter]}
\floatname{codelisting}{Listing}

\usepackage{amsthm}
\theoremstyle{definition}
\newtheorem{definition}{Definition}[section]
\theoremstyle{remark}
\AtBeginDocument{}

\makeatother
\makeatletter
\@ifpackageloaded{caption}{}{\usepackage{caption}}
\@ifpackageloaded{subcaption}{}{\usepackage{subcaption}}
\makeatother
\usepackage{bookmark}
\IfFileExists{xurl.sty}{\usepackage{xurl}}{} 
\makeatletter
\@ifundefined{xmpquote}{\newcommand{\xmpquote}[1]{#1}}{}
\makeatother
\hypersetup{
  pdftitle={Short-term load forecasting under EU-AI Act Requirements in Safety-Critical Environments},
  pdfauthor={Thomas Bartz-Beielstein; Inalbek Akiev; Soheil Ayati; Erik Birkholz; Henning Fritz Delius; Shervin Faramarzi Babadi; Fabian Alexander Feith; Mayank Dinesh Mehta; Lalo Mohamad; Marc Steffgen},
  pdfkeywords={\xmpquote{load
forecasting}, \xmpquote{LightGBM}, \xmpquote{gradient
boosting}, \xmpquote{hyperparameter
optimisation}, \xmpquote{ENTSO-E}, \xmpquote{safety-critical machine
learning}, \xmpquote{AI Act (AIA)}, \xmpquote{green AI}},
  colorlinks=true,
  linkcolor={blue},
  filecolor={Maroon},
  citecolor={Blue},
  urlcolor={Blue},
  pdfcreator={LaTeX via pandoc}}

\renewcommand{\today}{2026-11-27}
\newcommand{\runninghead}{A Preprint }
\renewcommand{\runninghead}{Working Paper }
\title{Short-term load forecasting under EU-AI Act Requirements in
Safety-Critical Environments}
\usepackage{etoolbox}
\makeatletter
\providecommand{\subtitle}[1]{
  \apptocmd{\@title}{\par {\large #1 \par}}{}{}
}
\makeatother
\subtitle{Results from a 41-day live challenge on the aggregated German
transmission-grid load}
\def\asep{\\\\\\ } 
\def\asep{\And }
\author{\textbf{Thomas
Bartz-Beielstein}~\orcidlink{0000-0002-5938-5158}\\\\THK-AI Research
Cluster,
Germany\\\\\href{mailto:thomas.bartz-beielstein@th-koeln.de}{thomas.bartz-beielstein@th-koeln.de}\asep\textbf{Inalbek
Akiev}\\\\Team Fabinalii. TH Köln,
Germany\\\\\href{mailto:inalbek.akiev@smail.th-koeln.de}{inalbek.akiev@smail.th-koeln.de}\asep\textbf{Soheil
Ayati}~\orcidlink{0009-0006-6820-7229}\\\\Team Hot Rod. TH Köln,
Germany\\\\\href{mailto:soheil.ayati@smail.th-koeln.de}{soheil.ayati@smail.th-koeln.de}\asep\textbf{Erik
Birkholz}\\\\Team Hot Rod. TH Köln,
Germany\\\\\href{mailto:erik.birkholz@smail.th-koeln.de}{erik.birkholz@smail.th-koeln.de}\asep\textbf{Henning
Fritz Delius}\\\\Team Hot Rod. TH Köln,
Germany\\\\\href{mailto:henning\_fritz.delius@smail.th-koeln.de}{henning\_fritz.delius@smail.th-koeln.de}\asep\textbf{Shervin
Faramarzi Babadi}~\orcidlink{0009-0009-2982-8964}\\\\Team Neura. TH
Köln,
Germany\\\\\href{mailto:shervin.faramarzi\_babadi@smail.th-koeln.de}{shervin.faramarzi\_babadi@smail.th-koeln.de}\asep\textbf{Fabian
Alexander Feith}\\\\Team Fabinalii. TH Köln,
Germany\\\\\href{mailto:fabian\_alexander.feith@smail.th-koeln.de}{fabian\_alexander.feith@smail.th-koeln.de}\asep\textbf{Mayank
Dinesh Mehta}\\\\Team Neura. TH Köln,
Germany\\\\\href{mailto:mayank\_dinesh.mehta@smail.th-koeln.de}{mayank\_dinesh.mehta@smail.th-koeln.de}\asep\textbf{Lalo
Mohamad}~\orcidlink{0009-0007-4464-0172}\\\\Team Weather Report. TH
Köln,
Germany\\\\\href{mailto:lalomohamad0212@gmail.com}{lalomohamad0212@gmail.com}\asep\textbf{Marc
Steffgen}\\\\Team Hot Rod. TH Köln,
Germany\\\\\href{mailto:marc.steffgen@smail.th-koeln.de}{marc.steffgen@smail.th-koeln.de}}
\date{2026-11-27}
\begin{document}
\maketitle
\begin{abstract}
Short-term load forecasting (STLF) is a core task in power-system
operations. Because power systems belong to the critical infrastructure,
forecasting in this setting is no longer only an accuracy problem.
Determinism, fail-safe handling, a minimal attack surface, and the
absence of dead code are software-engineering requirements. This report
presents results from a 41-day live challenge that evaluated an STLF
pipeline for the aggregated German transmission-grid load under the
requirements of the European Union's AI Act (AIA). The pipeline predicts
the 24 hourly load values of a target day from European Network of
Transmission System Operators for Electricity (ENTSO-E) data. It is
based on the open-source Python library \texttt{spotforecast2-safe},
which is designed for safety-critical forecasting environments, and
combines gap- and anomaly-aware preprocessing with calendar and weather
covariates and a forecasting algorithm. Forecast quality is compared
with the official ENTSO-E day-ahead forecast, which the
\texttt{spotforecast2-safe} pipeline outperforms. Transparent,
deterministic, low-cost, and auditable local models, referred to in this
report as BABL2L, remain competitive with energy-intensive pretrained
foundation models with more than 100 million parameters, such as
Chronos-2. The challenge implementation, the full submission history,
and the archived leaderboard are publicly available.
\end{abstract}
{\bfseries \emph Keywords}
\def\sep{\textbullet\ }
load forecasting \sep LightGBM \sep gradient
boosting \sep hyperparameter
optimisation \sep ENTSO-E \sep safety-critical machine learning \sep AI
Act (AIA) \sep 
green AI

\section{Introduction}\label{sec-intro}

Short-term load forecasting (STLF) is a core task in electricity-grid
management (Ullah et al. 2024). Forecast errors translate directly into
operational cost. This paper focuses on one concrete STLF problem posed
as a live forecasting challenge (Bartz-Beielstein 2026a) during the
summer term of 2026 at TH Köln. Several load-forecasting scenarios are
analysed in research and practice, see Table~\ref{tbl-hong16a}. During
the challenge, the task was to predict the 24 hourly values of the next
day for the aggregated German (DE) market-zone total load, measured in
megawatts. Predictions are based on historical data from ENTSO-E, the
European Network of Transmission System Operators for Electricity
(ENTSO-E 2024). Ground truth is the ENTSO-E actual load for the target
day. Submissions are ranked by mean absolute error (MAE) in megawatts,
averaged across all scored days, so lower values indicate better
forecasts. The official ENTSO-E day-ahead forecast, called the ``ENTSO-E
baseline'' in the remainder of this paper, competes in the leaderboard
alongside two seasonal-naive forecasts and current foundation models.
Because Möbius et al. (2025) report systematic under-prediction for the
German--Luxembourg bidding zone, the ENTSO-E baseline is probably
biased. They argue that the remaining errors retain enough
autoregressive structure for an error model alone to remove roughly one
fifth of their magnitude. The challenge rules therefore prohibit use of
the ENTSO-E baseline as a covariate unless the entry is explicitly
labelled as ENTSO-E-assisted through an \texttt{entsoe} suffix in its
name.

\begin{longtable}[]{@{}
  >{\raggedright\arraybackslash}p{(\linewidth - 8\tabcolsep) * \real{0.0723}}
  >{\raggedright\arraybackslash}p{(\linewidth - 8\tabcolsep) * \real{0.2530}}
  >{\raggedright\arraybackslash}p{(\linewidth - 8\tabcolsep) * \real{0.2410}}
  >{\raggedright\arraybackslash}p{(\linewidth - 8\tabcolsep) * \real{0.2169}}
  >{\raggedright\arraybackslash}p{(\linewidth - 8\tabcolsep) * \real{0.2169}}@{}}
\caption{Key features of different load forecasting problems (Hong and
Fan 2016). Long-term load forecasting (LTLF), medium term load
forecasting (MTLF), short-term load forecasting (STLF), very short term
load forecasting (VSTLF), spatial load forecasting (SLF), hierarchical
load forecasting (HLF), and probabilistic load forecasting
(PLF).}\label{tbl-hong16a}\tabularnewline
\toprule\noalign{}
\begin{minipage}[b]{\linewidth}\raggedright
\end{minipage} & \begin{minipage}[b]{\linewidth}\raggedright
Temporal resolution
\end{minipage} & \begin{minipage}[b]{\linewidth}\raggedright
Spatial resolution
\end{minipage} & \begin{minipage}[b]{\linewidth}\raggedright
Forecast horizon
\end{minipage} & \begin{minipage}[b]{\linewidth}\raggedright
Output format
\end{minipage} \\
\midrule\noalign{}
\endfirsthead
\toprule\noalign{}
\begin{minipage}[b]{\linewidth}\raggedright
\end{minipage} & \begin{minipage}[b]{\linewidth}\raggedright
Temporal resolution
\end{minipage} & \begin{minipage}[b]{\linewidth}\raggedright
Spatial resolution
\end{minipage} & \begin{minipage}[b]{\linewidth}\raggedright
Forecast horizon
\end{minipage} & \begin{minipage}[b]{\linewidth}\raggedright
Output format
\end{minipage} \\
\midrule\noalign{}
\endhead
\bottomrule\noalign{}
\endlastfoot
LTLF & Monthly/annual & N/A & Years & Point \\
MTLF & Days / Weeks & N/A & Weeks to months & Point \\
STLF & Hourly & N/A & Days & Point \\
VSTLF & Sub-hourly & N/A & Hours to days & Point \\
SLF & Monthly/annual & Small area & Years & Point \\
HLF & Hourly & Premise & Hours to years & Point \\
PLF & Hourly & N/A & Hours to years & Density/interval \\
\end{longtable}

The STLF pipeline, which the organiser made available to the student
teams as a starting point, treats the 24-hour prediction horizon as a
recursive multi-step forecasting problem. A Light Gradient-Boosting
Machine (LightGBM) regressor (Ke et al. 2017) is wrapped in a recursive
forecaster that predicts one hour at a time, with each prediction fed
back as input for the next step. During data preparation, the original
ENTSO-E series passes through anomaly-aware preprocessing. Outliers and
implausible values can be flagged, for example with an Isolation Forest
(Liu et al. 2008) or by manually defined bounds, and gaps are marked so
that corrupted history is not passed to the prediction model. The
recursive forecaster uses lagged load values together with calendar
covariates such as hour of day, day of week, and public holidays.
Hyperparameters are tuned with surrogate-model optimisation using
\texttt{spotoptim} (Bartz-Beielstein 2026b) and \texttt{Optuna} (Akiba
et al. 2019). The recursive STLF pipeline itself is provided by the
safety-critical package \texttt{spotforecast2-safe} (Bartz-Beielstein
and Bartz 2026), which vendors a reviewed subset of the
\texttt{skforecast} recursive algorithm (Amat Rodrigo and Escobar Ortiz
2024). Its \texttt{spotoptim}-tuned configuration is listed on the
challenge leaderboard as \texttt{spotoptim\ lgbm} and is referred to in
the text as the \texttt{spotoptim-lgbm} forecaster.

The European Union's Artificial Intelligence Act (AIA) (European
Parliament and Council of the European Union 2024) classifies an
artificial intelligence (AI) system used as a \emph{safety component} in
the supply of electricity as high-risk. In this context, determinism,
fail-safe handling, a minimal attack surface, and the absence of dead
code are crucial features. They become operationally relevant system
properties, especially when AI-based forecasts are used in
\emph{automated} decisions for critical infrastructure. STLF is
therefore no longer simply a prediction-accuracy problem. It is also a
software-engineering and compliance problem. This report considers the
implications for STLF methods that follow from the relevant legal
framework: the AIA (European Parliament and Council of the European
Union 2024), the Cyber Resilience Act (CRA) (European Parliament and
Council 2024b), the NIS-2 Directive (European Parliament and Council of
the European Union 2022a), which is transposed into the German
BSI-Gesetz (Deutscher Bundestag 2025), the Critical Entities Resilience
(CER) Directive (European Parliament and Council of the European Union
2022b), which is transposed into the German KRITIS-Dachgesetz
(Bundesministerium des Innern 2026), and the Product Liability Directive
(European Parliament and Council 2024a), which is transposed into the
German ProdHaftG-Novelle (Deutscher Bundestag 2026).

\begin{figure}[H]

\centering{

\pandocbounded{\includegraphics[keepaspectratio]{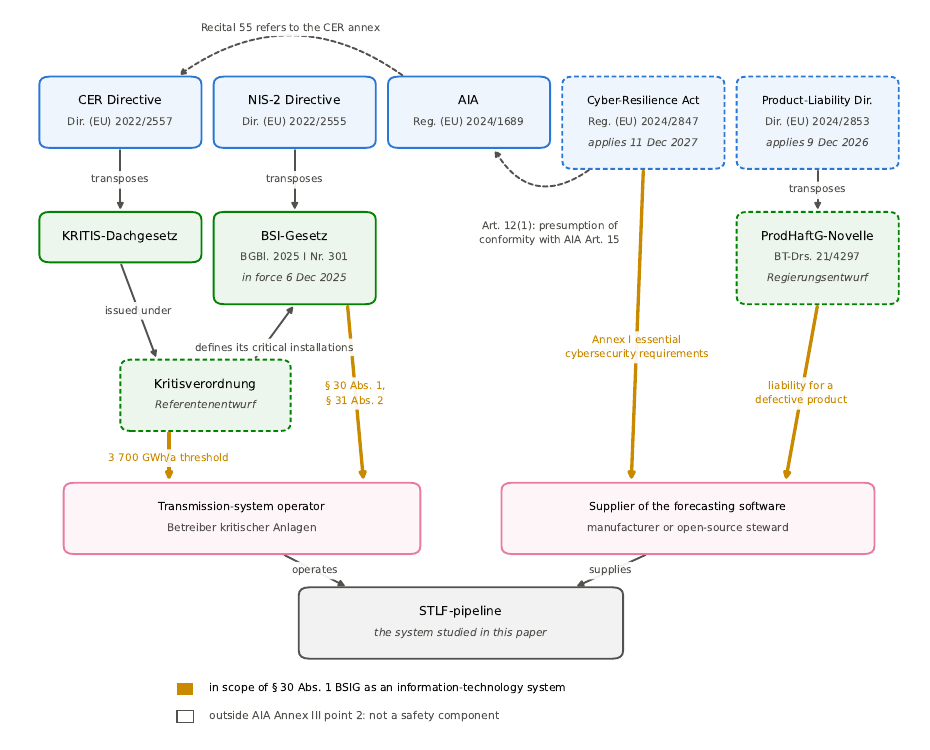}}

}

\caption{\label{fig-regulatory-map}Simplified summary of regulatory
instruments that are relevant for the STLF-forecasting pipeline.
\emph{Blue} marks Union instruments, \emph{green} their German
counterparts, and \emph{pink} the two parties. Solid arrows mark
derivation, either a directive transposed into German law or an
ordinance issued under a statute. The AIA and the Cyber Resilience Act
have no German counterpart because they bind directly. \emph{Dashed}
arrows mark a reference from one instrument to another. \emph{Ochre}
arrows mark where a duty attaches. Instruments that are still a draft or
not yet applicable are shown with a \emph{dashed} outline. The original
wordings of the provisions stated here are quoted in the appendix for
your convenience.}

\end{figure}%

Figure~\ref{fig-regulatory-map} provides a visual overview of their
relations. For convenience, the corresponding provisions are quoted in
the Appendix, e.g., Section~\ref{sec-app-aiact} presents the AIA
provisions relevant for this section.

\begin{figure}

\centering{

\includegraphics[width=0.5\linewidth,height=\textheight,keepaspectratio]{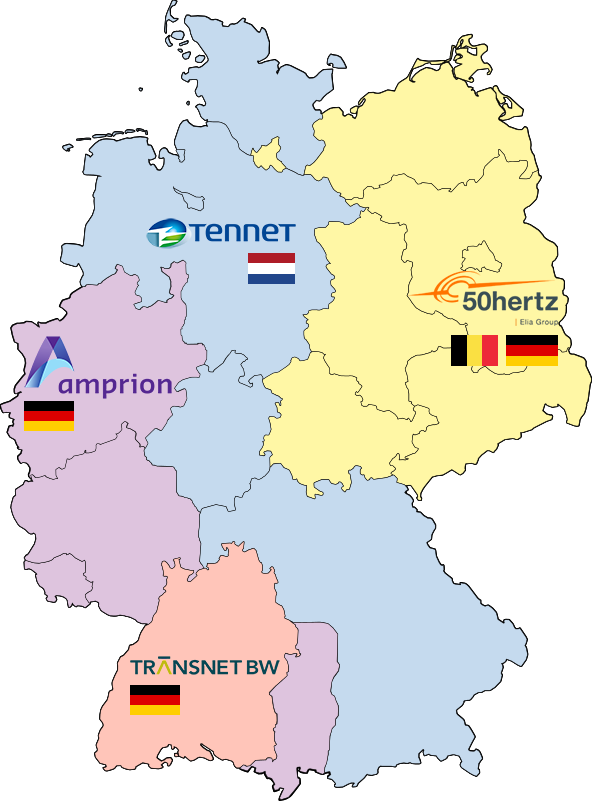}

}

\caption{\label{fig-regelzonen}Control areas of the four transmission
system operators (TSO) for electrical power in Germany. Attribution:
Francis McLloyd, CC BY-SA 3.0
\url{https://creativecommons.org/licenses/by-sa/3.0}, via Wikimedia
Commons.}

\end{figure}

The forecasting approaches considered in the challenge treat the German
market as a single aggregated zone. It is the largest zone in Europe by
annual electricity demand (ENTSO-E 2024). There are four local German
transmission zones (50Hertz, Amprion, TenneT, and TransnetBW), which are
shown in Figure~\ref{fig-regelzonen}.

The ENTSO-E Transparency Platform reports German load under two
aggregations: the bidding zone ``DE-LU'' (Germany and Luxembourg) and
the country aggregation ``DE'' (Germany without Luxembourg). The ``DE''
country aggregation is used as the target series of this paper and the
ground truth of the challenge. Its 2022 total of 482 TWh and hourly peak
of 78.7 GW lie approximately one percent below the DE-LU bidding-zone
values.

Supplying the general public with electricity is a critical service
under the German Kritisverordnung (Bundesministerium des Innern 2026).
Transmission networks become critical installations once final consumers
and redistributors withdraw 3,700 GWh per year
(Section~\ref{sec-app-kritisv-2}, Section~\ref{sec-app-kritisv-anhang}).
Although STLF pipelines are used to manage transmission and trading
functions, the Kritisverordnung governs only the \emph{physical}
resilience of installations and does \emph{not} mention forecasting,
software, or STLF models.

Information security is governed separately by the BSI-Gesetz (Deutscher
Bundestag 2025), which is an information-security statute throughout,
see Section~\ref{sec-app-bsig} and Section~\ref{sec-app-bsig-title}. The
definitions of critical installation, critical service, and security in
information technology (§ 2 Nr. 22, 24, and 39) are listed in
Section~\ref{sec-app-bsig-2}, the operator definition of § 28 Abs. 8 is
listed in Section~\ref{sec-app-bsig-28}, and the IT-security obligations
of § 30 Abs. 1 and § 31 Abs. 2 are shown in
Section~\ref{sec-app-bsig-30} and Section~\ref{sec-app-bsig-31}.

The BSI-Gesetz and the Kritisverordnung are connected, because both
transpose the European legislative package of 14 December 2022. The
BSI-Gesetz implements the NIS-2 Directive (Directive (EU) 2022/2555 on a
high common level of cybersecurity) (European Parliament and Council of
the European Union 2022a), which imposes cybersecurity risk-management
and reporting duties on essential and important entities and lists
energy first among its sectors of high criticality. The Kritisverordnung
is issued under the KRITIS-Dachgesetz, which transposes the sibling CER
Directive (Directive (EU) 2022/2557 on the resilience of critical
entities) (European Parliament and Council of the European Union 2022b).
The AIA also ties into this package, because it uses the CER Directive
for its definition of ``critical infrastructure''. Recital 55, quoted in
Section~\ref{sec-app-recital-55}, cites that directive's annex. The AIA
in turn is relevant to energy applications only where AI systems serve
as safety components Section~\ref{sec-app-recital-55}. STLF pipelines
are therefore located \emph{alongside} critical infrastructure rather
than \emph{inside} its safety perimeter. Nonetheless, we integrate
determinism, fail-safe handling, a minimal attack surface, and the
absence of dead code into the STLF pipeline.

Two further instruments, which both address the supplier rather than the
operator, complete the picture: the CRA and the revised Product
Liability Directive (PLD) (European Parliament and Council 2024a). The
CRA can be considered as the \emph{product-side} counterpart to the
operator-side risk management required by the BSI-Gesetz. CRA Article
12(1) links the two EU regulations directly: a product that is also a
high-risk AI system fulfils the cybersecurity requirement of Article 15
of the AIA once it satisfies the essential requirements of Annex I and
demonstrates this in its declaration of conformity
(Section~\ref{sec-app-cra-12}). The standards that would carry that
presumption are not finalised yet (European Commission 2025). The
revised PLD makes software a product in its own right
(Section~\ref{sec-app-pld-4}) and counts safety-relevant cybersecurity
requirements among the circumstances that determine defectiveness
(Section~\ref{sec-app-pld-7}). The PLD will be transposed into German
law (Deutscher Bundestag 2026).

The CRA and the PLD place open-source software supplied outside a
commercial activity beyond the manufacturer's obligations. This applies
to \texttt{spotforecast2-safe} only if it is not integrated into a
commercial product (Section~\ref{sec-app-cra-oss},
Section~\ref{sec-app-pld-2}). Because the CRA applies from 11 December
2027 and the PLD covers products offered after 9 December 2026, neither
instrument applies to the live challenge evaluated here.

The software design of the STLF reference pipeline provided to the
students as a template builds on two existing artefacts. The first is
the open energy-demand forecaster of Chagnet (2025), a software pipeline
for the French market that combines \texttt{skforecast}, LightGBM,
ENTSO-E data ingestion, weekly retraining, and Bayesian hyperparameter
tuning with Optuna. Its retargeted and extended version was adopted as a
pipeline template. The second is the open-source software
\texttt{spotforecast2-safe} (Bartz-Beielstein and Bartz 2026), which
implements the STLF pipeline as a reviewed, deterministic subset of the
\texttt{skforecast} recursive strategy (Amat Rodrigo and Escobar Ortiz
2024). This guarantees that the resulting system is reproducible and
auditable.

This report makes four contributions. It presents an end-to-end
day-ahead STLF pipeline for the German market zone that is
deterministic, reproducible, and auditable by construction. It reports
the accuracy of the system against the ENTSO-E baseline. It compares
recursive forecasting approaches with newer methods based on foundation
models and in-context learning. Finally, it discusses the implications
for operational and regulatory requirements in safety-critical
environments.

The two newer methods used in this challenge are Chronos-2 (Ansari et
al. 2025), a pretrained model with more than 100 million parameters, and
BABL2L\footnote{BABL2L was called MACL2L (Mac Learning to Learn) until
  November 2026, and its implementation package was renamed from
  \texttt{macl2l} to \texttt{babl2l}. All BABL2L results in this report
  were produced with the predecessor package \texttt{macl2l} during the
  challenge. The leaderboard identifiers \texttt{macl2l} and
  \texttt{macl2l\_entsoe} in the data bundle
  (Section~\ref{sec-app-data}) keep the old name.}, a small, locally
trained model that runs efficiently and does not require
high-performance hardware. Both models are evaluated on the same ENTSO-E
data and under the same challenge protocol as the recursive forecasters.

The remainder of the paper is organised as follows.
Section~\ref{sec-methods} describes the data, preprocessing, feature
construction, and evaluation protocol. Section~\ref{sec-recursive}
details the recursive multi-step forecasting algorithm, and
Section~\ref{sec-tuning} presents the \texttt{spotoptim} and Optuna
hyperparameter tuners. Section~\ref{sec-teams} states the rules that
every submission had to satisfy and records the teams' accounts of their
forecasters. Section~\ref{sec-results} reports the empirical findings.
Section~\ref{sec-discussion} interprets them with respect to operational
and regulatory requirements. Finally, Section~\ref{sec-conclusion}
summarises the work and outlines directions for future research.

\section{Materials and Methods}\label{sec-methods}

Table~\ref{tbl-notation} (in the Appendix, see
Section~\ref{sec-notation}) summarises the notation used throughout the
paper.

The STLF pipeline is a deterministic workflow that converts raw ENTSO-E
transmission-grid measurements into a day-ahead forecast. Each student
team received a reference implementation that could be modified freely
for its own submissions. The reference pipeline downloads aggregated
load data from the ENTSO-E Transparency Platform (ENTSO-E 2024),
annotates the series, and flags anomalies and gaps. Calendar and weather
covariates are then added, and a LightGBM regressor (Ke et al. 2017) is
used to build the forecaster. The resulting model predicts the 24-hour
horizon of a target day recursively. Forecast quality is evaluated daily
on the leaderboard by MAE against the actual load.

\subsection{Code design and process rules}\label{sec-code}

The properties that Section~\ref{sec-intro} requires of the STLF
pipeline and that every team of the challenge had to respect, namely
determinism, fail-safe handling, a minimal attack surface, and the
absence of dead code, are inherited from the \texttt{spotforecast2-safe}
package. The package is developed under the eight rules defined by
Bartz-Beielstein and Bartz (2026). Four code-development rules (CR-1 to
CR-4) restrict what the source code may contain. Four additional process
rules (PR-1 to PR-4) restrict how the package is developed, shipped, and
operated.

\begin{tcolorbox}[enhanced jigsaw, arc=.35mm, bottomrule=.15mm, bottomtitle=1mm, breakable, colback=white, colbacktitle=quarto-callout-note-color!10!white, colframe=quarto-callout-note-color-frame, coltitle=black, left=2mm, leftrule=.75mm, opacityback=0, opacitybacktitle=0.6, rightrule=.15mm, title=\textcolor{quarto-callout-note-color}{\faInfo}\hspace{0.5em}{Code-development rules}, titlerule=0mm, toprule=.15mm, toptitle=1mm]

The code-development rules are enforced at commit time (Bartz-Beielstein
and Bartz 2026):

\begin{itemize}
\tightlist
\item
  CR-1, no dead code, requires tests and executable docstring examples
  for functions and classes.
\item
  CR-2, deterministic transformations, requires that the same input
  results in the same bit-level output.
\item
  CR-3, fail-safe handling, requires that invalid or missing inputs
  raise an explicit exception.
\item
  CR-4, a minimal attack surface, keeps a short, versioned blocklist of
  forbidden dependencies.
\end{itemize}

\end{tcolorbox}

\begin{tcolorbox}[enhanced jigsaw, arc=.35mm, bottomrule=.15mm, bottomtitle=1mm, breakable, colback=white, colbacktitle=quarto-callout-note-color!10!white, colframe=quarto-callout-note-color-frame, coltitle=black, left=2mm, leftrule=.75mm, opacityback=0, opacitybacktitle=0.6, rightrule=.15mm, title=\textcolor{quarto-callout-note-color}{\faInfo}\hspace{0.5em}{Process rules}, titlerule=0mm, toprule=.15mm, toptitle=1mm]

At review or release time, the process rules are applied
(Bartz-Beielstein and Bartz 2026). This can be implemented by
continuous-integration workflows, for example on GitHub Actions or
GitLab CI/CD.

\begin{itemize}
\tightlist
\item
  PR-1, traceability, is implemented as documentation coverage: a
  release guard fails when a public symbol is undocumented.
\item
  PR-2, a documented threat model, keeps a STRIDE table (spoofing,
  tampering, repudiation, information disclosure, denial of service,
  elevation of privilege) beside the code it describes.
\item
  PR-3, supply-chain integrity, calls for a software bill of materials.
\item
  PR-4, a structured audit log, has every operational action emit a
  record using a pinned, versioned logging schema.
\end{itemize}

\end{tcolorbox}

\subsection{spotforecast2 and
spotforecast2-safe}\label{sec-spotforecast2}

\begin{definition}[spotforecast2-safe, the core
library]\protect\hypertarget{def-sf-safe}{}\label{def-sf-safe}

\texttt{spotforecast2-safe}, also referred to as \texttt{sf2-safe}, is a
specialised Python library for time-series forecasting in
safety-critical production environments. Unlike standard
machine-learning and deep-learning libraries, it adopts a safety-first
architecture by design. In particular, it implements the principles
summarised as CR-1 to CR-4 and PR-1 to PR-4 in Section~\ref{sec-code}.

\href{https://git-ce.rwth-aachen.de/thomas.bartz-beielstein/spotforecast2-safe/-/blob/main/MODEL_CARD.md}{MODEL\_CARD.md}
gives a detailed technical overview of its safety mechanisms. The
package is available as open source (Bartz-Beielstein 2026d).

\end{definition}

\begin{definition}[spotforecast2, the extended
library]\protect\hypertarget{def-sf}{}\label{def-sf}

\texttt{spotforecast2}, also referred to as \texttt{sf2}, extends
\texttt{spotforecast2-safe} with visualisation, hyperparameter tuning,
and additional features. It is available as open source
(Bartz-Beielstein 2026c).

\end{definition}

\subsection{Data and study domain}\label{sec-data}

The aggregated load of the DE market zone, taken as the ENTSO-E Actual
Total Load (ENTSO-E code 6.1.A),\footnote{The identifier 6.1.A is
  ENTSO-E's article-based catalogue number for a data item, while the
  official day-ahead forecast that is used as the ENTSO-E baseline below
  is the sibling item 6.1.B, the day-ahead total load forecast (ENTSO-E
  2024).} is the target value. The series is hourly, measured in MW, and
indexed by Coordinated Universal Time (UTC) timestamps. The ENTSO-E
Transparency Platform application programming interface (API) (ENTSO-E
2024) is used for retrieval, and the Open-Meteo API is used for
accessing the weather covariates. The recursive forecasters from the
STLF reference pipeline are trained on a window of three years of
history. Every submission must contain predictions for all 24 hours of
the next day (target day).

\begin{figure}

\centering{

\pandocbounded{\includegraphics[keepaspectratio,alt={Horizontal bar chart of the mean bias of each leaderboard entry. Values ordered by magnitude around zero.}]{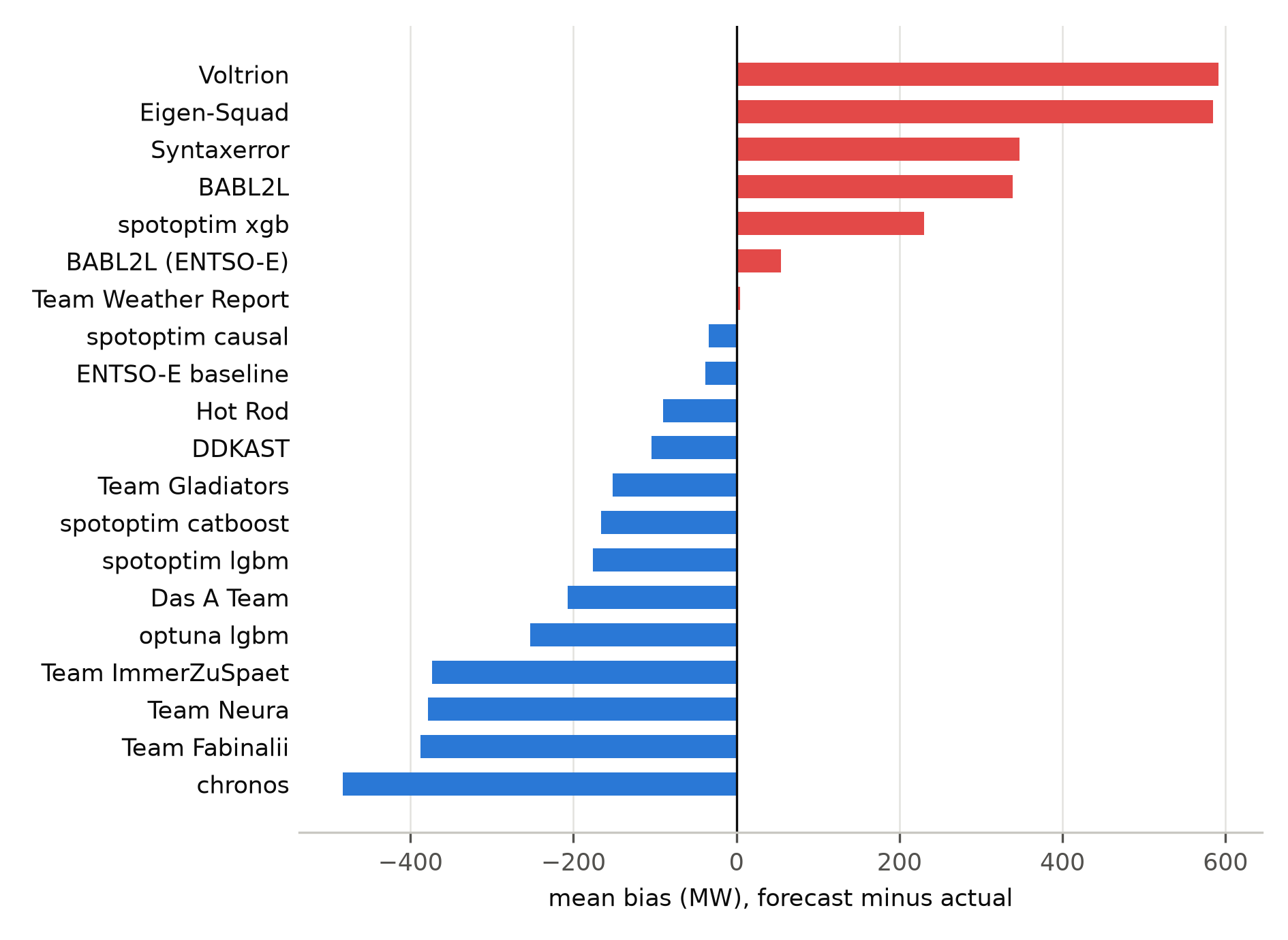}}

}

\caption{\label{fig-bias}Mean bias per leaderboard entry, in MW. Values
computed from the challenge leaderboard data
(\url{https://bartzbeielstein.github.io/challenge-leaderboard/},
snapshot of 21 July 2026, completed live phase of 41 scored days). The
bias is defined as the mean of forecast minus actual (see
Equation~\ref{eq-bias}). A negative value indicates systematic
under-forecast. The ENTSO-E baseline reaches \(-38.42\) MW.}

\end{figure}

Following Möbius et al. (2025), the ENTSO-E baseline is treated as an
operational forecast rather than an unbiased reference. Analysing the
published DE-LU series from 2016 to 2019, Möbius et al. (2025) report
systematic under-prediction, with an average bias of 881 MW, an MAE of
1776 MW, or 3.14\% of mean load, and autocorrelated hourly errors. The
direction of the error is time-dependent: under-prediction occurs on
weekdays and over-prediction at weekends, and the largest deviations
occur in the morning and evening hours (see also
Section~\ref{sec-error-diagnostics}). Table 2 in Möbius et al. (2025)
shows that modelling the error series on its own, i.e., without any
load-specific covariate, lowers its root mean squared error (RMSE) by
about 21\%. Therefore, the mean bias of Equation~\ref{eq-bias} and the
under-prediction rate (UPR) of Equation~\ref{eq-upr} are reported on the
challenge leaderboard together with the MAE, the RMSE, and the mean
absolute percentage error (MAPE). Figure~\ref{fig-bias} shows that over
the 41 scored days of the completed challenge the ENTSO-E baseline has a
mean bias of \(-38.42\) MW against an MAE of \(2155.61\) MW. Its
forecasts fall below the actual load in 51.4\% of hours. The direction
reported in Möbius et al. (2025) is still observable, but the constant
component is more than an order of magnitude smaller. Correcting the
mean bias alone is therefore not sufficient for the challenge.

\subsection{Overview of load-forecasting methods}\label{sec-overview}

Table~\ref{tbl-stlf-taxonomy} shows the forecasters studied in this
paper within the wider landscape of forecasting methods. The reader is
referred to Ullah et al. (2024) and Hyndman et al. (2026) for a
comprehensive survey of the field. Amat Rodrigo and Escobar Ortiz (2026)
provide a description of the \texttt{skforecast} package and
Bartz-Beielstein (2026b) describes the \texttt{spotoptim} package.

The recursive spotoptim-lgbm forecaster belongs to the category called
``time series forecasting with machine learning'' (TSFML)
(Bartz-Beielstein and Bartz 2026; Amat Rodrigo and Escobar Ortiz 2024;
Amat Rodrigo and Escobar Ortiz 2026). In TSFML the forecasting problem
is reduced to tabular supervised regression: the series is recast as a
matrix of lagged values and extended with exogenous covariates.
\texttt{skforecast} and \texttt{spotforecast2-safe} allow any regressor
with a \texttt{scikit-learn} interface to be trained in this way
(Bontempi et al. 2013). Multi-step forecasts can be computed by applying
the fitted one-step model recursively, which implements one of the
multi-step strategies reviewed by Ben Taieb et al. (2012). Model
selection and evaluation rely on backtesting over rolling forecasting
origins (Tashman 2000; Bergmeir and Benı́tez 2012). In the time-series
classification framework of Januschowski et al. (2020) the TSFML
approach is considered ``machine-learning based'' because of this
transformation and not because a neural architecture is used.
Table~\ref{tbl-stlf-taxonomy} therefore lists TSFML as a separate
category. The applicability of TSFML with gradient-boosting pipelines to
the ENTSO-E prediction task has been demonstrated by Chagnet (2025).
LightGBM (Ke et al. 2017) and related gradient-boosted decision-tree
implementations are among the most widely used algorithms in the TSFML
approach, as shown, for example, by the winning entries of the M5
accuracy competition (Makridakis et al. 2020; Januschowski et al. 2022;
Ben Taieb and Hyndman 2014). As reported by Elsayed et al. (2021),
LightGBM approaches are competitive with deep-learning architectures on
tabularised forecasting tasks. Whether TSFML approaches are competitive
with recent large pretrained foundation models is an open question.

\texttt{spotforecast2-safe} implements the TSFML approach through its
vendored subset of the \texttt{skforecast} recursive strategy
(Section~\ref{sec-spotforecast2}), and Section~\ref{sec-recursive}
explains the resulting spotoptim-lgbm forecaster in detail.

\begin{longtable}[]{@{}
  >{\raggedright\arraybackslash}p{(\linewidth - 2\tabcolsep) * \real{0.3333}}
  >{\raggedright\arraybackslash}p{(\linewidth - 2\tabcolsep) * \real{0.6667}}@{}}
\caption{Classification of forecasting methods, based on Hyndman et al.
(2026). It was extended by the category \emph{time series forecasting
with machine learning} (Bontempi et al. 2013; Januschowski et al. 2020).
ARIMA denotes the \emph{autoregressive integrated moving average model},
and ETS abbreviates \emph{error, trend, seasonal}, the state-space form
of exponential smoothing. The last row adds the three forecasters
evaluated in this paper: the recursive spotoptim-lgbm forecaster, a
member of the TSFML category, and the pretrained models Chronos-2 and
BABL2L, members of the foundation-model
category.}\label{tbl-stlf-taxonomy}\tabularnewline
\toprule\noalign{}
\begin{minipage}[b]{\linewidth}\raggedright
Category
\end{minipage} & \begin{minipage}[b]{\linewidth}\raggedright
Members
\end{minipage} \\
\midrule\noalign{}
\endfirsthead
\toprule\noalign{}
\begin{minipage}[b]{\linewidth}\raggedright
Category
\end{minipage} & \begin{minipage}[b]{\linewidth}\raggedright
Members
\end{minipage} \\
\midrule\noalign{}
\endhead
\bottomrule\noalign{}
\endlastfoot
Time series regression models & linear model, predictor selection,
nonlinear regression \\
Exponential smoothing & simple exponential smoothing, trend and seasonal
methods, ETS state-space models \\
ARIMA models & autoregressive models, moving average models, seasonal
and non-seasonal ARIMA models \\
Dynamic regression models & regression with ARIMA errors, dynamic
harmonic regression, lagged predictors \\
Forecasting hierarchical and grouped time series & bottom-up and
top-down approaches, forecast reconciliation \\
Advanced forecasting methods & complex seasonality, Prophet, vector
autoregressions, bootstrapping and bagging \\
Time series forecasting with machine learning & reduction to tabular
regression on lagged values, recursive and direct multi-step strategies,
gradient-boosted decision trees, global forecasting models \\
Neural networks & multilayer perceptron, modern neural network
architectures \\
Foundation forecasting models & transfer learning, pretrained foundation
models \\
This paper & spotoptim-lgbm, Chronos-2, BABL2L, implementations from
student teams \\
\end{longtable}

\subsection{Reference pipeline provided to the
teams}\label{sec-template}

The three subsections that follow describe the TSFML-reference pipeline
handed to every student team at the start of the challenge: the
anomaly-aware preparation of the raw ENTSO-E series, the covariates
derived from it, and the resulting data-set. Together with the recursive
forecaster of Section~\ref{sec-recursive}, they constitute the
spotoptim-lgbm configuration. This configuration is both the
organiser-operated entry evaluated in Section~\ref{sec-results} and the
starting point that teams were free to modify. Section~\ref{sec-teams}
shows the rules that governed those modifications and presents the
modifications made by the student teams.

\subsubsection{Outlier detection and data
preparation}\label{sec-outliers}

Because the forecasting system operates in a safety-critical setting
governed by the AIA (European Parliament and Council of the European
Union 2024), gaps and corruptions in the input series are annotated and
treated explicitly (``healed'') rather than silently imputed. This
gap-aware design choice follows the data-governance principles of
Bartz-Beielstein and Bartz (2026). The production preparation follows
three stages, with a resampling boundary between the first two:
target-corruption checks on the native fifteen-minute data, anomaly
flagging after hourly resampling, and gap imputation thereafter. A
step-to-step walkthrough of the code used in these three stages is
presented in Section~\ref{sec-software}.

\paragraph{Stage 1: Target-corruption
flagging}\label{stage-1-target-corruption-flagging}

The first stage applies domain-specific target-corruption rules. In the
multi-task execution, \texttt{prepare\_data} runs it before resampling,
while the fifteen-minute observations needed for its intra-hour checks
are still available. These rules were tailored to the characteristic
dropouts of the ENTSO-E Actual Load series: (i) An \emph{intra-hour
range check} verifies the consistency of the four fifteen-minute slots
making up an hour. (ii) An \emph{adjacent-step check} rejects
implausible jumps between consecutive fifteen-minute slots, and (iii) a
\emph{deviation check} marks excessive shortfalls below a reference
series, namely the ENTSO-E baseline. The configured corruption policy
then determines the response: \texttt{abort} refuses the run,
\texttt{heal} marks the flagged target slots as missing for later
interpolation, and \texttt{truncate} removes the corrupted tail from the
usable target history. After this check, the pipeline resamples the data
to hourly values before Stage 2's \texttt{detect\_outliers} call.

\paragraph{Stage 2: Unsupervised anomaly
flagging}\label{stage-2-unsupervised-anomaly-flagging}

The second stage screens the hourly series for outliers and anomalies.
It separates detection from removal, because only user-specified
hard-coded plausibility bounds trigger removal. Values outside those
bounds are set to missing, and the subsequent imputation stage handles
them as gaps. The Isolation-Forest detector (Liu et al. 2008), fitted
independently to each column with a contamination parameter that states
the expected fraction of anomalies, is only advisory in the production
pipeline: it reports suspicious points to the log for operator review
and does not modify the data.

\texttt{spotforecast2-safe} also provides a cleaning variant that
returns a copy with flagged points set to missing, together with an
indicator of which points were flagged
(Section~\ref{sec-software-stage2}). It leaves its input series
unchanged. A fixed random seed together with the protocolled
contamination parameter keeps the flagging, the whole preparation
pipeline, and the recursive forecaster of Section~\ref{sec-recursive}
reproducible across runs.

\paragraph{Stage 3: Gap imputation}\label{stage-3-gap-imputation}

The third stage fills gaps created by hard-bound removal or by a
target-corruption policy. The guiding idea is that an imputed value may
complete the series but should not be trusted as training evidence.
Every imputed hour, together with the subsequent hours whose features
look back into the gap, is therefore assigned a training weight of zero.
The values themselves are copied from the nearest observed hours by
default or interpolated linearly between the gap's endpoints, and both
the default and the healing configuration apply the zero-weight rule.

Gaps at the recent end of the series are never filled: where no
observation exists yet, the series is truncated to the last observed
hour and the forecast span is extended to compensate. Whether a flagged
stretch may be healed at all is governed by the stricter policy of Stage
1, whose default is to abort the run. When Stage 1 heals a flagged
stretch, that decision changes the later imputation choice to weighted
interpolation so that the repaired target values receive the same
zero-weight protection. Consequently, imputation is limited and
user-controlled. Advisory Isolation-Forest detections alone do not
create gaps to fill.

\begin{tcolorbox}[enhanced jigsaw, arc=.35mm, bottomrule=.15mm, bottomtitle=1mm, breakable, colback=white, colbacktitle=quarto-callout-note-color!10!white, colframe=quarto-callout-note-color-frame, coltitle=black, left=2mm, leftrule=.75mm, opacityback=0, opacitybacktitle=0.6, rightrule=.15mm, title=\textcolor{quarto-callout-note-color}{\faInfo}\hspace{0.5em}{Role and usage of the ENTSO-E baseline forecast}, titlerule=0mm, toprule=.15mm, toptitle=1mm]

Because the ENTSO-E baseline is the comparison standard of the
challenge, standard entries must not use it as a model input.
\texttt{spotforecast2-safe} implements this rule in code rather than
merely stating it.

The baseline serves three non-predictive roles. (i) The deviation check
of the first stage uses it as a reference to spot implausible shortfalls
in the measured load. (ii) Warn-only plausibility checks compare the
shape and level of a finished forecast against a reference profile
before submission. They alert the operator without blocking or altering
the submission. (iii) Diagnostic plots overlay the baseline on the
submitted forecast for visual inspection after the fact.

Three mechanisms ensure that the baseline predictions are not used for
model training. (i) The information flow: every check compares numbers
and raises a flag. No data is copied from the baseline into the training
data, and where a flagged load value is repaired, the replacement is
interpolated from the load series itself. (ii) An automatic guard: after
training, the function \texttt{assert\_no\_leakage} inspects the
training data, the selected features, and the features recorded inside
the fitted model, and aborts the run before a submission is written if
the baseline appears in any of them. (iii) The configuration: the
switches that would admit the baseline as a covariate exist in the
library but are switched off in every standard entry. In plain terms,
the pipeline treats the official forecast the way an examiner treats a
reference solution: it may be used to judge and to flag, but never to
copy from.

The only exception is clearly defined: an entry may use the baseline for
predictions if it carries the \texttt{entsoe} suffix required by the
challenge rules.

\end{tcolorbox}

\subsubsection{Covariates}\label{sec-covariates}

The exogenous feature vector \(\mathbf{x}_t\) contains the exogenous
values that are available in addition to the load history. It comprises
three groups. The deterministic calendar features are the periodic
components month, week of year, day of week, and hour of day, which
enter the model through a cyclical sine/cosine encoding so that the
encoded coordinates wrap around the period. The holiday-derived integer
indicators are computed from the German holiday calendar (nation-wide
holidays plus those of the state of North Rhine-Westphalia). They
consist of a public-holiday flag, three adjacency flags marking the day
before a holiday, the day after a holiday, and bridging working days
(\emph{Brückentage}), and a day-type pair consisting of a working-day
flag and a four-class day-type code that distinguishes working days,
Saturdays, Sundays, and public holidays, with holidays taking
precedence. Weekend information therefore enters through the day-type
classes rather than through a separate weekend flag, and the integer
coding is unproblematic for a tree-based learner, which splits it by
thresholding. The weather covariates are variables such as air
temperature from a numerical weather service (Open-Meteo).

A key constraint of the TSFML forecaster decides which covariates are
admissible: every exogenous covariate must be known, or itself forecast,
over the entire prediction horizon, because the value
\(\mathbf{x}_{T+h}\) is required at every step \(h = 1,\dots,H\) of the
recursion described in Section~\ref{sec-recursive}, where \(T\) denotes
the forecast origin and \(H\) the horizon. The calendar and holiday
features satisfy this requirement trivially, because they are
deterministic functions of the timestamp and the published holiday
calendar and can be evaluated for any future hour. In contrast, weather
and other measured drivers are not deterministically known. They must be
supplied by a forecast of their own that spans the target day before
they can be included in \(\mathbf{x}_{T+h}\).

\subsubsection{The complete data-set}\label{sec-dataset}

Table~\ref{tbl-dataset} lists the complete data-set of the recursive
forecaster: the endogenous target series, the exogenous covariate groups
that form \(\mathbf{x}_t\), and the one reference series that is
deliberately not included in the TSFML model. The ENTSO-E Actual Total
Load is the only endogenous data. After the preparation of
Section~\ref{sec-outliers} the native fifteen-minute series is
aggregated to hourly mean values. The forecaster computes the
autoregressive inputs from this single series, i.e., the lags
\(\mathcal{L}\) of Table~\ref{tbl-notation} as well as a rolling mean
over the preceding 72 hours that summarises the recent load level.

\begin{longtable}[]{@{}
  >{\raggedright\arraybackslash}p{(\linewidth - 8\tabcolsep) * \real{0.1333}}
  >{\raggedright\arraybackslash}p{(\linewidth - 8\tabcolsep) * \real{0.2667}}
  >{\raggedright\arraybackslash}p{(\linewidth - 8\tabcolsep) * \real{0.2000}}
  >{\raggedright\arraybackslash}p{(\linewidth - 8\tabcolsep) * \real{0.1333}}
  >{\raggedright\arraybackslash}p{(\linewidth - 8\tabcolsep) * \real{0.2667}}@{}}
\caption{The complete data-set of the TSFML forecaster. The first row
lists the endogenous target, the middle rows the exogenous covariate
groups of \(\mathbf{x}_t\), and the last row the reference series that
the leakage guard of Section~\ref{sec-outliers} keeps out of the
model.}\label{tbl-dataset}\tabularnewline
\toprule\noalign{}
\begin{minipage}[b]{\linewidth}\raggedright
Group
\end{minipage} & \begin{minipage}[b]{\linewidth}\raggedright
Series
\end{minipage} & \begin{minipage}[b]{\linewidth}\raggedright
Source
\end{minipage} & \begin{minipage}[b]{\linewidth}\raggedright
Cadence
\end{minipage} & \begin{minipage}[b]{\linewidth}\raggedright
Enters the model as
\end{minipage} \\
\midrule\noalign{}
\endfirsthead
\toprule\noalign{}
\begin{minipage}[b]{\linewidth}\raggedright
Group
\end{minipage} & \begin{minipage}[b]{\linewidth}\raggedright
Series
\end{minipage} & \begin{minipage}[b]{\linewidth}\raggedright
Source
\end{minipage} & \begin{minipage}[b]{\linewidth}\raggedright
Cadence
\end{minipage} & \begin{minipage}[b]{\linewidth}\raggedright
Enters the model as
\end{minipage} \\
\midrule\noalign{}
\endhead
\bottomrule\noalign{}
\endlastfoot
Endogenous & Actual Total Load \(y_t\) of the DE market zone & ENTSO-E
Transparency Platform (code 6.1.A) & 15 min, aggregated to hourly means
& prediction target; autoregressive lags \(\mathcal{L}\) and a 72-hour
rolling mean \\
Exogenous, calendar & hour of day, day of week, week of year, month &
deterministic function of the UTC timestamp & hourly & one sine/cosine
pair per feature (eight columns) \\
Exogenous, solar & sunrise hour, sunset hour & astronomical computation
for the reference location & hourly & one sine/cosine pair per feature
(four columns) \\
Exogenous, holiday and day-type & public-holiday flag, day before/after
holiday, \emph{Brückentag}, working-day flag, four-class day type &
deterministic function of the timestamp and the German holiday calendar
(DE, state NW) & hourly & six integer columns, used as delivered \\
Exogenous, weather & air temperature (2 m), relative humidity (2 m),
precipitation, rain, snowfall, weather code, mean-sea-level pressure,
surface pressure, and total, low, mid, and high cloud cover &
Open-Meteo: archive for the history, weather forecast over the horizon &
hourly & twelve columns, used as delivered \\
Reference, non-predictive & day-ahead load forecast of the DE market
zone & ENTSO-E Transparency Platform (code 6.1.B) & 15 min & never a
covariate; deviation QC, shape check, and diagnostic plots only \\
\end{longtable}

Every exogenous covariate satisfies the admissibility constraint of
Section~\ref{sec-covariates}. In total, the TSFML model uses thirty
exogenous columns in addition to the lagged target\footnote{The
  covariate configuration described here is the one active from 19 July
  2026 onward. The 39 leaderboard days scored before that date were
  produced with a reduced set of twenty-four exogenous columns.}.

\subsection{Challenge design}\label{sec-challenge}

The challenge ran as a public leaderboard repository on GitHub, and its
archived final state is available at
\url{https://bartzbeielstein.github.io/challenge-leaderboard/}. Eleven
student teams from two courses at TH Köln participated. In addition to
the student teams' entries, eight organiser-operated models, the ENTSO-E
baseline, and two seasonal-naive benchmarks computed from the ground
truth were scored on the leaderboard. For every target day, the 24
forecasts were due by 23:59 UTC on the preceding day. They were
submitted to the repository through a fork-and-pull-request workflow.
Each day was scored after the official ENTSO-E values were published.
Whenever ENTSO-E revised the actuals within a 21-day look-back window,
the affected scores were recomputed. Two scoring rules managed missing
and corrected submissions: (i) a team that missed a day had its most
recent submission carried forward and scored in its place, and (ii)
every team held a single joker to replace one already-scored day by a
fresh forecast. After a preliminary testing phase, the live challenge
started on 10 June 2026 and closed with its 41st target day on 20 July
2026. Every team's result had to be reproduced by a different team, and
this peer reproduction was certified for all eleven teams.

\subsection{Evaluation metrics}\label{sec-metrics}

All metrics are computed over the \(H=24\) hourly forecasts of the
target day, i.e., the hours \(T+1,\dots,T+H\) that follow the forecast
origin \(T\). Each forecast \(\hat{y}_{T+h\mid T}\) is compared with the
reported ENTSO-E load \(y_{T+h}\) in the notation of
Table~\ref{tbl-notation}.

In addition to the primary ranking metric, the MAE
(Definition~\ref{def-mae}), we report the RMSE, the MAPE, the mean bias,
the UPR, and, for comparability across series of different scale, the
mean absolute scaled error (MASE). Mathematical definitions of all
errors used in this report are given in the Appendix
(Section~\ref{sec-definitions}).

\section{The recursive LightGBM forecaster}\label{sec-recursive}

The TSFML forecaster wraps a \texttt{scikit-learn}-compatible regressor
into a recursive autoregressive multi-step predictor. The regressor used
here is a LightGBM gradient-boosted decision-tree (GBDT) ensemble (Ke et
al. 2017; Friedman 2001), and the recursive wrapping strategy is ported
from \texttt{skforecast} (Amat Rodrigo and Escobar Ortiz 2024) into the
deterministic engine of \texttt{spotforecast2-safe} (Bartz-Beielstein
and Bartz 2026). Forecasting the full day requires lags that extend
beyond the forecast origin \(T\), where the true load is not yet known.
The recursive strategy resolves this by substituting earlier forecasts
for the missing actuals. Because the predictor also requires future
covariate vectors, every covariate \(\mathbf{x}_{T+h}\) must be
available across the full horizon at prediction time. Deterministic
functions of the timestamp satisfy this requirement exactly, whereas
weather-derived covariates must be supplied by a forecast.

In the baseline configuration, the TSFML forecaster operates on a single
aggregated German Actual Total Load series with a daily seasonal
encoding, a lag set selected by the tuner from the pool of
Table~\ref{tbl-searchspace}, a horizon of \(H=24\) hours, a three-year
training window, and a fixed random seed.

\subsection{Choice of the GBDT implementation}\label{sec-gbdt-choice}

The challenge forecaster is built on LightGBM, one of the three most
widely used GBDT implementations alongside XGBoost (Chen and Guestrin
2016) and CatBoost (Prokhorenkova et al. 2018). The three differ in two
respects that matter for this application: how each tree is grown and
how categorical columns are handled.

Tree growth separates the implementations most visibly. XGBoost expands
its trees level by level, which keeps model growth predictable and
comparatively robust against overfitting, at the cost of slower
training. LightGBM grows the most promising leaf first, which is fast
but prone to overfitting on small data sets. CatBoost builds trees that
split on a single feature per level, which makes prediction very fast.
The treatment of categorical columns differs just as much. XGBoost
splits natively on sets of category values but applies no smoothing, so
rare categories invite overfitting. LightGBM splits on integer codes
when a column is declared categorical and silently treats an undeclared
column as numeric. CatBoost derives its category encoding from preceding
rows only, so the encoding cannot leak the target. These differences
yield simple selection guidance. XGBoost is the first choice if
numerical features dominate, LightGBM is preferable when the data have
millions of rows or iteration speed matters, and CatBoost is recommended
when many categorical columns are present, especially of high
cardinality.

For the present application, a long hourly series with numeric and
integer-coded covariates and a daily refit schedule, LightGBM is a
natural choice. Since \texttt{spotforecast2-safe} is modular, the
forecaster can be replaced by any other \texttt{scikit-learn}-compatible
regressor, including XGBoost and CatBoost, without changing the
recursive strategy or the tuning procedure. For that reason, all three
methods are evaluated in this paper.

\section{Hyperparameter optimisation}\label{sec-tuning}

The hyperparameters \(\boldsymbol{\theta}\) are not chosen by hand but
by minimising an out-of-sample error estimate obtained by time-series
cross-validation (CV). Formally, tuning searches the space \(\Theta\)
for

\begin{equation}\protect\phantomsection\label{eq-tuning}{
\boldsymbol{\theta}^{*} = \arg\min_{\boldsymbol{\theta}\in\Theta} \operatorname{CV\text{-}MAE}(\boldsymbol{\theta}),
}\end{equation}

where \(\operatorname{CV\text{-}MAE}(\boldsymbol{\theta})\) is the MAE
of Section~\ref{sec-metrics} for the forecaster, estimated by CV at the
candidate configuration \(\boldsymbol{\theta}\). The minimiser
\(\boldsymbol{\theta}^{*}\) is then refitted on the full training window
before evaluation. Table~\ref{tbl-searchspace} lists the nine dimensions
of the search space \(\Theta\): eight LightGBM hyperparameters and, as a
ninth categorical dimension, the choice of the autoregressive lag set.

\begin{longtable}[]{@{}
  >{\raggedright\arraybackslash}p{(\linewidth - 6\tabcolsep) * \real{0.2500}}
  >{\raggedright\arraybackslash}p{(\linewidth - 6\tabcolsep) * \real{0.3333}}
  >{\raggedright\arraybackslash}p{(\linewidth - 6\tabcolsep) * \real{0.2500}}
  >{\raggedright\arraybackslash}p{(\linewidth - 6\tabcolsep) * \real{0.1667}}@{}}
\caption{The hyperparameter search space \(\Theta\) handed to
\protect\texttt{spotoptim} and, identically, to Optuna. The upper bounds
of \protect\texttt{learning\_rate} and \protect\texttt{n\_estimators}
are raised from the package defaults of 0.1 and 1000. Every other
dimension keeps its default
bounds.}\label{tbl-searchspace}\tabularnewline
\toprule\noalign{}
\begin{minipage}[b]{\linewidth}\raggedright
Hyperparameter
\end{minipage} & \begin{minipage}[b]{\linewidth}\raggedright
Meaning
\end{minipage} & \begin{minipage}[b]{\linewidth}\raggedright
Range
\end{minipage} & \begin{minipage}[b]{\linewidth}\raggedright
Type and scale
\end{minipage} \\
\midrule\noalign{}
\endfirsthead
\toprule\noalign{}
\begin{minipage}[b]{\linewidth}\raggedright
Hyperparameter
\end{minipage} & \begin{minipage}[b]{\linewidth}\raggedright
Meaning
\end{minipage} & \begin{minipage}[b]{\linewidth}\raggedright
Range
\end{minipage} & \begin{minipage}[b]{\linewidth}\raggedright
Type and scale
\end{minipage} \\
\midrule\noalign{}
\endhead
\bottomrule\noalign{}
\endlastfoot
\texttt{num\_leaves} & maximum number of leaves per tree & \([8, 256]\)
& integer, linear \\
\texttt{max\_depth} & maximum depth of a tree & \([3, 16]\) & integer,
linear \\
\texttt{learning\_rate} & shrinkage of each tree's contribution &
\([10^{-4}, 0.3]\) & continuous, \(\log_{10}\) \\
\texttt{n\_estimators} & number of trees \(K\) & \([10, 4000]\) &
integer, \(\log_{10}\) \\
\texttt{bagging\_fraction} & fraction of training rows drawn per
boosting iteration & \([0.5, 1]\) & continuous, linear \\
\texttt{feature\_fraction} & fraction of features considered per tree &
\([0.5, 1]\) & continuous, linear \\
\texttt{reg\_alpha} & \(L_1\) regularisation weight & \([0.01, 100]\) &
continuous, linear \\
\texttt{reg\_lambda} & \(L_2\) regularisation weight & \([0.01, 100]\) &
continuous, linear \\
lag set \(\mathcal{L}\) & autoregressive lags & six candidate sets, see
text & categorical \\
\end{longtable}

The six candidate lag sets are \(\{1,\dots,24\}\), \(\{1,\dots,48\}\),
\(\{1,2,24,48\}\), \(\{1,2,23,24,47,48\}\),
\(\{1,2,11,12,23,24,167,168\}\), and
\(\{1,2,3,11,12,22,23,24,47,48,167,168\}\).

The search space \(\Theta\) is the package-default LightGBM
hyperparameter space of \texttt{spotforecast2} with two of its upper
bounds raised. The learning rate is searched over \([10^{-4},\, 0.3]\)
and the number of boosting iterations, that is the number of trees, over
\([10,\, 4000]\), both on a \(\log_{10}\) scale.

Two optimisers explore this identical space. The primary one is
\texttt{spotoptim}, a surrogate-model sequential optimiser that fits a
cheap surrogate to the configurations evaluated so far and proposes each
next configuration by optimising an acquisition criterion over that
surrogate (Bartz-Beielstein 2026b). It is a Python implementation of the
sequential parameter optimisation methodology documented for R by Bartz
et al. (2022). \texttt{spotoptim} starts with an initial design of 50
configurations, followed by further evaluations up to a total budget of
200, with active restarts. As a reference optimiser, the same problem is
handed to the Optuna Tree-structured Parzen Estimator (TPE) sampler
(Akiba et al. 2019). Because both optimisers receive the identical space
\(\Theta\) and the identical evaluation budget, they can be compared
directly.

The inner objective \(\operatorname{CV\text{-}MAE}\) in
Equation~\ref{eq-tuning} is computed by rolling-origin CV on the
three-year training window. Each of the ten folds advances the forecast
origin forward in time, predicts a block of 24 hours, and refits the
model every seven days. No future observation is used for predicting the
past, because every origin uses only data strictly preceding it.

The orchestration of these studies, including data loading, bookkeeping,
and parallel execution of the \texttt{spotoptim} and Optuna tasks, is
handled by the multi-task driver \texttt{spotforecast2}. The
deterministic forecasting engine and the required data-preparation
routines are located in \texttt{spotforecast2-safe}.

\section{Team forecasters}\label{sec-teams}

Every team started from the reference TSFML pipeline of
Section~\ref{sec-template} together with the recursive forecaster of
Section~\ref{sec-recursive}, that is, from the spotoptim-lgbm
configuration built on \texttt{spotforecast2-safe}. From there a team
could submit that configuration unchanged, extend it with further
covariates, a different regressor, or a different tuning strategy, or
replace it entirely with a forecaster of its own design. Whichever
approach a team selected, the four code-development rules CR-1 to CR-4
of Section~\ref{sec-code} were mandatory, and the process rules PR-1 to
PR-4 governed how a package was developed and shipped. The subsections
below summarise the student approaches in each team's own wording. Teams
are ordered alphabetically.

\subsection{Team Fabinalii}\label{sec-team-team-fabinalii}

We began with the Section~\ref{sec-template} pipeline and kept its main
components unchanged: the recursive LightGBM forecaster, the calendar
features, and the Open-Meteo weather data. We also kept a fixed
configuration with 400 trees, a learning rate of 0.05, and 63 leaves,
and we did not perform automated tuning. On 20 June, our submission
reached an MAE of 7,371 MW, the worst result in the field. The best
entries were close to 1,000 MW on that day, so the problem was clearly
in our pipeline. When we compared our forecasts with the actual load, we
found the cause: the short lags 1 to 3 carried Friday's hourly momentum
into Saturday, and the model created a weekday morning ramp on a weekend
day. To address this, we made the lag set depend on the weekday of the
target day. Monday, Saturday, and Sunday use only daily lags from 24 up
to 168 hours. Tuesday to Friday use the reference lags 1, 2, 3, 24, and
168. A backtest with about 43 target days per weekday confirmed the
improvement on weekend days, although it also introduced losses during
the week. We also revisited the 90-day training window, which we had
chosen arbitrarily. Backtests over 131, and later 194, target days
improved steadily at first and then showed diminishing returns, so we
selected 730 days. As for the code rules of Section~\ref{sec-code}, all
submissions came from the registered handles of the three team members
(CR-1). The seed is fixed, LightGBM runs deterministically, and each
submission includes a training snapshot and log, so forecasts can be
replayed exactly (CR-2). Gaps raise an error instead of being filled
silently. After an invalid ENTSO-E value of 1,124,884 MW slipped through
and broke a forecast, we added a plausibility check: load values outside
20,000 to 100,000 MW are flagged and corrected with an audit note
(CR-3). We kept the same runtime dependencies, namely
\texttt{spotforecast2-safe}, \texttt{lightgbm}, \texttt{pandas}, and
\texttt{numpy}, none of which were on the deny-list (CR-4).

\subsection{Team Hot Rod}\label{sec-team-team-hot-rod}

\subsubsection{Forecast construction}\label{forecast-construction}

Team Hot Rod replaced the recursive reference forecaster with a direct
24-hour pipeline called \emph{Voltcrown}, while retaining
\texttt{spotforecast2-safe} for ENTSO-E data access and safety-oriented
preprocessing. One shared LightGBM regressor evaluates the 24
target-hour feature rows in one operation. A forecast produced for an
earlier hour is never fed into a later target hour. The model is
retrained for every target day. Five regressors use the same
configuration: 3,000 trees, a learning rate of 0.05, 31 leaves, a
minimum of 40 samples per leaf, row subsampling of 0.7, and feature
subsampling of 0.9. They use fixed seeds 2026 to 2030, and their hourly
predictions are averaged. The subsampling is the principal regulariser,
and the additional seeds mainly reduce sensitivity to a single unlucky
fit.

Training uses the ENTSO-E German Actual Total Load from 2018--2019
together with 2024 up to the last observation available at the
submission time. The years 2020--2023 are excluded because backtests
indicated a structural mismatch between the pandemic and energy-crisis
period and the regime being forecast. The target is the hourly German
load in MW. Neither another team's forecast nor the ENTSO-E day-ahead
load forecast enters the model.

\subsubsection{Covariates and publication-time
lags}\label{covariates-and-publication-time-lags}

The autoregressive base consists of weekly lags at 168, 336, 504, and
672 hours and the mean, maximum, minimum, and standard deviation of the
preceding weekly profile. Calendar covariates comprise hour, weekday,
month, their cyclical encodings, weekend status, indicators for
individual public holidays across all 16 German states, days adjacent to
holidays, and bridge days. Astronomical features comprise daylight
duration, a daylight indicator, and, in the final version, the hourly
solar-elevation angle at a fixed German reference point (51° N, 10° E).

Weather is aggregated across Berlin, Hamburg, Munich, Rhine--Ruhr,
Frankfurt, and Stuttgart using fixed population-proxy weights. The
variables are 2 m temperature, cloud cover, 10 m wind speed, shortwave
radiation, and heating- and cooling-degree features. Historical rows use
observed Open-Meteo weather. For the target day, the final pipeline uses
the mean of three numerical weather prediction (NWP) products exposed by
Open-Meteo: ICON Seamless, ECMWF IFS 0.25°, and GFS Seamless. Each
forecast vintage is cached with the run so that a later revision cannot
silently alter the inputs used for that submission.

The most recent ENTSO-E actuals are not always published when a
day-ahead forecast must be made. We therefore apply the same publication
anchor to historical training rows and target rows. Weekly lags are
already settled. For 24- and 48-hour lags, an 800-tree auxiliary
LightGBM nowcaster supplies only those lagged hours lying beyond the
anchor, and published values are retained wherever available. The
nowcaster itself is fitted only on settled rows and uses 48- and 72-hour
raw lags in addition to the base covariates. This construction makes the
short lags available without leaking values that were unpublished at
submission time or creating a train--serve mismatch.

\subsubsection{Development sequence and adoption
rule}\label{development-sequence-and-adoption-rule}

The live entry was a versioned sequence rather than a single immutable
model. Version 7.5 introduced the five-seed subsampled ensemble and
removed the scalar output-bias correction used by the earlier v4.5
fallback. Version 7.9 added the publication-aware short lags, multi-NWP
target weather, and a trailing weekly regime-offset feature. Version
7.9.5 added automatic corrupt-input detection without changing the
forecast when no corruption was flagged. A subsequent holdout showed
that the regime-offset feature amplified the worst rolling seven-day
error, so v7.10 removed it. The final v7.11 retained that offset-free
chassis and added hourly solar elevation. Consequently, the final
leaderboard score evaluates the operational sequence of these versions
and must not be attributed to v7.11 alone.

The live-replica backtest was the team's principal local decision
instrument. At every historical forecast origin, it reconstructed the
information set available at D-1: target-day weather came from archived
forecasts rather than observations, actual load was censored at a
publication anchor, and the production feature, nowcasting, and
training-window rules were rerun. This allowed candidates to be screened
over hundreds of historical forecast origins without deploying every
idea to the live leaderboard and then waiting several days for a stable
rolling score. Because a historical window cannot remove year or regime
drift, candidates selected on one period were checked once on a later,
fresh holdout.

Daily errors cluster into weather and regime episodes, so uncertainty in
paired model differences was estimated by moving-block bootstrap with
seven-day blocks. Adoption required a reduction in paired mean MAE whose
95\% interval excluded zero, together with no regression in the
worst-ten-day mean or the worst rolling seven-day window. This tail gate
rejected changes that looked favourable only on average or over short
windows. The exact released implementations, model cards, and evaluation
tools are available in the
\href{https://github.com/SoheilAyati/Hot-Rod-Public}{Hot Rod public
repository}.

\subsubsection{Compliance with CR-1 to
CR-4}\label{compliance-with-cr-1-to-cr-4}

CR-1 is addressed by packaging each released version as a self-contained
executable submission path and by testing the reusable validation,
scoring, comparison, and backtesting functions. The final challenge
entry was also reproduced independently under the challenge's
peer-reproduction rule. CR-2 is addressed through fixed seeds,
deterministic feature construction, and run-specific snapshots of
external inputs. CR-3 is addressed by freshness, coverage, timestamp,
and missing-value checks that raise explicit errors. In addition, a
conservative guard scans the most recent 14 days for runs of at least
two hours deviating by more than 8 GW from a holiday-aware six-week
profile. Affected target values are excluded from training, affected lag
sources are replaced from an earlier clean same-weekday hour, and every
action is logged. CR-4 is addressed by a LightGBM-only modelling stack
and the dependency restrictions inherited from
\texttt{spotforecast2-safe}. Each run records its publication anchor,
nowcasted-lag counts, weather snapshot, exclusions, model configuration,
and output summary in a structured audit log (PR-4).

\subsection{Team Neura}\label{team-neura}

Team Neura built its forecaster directly on \texttt{spotforecast2-safe},
using the \texttt{ForecasterRecursiveLGBM} model,
\texttt{Period}/\texttt{ExogBuilder} calendar features, and
preprocessing utilities, rather than adopting the defaults of the
reference pipeline of Section~\ref{sec-template} unchanged. In
particular, the team replaced the reference pipeline's single-location
weather representation with its own four-city weather data and used a
fixed 168-hour lag window instead of the reference optimisation-based
configuration.

Hourly actual-load data for Germany were obtained from ENTSO-E and used
as the forecasting target. Incremental caching was implemented to avoid
repeated full downloads, together with a 36-hour freshness check on the
last published actual load. Missing values, detected anomalies, and a
known faulty data period were handled using the
\texttt{spotforecast2-safe} annotation and weighting approach. A
168-hour window was used to zero-weight observations whose lag features
could still be affected by compromised values.

Calendar features were combined with weather data from four German
cities: Frankfurt, Munich, Berlin, and Hamburg. Weather variables
included temperature, wind, and derived heating and cooling degree
features. The multi-city weather information was aggregated to better
represent nationwide German electricity demand.

The final model was a recursive LightGBM forecaster using 168 hourly
lags. LightGBM was configured with 400 trees, a learning rate of 0.05,
63 leaves, and a fixed random seed of 2026. Hyperparameters were set
manually rather than tuned by \texttt{spotoptim} or Optuna.
Deterministic settings were used to improve reproducibility between
runs.

The submitted pipeline satisfies the code-development and process rules
of Section~\ref{sec-code} as follows. CR-1 (no dead code) is satisfied
because the submitted pipeline contains no unused or unreachable code.
All included code paths contribute to data preparation, validation,
feature construction, model training, evaluation, or forecasting. CR-2
(deterministic transformations) is addressed through the fixed random
seed and deterministic LightGBM settings. CR-3 (fail-safe handling) is
addressed through data-coverage, freshness, timestamp-alignment, and NaN
checks that raise explicit errors rather than silently proceeding with
stale, misaligned, or incomplete data, and through the annotation and
zero-weighting of missing, anomalous, and known faulty observations
rather than treating them as reliable training data. CR-4 (minimal
attack surface) is addressed by limiting additional dependencies and
external data access to components required by the forecasting pipeline,
while relying on the dependency controls provided by
\texttt{spotforecast2-safe}. PR-4 (structured audit log) is addressed by
logging key preprocessing, training, and data-quality decisions,
including freshness checks, masked observations, applied weights, and
the random seed.

\subsection{Team Weather Report}\label{sec-team-team-weather-report}

\subsubsection{Covariate selection}\label{sec-covariate-selection}

We kept the reference pipeline of Section~\ref{sec-template} unchanged.
Our main modelling change concerned covariate selection rather than the
introduction of new feature transformations. The underlying library
already provided several optional weather features, but these were
disabled by default and were not part of the reference configuration.
After inspecting the library implementation, we evaluated a bundled
configuration, referred to in this project as \emph{rich}, that
consisted of rolling-window weather aggregates, heating- and
cooling-degree-hour features, apparent temperature and dew point, and
population-weighted weather across 15 German load centres.

We assessed the configuration in a controlled three-date ablation in
which the non-weather covariates, lag selection, backtest configuration,
and optimisation budget were held fixed. Relative to the baseline, the
\emph{rich} configuration reduced average MAE by approximately 29\%.

\subsubsection{Tuning}\label{sec-wr-tuning}

\texttt{spotoptim} returned this same configuration unchanged on all
three dates of the covariate ablation, and we subsequently used it as
the reference for tuning. Within it, \texttt{bagging\_fraction} and
\texttt{reg\_lambda} sat at their lower bounds. That both parameters
landed exactly at their bounds suggested that the search space itself,
rather than the model, might be constraining the result. To test that
suspicion, we widened these bounds (0.5 to 0.3 and 0.001 to 0.0001),
together with a larger budget of \texttt{n\_trials=10} and
\texttt{n\_initial=5}. The resulting configuration achieved an average
MAE of 750 MW, a reduction of about 40\% from the 1258 MW reference
baseline, on the same seven-date backtest, and we used it for subsequent
challenge submissions.

Structurally, it also looked different from the reference: fewer,
shallower trees (\texttt{num\_leaves} fell from 217 to 98 and
\texttt{max\_depth} from 24 to 5), a higher learning rate
(\texttt{learning\_rate} rose from 0.0154 to 0.070), and shifted
regularisation (\texttt{reg\_alpha} from 1.58 to 2.44 and
\texttt{reg\_lambda} from 0.0013 to 0.000193).

\subsubsection{Compliance with CR-1 to
CR-4}\label{compliance-with-cr-1-to-cr-4-1}

Because our submission extends beyond the reference pipeline, more of
the code falls outside \texttt{spotforecast2-safe}'s own guarantees, so
we address CR-1 to CR-4 of Section~\ref{sec-code} partly through that
engine and partly through measures we added directly to the submission
and backtest scripts.

CR-1: We test our functions across the submission and backtest scripts,
including data handling and retry logic for the ENTSO-E download, the
weather cache, and the submission push step.

CR-2: We fixed the reference pipeline's \texttt{random\_state} to 42,
our own choice rather than the library's default of 314159, which also
governs the \texttt{spotoptim} search path. As required of every team in
the challenge, another team independently reproduced our submission's
output exactly on a different machine, confirming determinism.

CR-3: We wrote guards ourselves (\texttt{assert\_no\_leakage},
\texttt{assert\_coverage}, \texttt{assert\_contract}), each raising an
explicit error rather than logging or continuing silently. Our retry
wrapper follows the same rule: a failed ENTSO-E download raises an error
unless \texttt{-\/-allow-stale} is passed explicitly, which falls back
to cached data with a logged warning.

CR-4: We check which packages our two scripts import against the
pipeline's blocklist with a test
(\texttt{test\_importing\_submit\_script\_loads\_no\_blocked\_dependency}),
though it only covers our own files, not the lock file shared across the
whole repository.

\section{Results}\label{sec-results}

This section reports the final outcome of the challenge. Every number,
table, and figure in the manuscript is computed at render time. The
analysis uses the finalised state of 21 July 2026, after the last target
day had been scored and the last ENTSO-E data revisions had been
applied. Every daily score reported below is recomputed from the data
set shipped with this paper, which stores the full leaderboard history.
The computation reproduces the mean MAE and the rank of every entry on
the published leaderboard
(\url{https://bartzbeielstein.github.io/challenge-leaderboard/}). The
live phase comprises the 41 target days from 10 June to 20 July 2026.
Although the challenge formally closed on 22 July 2026, 20 July is the
last scored target day. All metrics are those of
Section~\ref{sec-metrics}, aggregated as means over each entry's scored
days, and ranking follows the challenge rule of ascending mean MAE, with
the number of scored days breaking ties.

\subsection{Leaderboard}\label{sec-leaderboard}

Table~\ref{tbl-leaderboard} lists the 20 leaderboard entries of the live
phase: 11 student teams, eight organiser-operated reference models, and
the ENTSO-E baseline, together with two seasonal-naive benchmarks
computed from the committed ground truth. The scoring rules from
Section~\ref{sec-challenge} influence these numbers. Team Weather Report
had 10 of its 41 days carried forward, so its last place partly reflects
missed submissions rather than model quality, while three further teams
had one carried-forward day each. In addition, seven teams spent their
one-time joker to replace a single already-scored day (Das A Team, Team
Fabinalii, Syntaxerror, Team Neura, DDKAST, Voltrion, and Team Weather
Report, whose joker substituted a fresh submission for its
carried-forward final day). The reference models joined the live phase
on different dates, which is why their day counts differ.

Three observations stand out. First, every entry achieved a lower mean
MAE than the weekly seasonal naive on its own scored days, and the daily
naive trails the entire field by a wide margin, so the challenge
separated genuine forecasting skill from trivial persistence. Second,
the most accurate full-coverage entry was the student team Hot Rod with
a mean MAE of 1227.8 MW, 43.0\% below the ENTSO-E baseline's 2155.6 MW,
while the two reference models ranked above Hot Rod cover only 26 to 29
of the 41 days. Third, the spotoptim-lgbm forecaster introduced as the
TSFML baseline in this report, which appears on the leaderboard under
the entry name spotoptim lgbm, reached rank 6 with a mean MAE of 1369.1
MW over 35 days. It entered the live phase on 16 June, six days after
the restart.

\begin{longtable}[]{@{}
  >{\raggedleft\arraybackslash}p{(\linewidth - 16\tabcolsep) * \real{0.0708}}
  >{\raggedright\arraybackslash}p{(\linewidth - 16\tabcolsep) * \real{0.2389}}
  >{\raggedleft\arraybackslash}p{(\linewidth - 16\tabcolsep) * \real{0.1062}}
  >{\raggedleft\arraybackslash}p{(\linewidth - 16\tabcolsep) * \real{0.1150}}
  >{\raggedleft\arraybackslash}p{(\linewidth - 16\tabcolsep) * \real{0.1062}}
  >{\raggedleft\arraybackslash}p{(\linewidth - 16\tabcolsep) * \real{0.1150}}
  >{\raggedleft\arraybackslash}p{(\linewidth - 16\tabcolsep) * \real{0.0973}}
  >{\raggedleft\arraybackslash}p{(\linewidth - 16\tabcolsep) * \real{0.0708}}
  >{\raggedleft\arraybackslash}p{(\linewidth - 16\tabcolsep) * \real{0.0796}}@{}}

\caption{\label{tbl-leaderboard}Final live-phase leaderboard (41 target
days, 10 June to 20 July 2026), ranked by mean MAE with the number of
scored days breaking ties. Entries marked with an asterisk are
organiser-operated reference models and the remaining ranked entries are
student teams. Days counts scored target days, with carried-forward days
in parentheses. The two seasonal-naive rows are benchmarks computed from
the committed ground truth over all 41 days. They are shown unranked
here and appear as trailing pseudo-entries on the public leaderboard.}

\tabularnewline

\toprule\noalign{}
\begin{minipage}[b]{\linewidth}\raggedleft
Rank
\end{minipage} & \begin{minipage}[b]{\linewidth}\raggedright
Entry
\end{minipage} & \begin{minipage}[b]{\linewidth}\raggedleft
MAE (MW)
\end{minipage} & \begin{minipage}[b]{\linewidth}\raggedleft
RMSE (MW)
\end{minipage} & \begin{minipage}[b]{\linewidth}\raggedleft
MAPE (\%)
\end{minipage} & \begin{minipage}[b]{\linewidth}\raggedleft
Bias (MW)
\end{minipage} & \begin{minipage}[b]{\linewidth}\raggedleft
UPR (\%)
\end{minipage} & \begin{minipage}[b]{\linewidth}\raggedleft
MASE
\end{minipage} & \begin{minipage}[b]{\linewidth}\raggedleft
Days
\end{minipage} \\
\midrule\noalign{}
\endfirsthead

\toprule\noalign{}
\begin{minipage}[b]{\linewidth}\raggedleft
Rank
\end{minipage} & \begin{minipage}[b]{\linewidth}\raggedright
Entry
\end{minipage} & \begin{minipage}[b]{\linewidth}\raggedleft
MAE (MW)
\end{minipage} & \begin{minipage}[b]{\linewidth}\raggedleft
RMSE (MW)
\end{minipage} & \begin{minipage}[b]{\linewidth}\raggedleft
MAPE (\%)
\end{minipage} & \begin{minipage}[b]{\linewidth}\raggedleft
Bias (MW)
\end{minipage} & \begin{minipage}[b]{\linewidth}\raggedleft
UPR (\%)
\end{minipage} & \begin{minipage}[b]{\linewidth}\raggedleft
MASE
\end{minipage} & \begin{minipage}[b]{\linewidth}\raggedleft
Days
\end{minipage} \\
\midrule\noalign{}
\endhead
\bottomrule\noalign{}
\endlastfoot
1 & BABL2L (ENTSO-E)* & 1137.1 & 1403.7 & 2.20 & 54.2 & 48.9 & 0.75 &
29 \\
2 & BABL2L* & 1156.0 & 1399.7 & 2.26 & 339.3 & 41.7 & 0.76 & 26 \\
3 & Hot Rod & 1227.8 & 1452.4 & 2.42 & −90.1 & 51.6 & 0.81 & 41 \\
4 & chronos* & 1345.8 & 1606.8 & 2.61 & −482.6 & 62.0 & 0.89 & 39 \\
5 & optuna lgbm* & 1361.1 & 1615.5 & 2.67 & −253.1 & 55.6 & 0.90 & 39 \\
6 & spotoptim lgbm* & 1369.1 & 1626.0 & 2.70 & −175.8 & 56.0 & 0.91 &
35 \\
7 & Team Neura & 1404.8 & 1631.3 & 2.79 & −378.5 & 60.5 & 0.93 & 41 \\
8 & spotoptim catboost* & 1417.3 & 1690.1 & 2.77 & −166.2 & 54.3 & 0.93
& 27 \\
9 & spotoptim causal* & 1453.5 & 1745.7 & 2.95 & −34.1 & 57.3 & 0.96 &
16 \\
10 & Das A Team & 1459.8 & 1723.0 & 2.86 & −207.3 & 57.6 & 0.96 & 35 \\
11 & spotoptim xgb* & 1526.3 & 1793.7 & 3.03 & 230.1 & 47.5 & 1.01 &
27 \\
12 & Team Gladiators & 1639.1 & 1882.9 & 3.28 & −151.8 & 55.9 & 1.09 &
41 \\
13 & Team Fabinalii & 1672.7 & 1957.2 & 3.33 & −387.5 & 60.8 & 1.11 &
41 \\
14 & Eigen-Squad & 1742.5 & 2009.2 & 3.49 & 584.6 & 42.6 & 1.16 & 41
(1) \\
15 & DDKAST & 1745.2 & 2018.2 & 3.44 & −103.8 & 52.3 & 1.16 & 41 \\
16 & Syntaxerror & 1779.5 & 2096.6 & 3.53 & 347.7 & 49.2 & 1.18 & 41
(1) \\
17 & Voltrion & 1897.6 & 2185.6 & 3.78 & 591.1 & 35.9 & 1.26 & 41 \\
18 & Team ImmerZuSpaet & 1962.4 & 2362.3 & 3.84 & −373.8 & 55.6 & 1.31 &
41 (1) \\
19 & ENTSO-E baseline & 2155.6 & 2434.8 & 4.34 & −38.4 & 51.4 & 1.43 &
41 \\
20 & Team Weather Report & 2271.5 & 2613.4 & 4.52 & 4.8 & 54.2 & 1.51 &
41 (10) \\
-- & Seasonal naive, s = 168 h & 2501.7 & 2823.5 & 4.94 & −503.4 & 58.6
& 1.66 & 41 \\
-- & Seasonal naive, s = 24 h & 3994.8 & 4502.7 & 7.95 & 58.4 & 45.6 &
2.65 & 41 \\

\end{longtable}

Because the mean MAE is averaged over each entry's own scored days,
entries with different coverage are not directly comparable (Hewamalage
et al. 2023). Paired comparisons on shared days give a sharper picture.

\subsection{Paired comparisons on shared days}\label{sec-paired}

Throughout this report, differences between two entries are tested with
two-sided paired t-tests on the daily MAE values of their shared days.
The normality this requires is unproblematic, because each daily MAE is
itself a mean of 24 hourly errors and the paired samples span 26 to 41
days, a setting considered safe for parametric testing.

On the 35 days scored for both, the spotoptim-lgbm forecaster and its
Optuna-tuned sibling were statistically indistinguishable, with mean MAE
1369.1 versus 1371.6 MW (\(p = 0.98\)). Compared with the ENTSO-E
baseline, the spotoptim-lgbm forecaster was more accurate on 26/35
shared days, with mean MAE 1369.1 versus 2097.9 MW, a reduction of
34.7\% (\(p = 0.0001\)). Hot Rod's nominal advantage over the
spotoptim-lgbm forecaster on the shared days (1217.1 versus 1369.1 MW)
does not reach the 5\% level (\(p = 0.13\)), so the final ranking gap
between the two is not statistically resolved at this sample size.

\subsection{Statistical significance of the
differences}\label{sec-significance}

Whether the observed differences are statistically meaningful was tested
with a repeated-measures analysis of variance (ANOVA) on the daily MAE
matrix, blocked by target day, followed by pairwise two-sided paired
t-tests with Holm correction, the parametric variant of the comparison
protocol of Hewamalage et al. (2023). The panel comprises all eleven
student teams together with the ENTSO-E baseline and the two seasonal
naives. Two selections keep the comparison manageable. The panel is
restricted to the 35 target days scored for every one of its entries,
because Das A Team submitted its first forecast on 16 June, which drops
the six opening days of the live phase. The organiser-operated reference
models are excluded, so the panel compares the student field with the
ENTSO-E baseline and the naive benchmarks. The ANOVA rejects equality of
the fourteen entries decisively (\(F(13, 442) = 9.21\),
\(p < 10^{-16}\)).

\begin{figure}[H]

\centering{

\pandocbounded{\includegraphics[keepaspectratio]{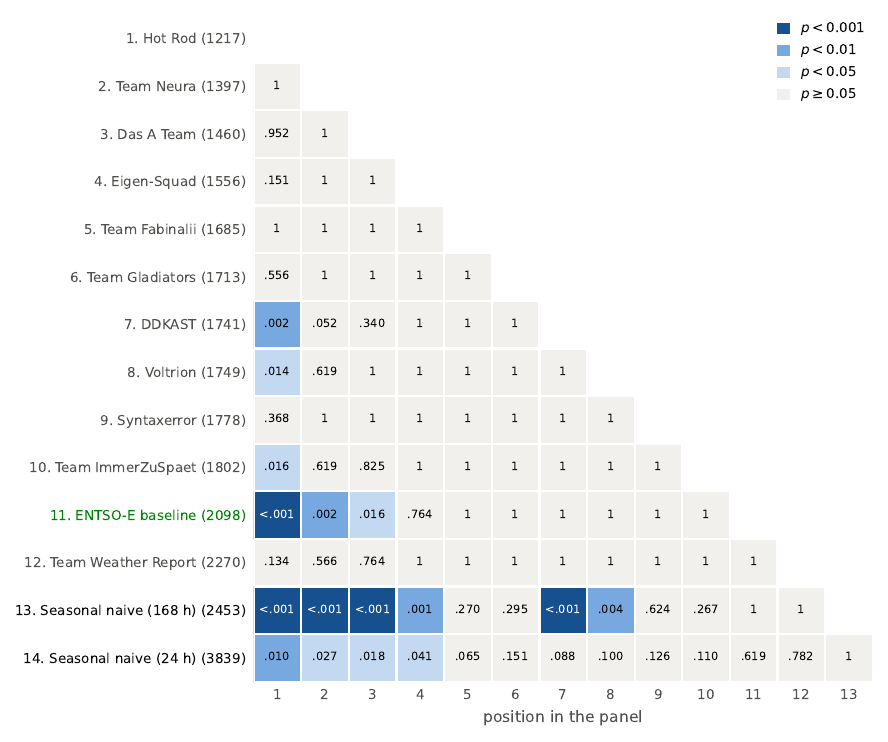}}

}

\caption{\label{fig-significance-matrix}Student teams only.
Holm-adjusted p-values of all 91 pairwise two-sided paired t-tests
behind Figure~\ref{fig-cd-diagram}, over the same 35 target days.
Entries are ordered by mean daily MAE and numbered accordingly, so each
cell compares the entry naming its row with the entry whose number
labels its column. Darker cells indicate stronger evidence of a
difference. Values below 0.001 are shown as \textless.001.}

\end{figure}

Figure~\ref{fig-significance-matrix}, which reports 91 Holm-adjusted
p-values, reveals three findings. Hot Rod, Team Neura, and Das A Team
are significantly more accurate than the ENTSO-E baseline (Holm-adjusted
\(p < 10^{-4}\), \(p = 0.0017\), and \(p = 0.016\)), and no other
student team separates from it after correction (adjusted
\(p \ge 0.76\)). The ENTSO-E baseline itself is statistically
indistinguishable from the weekly seasonal naive at the day level over
this period (\(p = 1.0\) after correction), and even from the daily
naive (adjusted \(p = 0.62\)), although the latter's mean daily MAE lies
\(1,741\) MW higher, because the daily naive's error fluctuates between
\(678\) and \(11,501\) MW across the panel days. Only the four leading
student teams separate from the daily naive (adjusted \(p \le 0.04\)).
The leading student teams are largely mutually indistinguishable, with
adjusted \(p \ge 0.15\) between Hot Rod and the five teams that follow
it. Hot Rod does separate from the mid-field trio of DDKAST, Voltrion,
and Team ImmerZuSpaet (adjusted \(p \le 0.016\)) but not from
last-placed Team Weather Report (\(p = 0.13\)), whose ten
carried-forward days inflate its day-to-day variance. Thirty-five shared
days thus separate the ends of the field but not its neighbours, the
sample-size effect described by Hewamalage et al. (2023).

\subsection{Sensitivity of the panel to the coverage
requirement}\label{sec-panel-sensitivity}

\begin{figure}[H]

\centering{

\pandocbounded{\includegraphics[keepaspectratio]{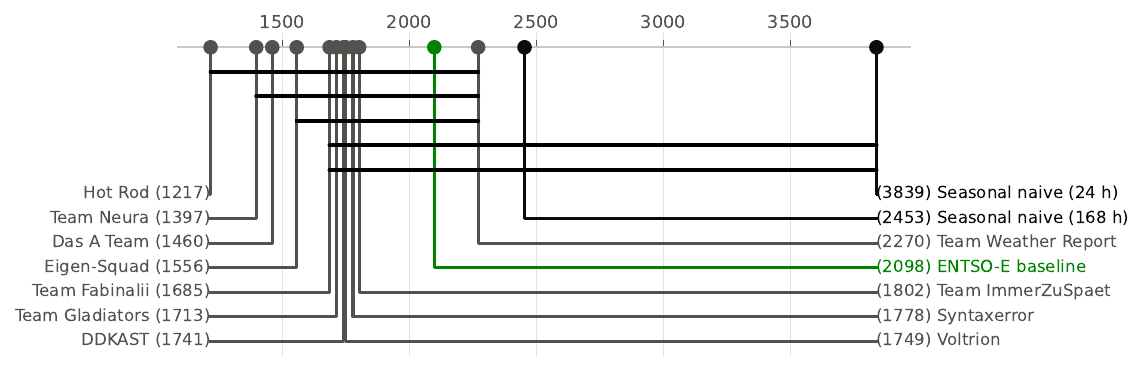}}

}

\caption{\label{fig-cd-diagram}Mean daily MAE (MW) of the fourteen panel
entries (the eleven student teams, the ENTSO-E baseline, and the two
seasonal naives) over the 35 live days scored for all of them, in the
diagram layout of Demšar (2006). Lower is better. A crossbar marks a
maximal set of entries that are mutually indistinguishable under
two-sided paired t-tests with Holm correction at the five percent level.
Because each bar is drawn as a plain span between its two outermost
members, it may also cover entries that are not part of the set.
Figure~\ref{fig-significance-matrix} resolves those cases.}

\end{figure}

There is a trade-off between (i) how many entries the panel compares and
(ii) how many days it compares them over. The same set of entries as in
Figure~\ref{fig-significance-matrix} is used to generate a
critical-difference diagram in Figure~\ref{fig-cd-diagram}, which
visualises the pairwise comparisons of the 35-day panel. The crossbars
mark maximal sets of mutually indistinguishable entries under
Holm-corrected two-sided paired t-tests at the five percent level. The
topmost bar spans Hot Rod to Team Weather Report, which are mutually
indistinguishable (\(p = 0.13\)), but it also passes over the ENTSO-E
baseline, from which Hot Rod does separate (\(p < 10^{-4}\)).

\begin{figure}[H]

\centering{

\pandocbounded{\includegraphics[keepaspectratio]{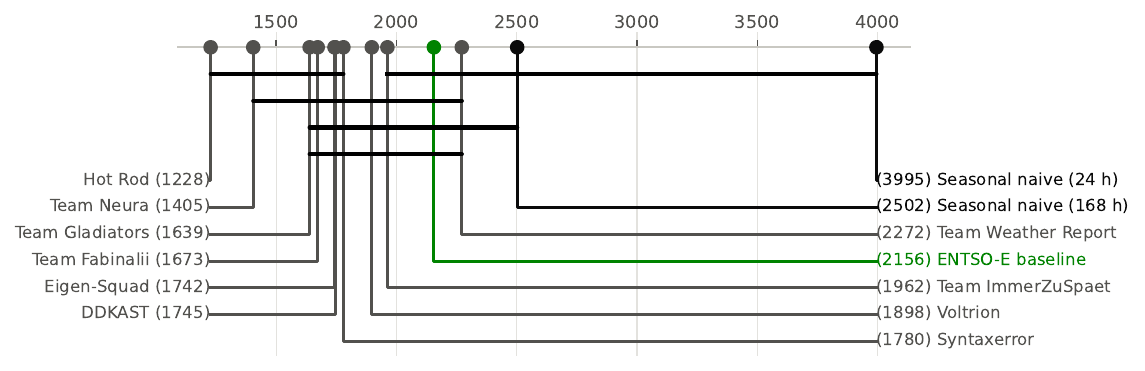}}

}

\caption{\label{fig-cd-40day}Critical-difference diagram over the
entries scored on at least 40 of the 41 target days, which admits
thirteen entries over the full 41-day window. Das A Team is absent
because its first submission was on 16 June. Layout and crossbar
semantics follow Figure~\ref{fig-cd-diagram}.}

\end{figure}

Admitting only entries scored on at least 40 of the 41 target days, as
in Figure~\ref{fig-cd-40day}, drops Das A Team and restores the full
41-day window for the remaining 13 entries (\(F(12, 480) = 10.26\),
\(p < 10^{-17}\)).

Only Hot Rod and Team Neura separate from the ENTSO-E baseline over the
full 41-day window (adjusted \(p < 10^{-4}\) and \(p = 0.0026\)),
because Das A Team, the third team that separates from it in
Figure~\ref{fig-cd-diagram}, is no longer in the panel.

A comparison of Figure~\ref{fig-cd-diagram} with
Figure~\ref{fig-cd-40day} gives further insight. Of their 78 common
pairs, 9 cross the five percent threshold, in both directions: Hot Rod
and Team Weather Report separate over 41 days (\(p = 0.034\)) but not
over 35 (\(p = 0.13\)), whereas Eigen-Squad and the weekly naive
separate over 35 days (\(p = 0.0012\)) but not over 41 (\(p = 0.30\)).
Five of the 9 crossings involve the daily naive, whose volatile
day-level comparisons sit near the threshold throughout.
Figure~\ref{fig-significance-matrix} avoids some of the ambiguities of
the critical-difference diagrams, because it shows the full matrix of
pairwise comparisons with their Holm-adjusted p-values.

\subsection{Rank stability across the metrics}\label{sec-rank-stability}

\begin{figure}[H]

\centering{

\pandocbounded{\includegraphics[keepaspectratio]{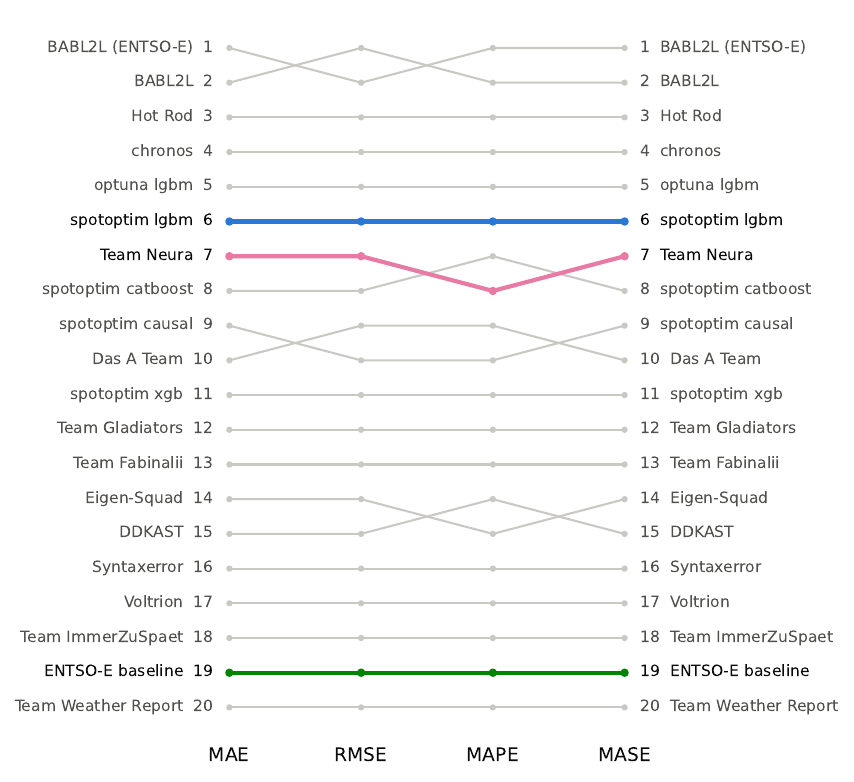}}

}

\caption{\label{fig-rank-stability}Leaderboard rank of every entry under
the four error metrics, each aggregated as the mean of daily values.
Grey lines are entries whose rank changes by at most one position. The
highlighted entries are the spotoptim-lgbm forecaster (blue), the
ENTSO-E baseline (green), and Team Neura (magenta).}

\end{figure}

Although MAE alone determines the official ranking, it is of interest
how the leaderboard changes under alternative error metrics.
Figure~\ref{fig-rank-stability} re-ranks all entries by mean RMSE, MAPE,
and MASE. Kendall's \(\tau\) between the MAE ranking and the three
alternatives is 0.979, 0.968, and 1.000, and every movement in the
figure is a swap of adjacent or near-adjacent entries.

Under RMSE, which penalises large hourly errors, BABL2L overtakes its
ENTSO-E-informed variant by a 3 MW margin, and Das A Team edges past
spotoptim causal. MAPE re-weights each day by the inverse of its demand
level, which penalises entries whose errors fall on low-demand days, and
Team Neura drops from rank 7 to 8. The MASE of Equation~\ref{eq-mase} is
an almost affine copy of the MAE, because its in-sample scaling factor
varies little across target days (1451 to 1530 MW): 10 of the 20 entries
stay below one and thus beat the average one-step naive error. The
ranking is therefore robust to the choice of metric: for a single hourly
series far from zero, the scale-dependent MAE is an appropriate primary
metric, and the alternatives change little.

\subsection{Pretrained in-context models}\label{sec-foundation}

Three reference entries in Table~\ref{tbl-leaderboard} are non-TSFML
models: they perform no gradient training at submission time, in
contrast to the daily-refit recursive gradient-boosting pipelines of
Section~\ref{sec-recursive}.

The two BABL2L models share one submitter and one model family. At
submission time the model conditions in-context on the most recent 540
to 600 day-rows without updating its weights. The two identities differ
in exactly one input. BABL2L (ENTSO-E) receives the ENTSO-E baseline as
a fifth covariate, and this single channel enters three ways: as a
feature column of the in-context forecaster, through the same feature
rows into a ridge regression whose output is averaged with the model
output, and as the third voter of a median safety guard. Plain BABL2L
excludes that channel everywhere and retains the renewable-generation
forecasts and the day-ahead price. Both identities blend, clip, and
spike-repair the raw model output.

The chronos entry is the time-series foundation model Chronos-2, applied
zero-shot and univariate (Ansari et al. 2024, 2025): the actual load
series is its only input, with no covariates, no calendar features, and
no training or tuning. From a context of the trailing 180 days it emits
the 0.1, 0.5, and 0.9 quantiles of the next 24 hours in one direct
multi-step pass, and the submission is the 0.5 quantile, the MAE-optimal
point summary under the challenge's ranking metric. The ENTSO-E baseline
serves only a warn-only plausibility check and is never a model input.

Table~\ref{tbl-foundation} compares the BABL2L- and chronos-based
results on shared scored days. They are absent from the complete-panel
test of Section~\ref{sec-significance} because they joined the live
phase late. Two results stand out. First, the two BABL2L identities are
statistically indistinguishable on their 26 shared days (mean MAE 1156.0
versus 1171.3 MW, \(p = 0.80\)), with the plain variant nominally ahead.
The ENTSO-E channel bought no measurable accuracy, and the first place
of BABL2L (ENTSO-E) over its sibling in Table~\ref{tbl-leaderboard} is a
coverage artefact of three additional scored days, the same pattern as
the spotoptim and Optuna comparison of Section~\ref{sec-paired}. Second,
the in-context entries match or beat the locally trained recursive
forecasters. BABL2L (ENTSO-E) is significantly more accurate than
spotoptim lgbm on 29 shared days (\(p = 0.019\)), and plain BABL2L than
spotoptim xgb (\(p = 0.025\)). Zero-shot chronos is statistically tied
with the daily-retuned optuna lgbm over 39 shared days (1345.8 versus
1361.1 MW, \(p = 0.90\)), a remarkable result for a univariate model
without tuning. Between the two in-context families, BABL2L leads
chronos on their 26 shared days (1156.0 versus 1283.1 MW, \(p = 0.22\)),
a nominal but not significant margin at this sample size. Whether
pretrained in-context models displace tuned gradient-boosting pipelines
in this setting is taken up in Section~\ref{sec-discussion}.

\begin{longtable}[]{@{}
  >{\raggedright\arraybackslash}p{(\linewidth - 10\tabcolsep) * \real{0.3956}}
  >{\raggedleft\arraybackslash}p{(\linewidth - 10\tabcolsep) * \real{0.0879}}
  >{\raggedleft\arraybackslash}p{(\linewidth - 10\tabcolsep) * \real{0.1538}}
  >{\raggedleft\arraybackslash}p{(\linewidth - 10\tabcolsep) * \real{0.1538}}
  >{\raggedleft\arraybackslash}p{(\linewidth - 10\tabcolsep) * \real{0.1319}}
  >{\raggedleft\arraybackslash}p{(\linewidth - 10\tabcolsep) * \real{0.0769}}@{}}

\caption{\label{tbl-foundation}Paired comparisons of the pretrained
in-context entries on shared scored days: mean MAE of each side over
exactly those days, the number of days on which entry A was more
accurate, and the two-sided paired t-test p-value. Entries are compared
only on days scored for both.}

\tabularnewline

\toprule\noalign{}
\begin{minipage}[b]{\linewidth}\raggedright
Comparison (A vs B)
\end{minipage} & \begin{minipage}[b]{\linewidth}\raggedleft
Days
\end{minipage} & \begin{minipage}[b]{\linewidth}\raggedleft
MAE A (MW)
\end{minipage} & \begin{minipage}[b]{\linewidth}\raggedleft
MAE B (MW)
\end{minipage} & \begin{minipage}[b]{\linewidth}\raggedleft
A better
\end{minipage} & \begin{minipage}[b]{\linewidth}\raggedleft
\(p\)
\end{minipage} \\
\midrule\noalign{}
\endfirsthead

\toprule\noalign{}
\begin{minipage}[b]{\linewidth}\raggedright
Comparison (A vs B)
\end{minipage} & \begin{minipage}[b]{\linewidth}\raggedleft
Days
\end{minipage} & \begin{minipage}[b]{\linewidth}\raggedleft
MAE A (MW)
\end{minipage} & \begin{minipage}[b]{\linewidth}\raggedleft
MAE B (MW)
\end{minipage} & \begin{minipage}[b]{\linewidth}\raggedleft
A better
\end{minipage} & \begin{minipage}[b]{\linewidth}\raggedleft
\(p\)
\end{minipage} \\
\midrule\noalign{}
\endhead
\bottomrule\noalign{}
\endlastfoot
BABL2L vs BABL2L (ENTSO-E) & 26 & 1156.0 & 1171.3 & 13/26 & 0.804 \\
BABL2L vs chronos & 26 & 1156.0 & 1283.1 & 15/26 & 0.218 \\
BABL2L (ENTSO-E) vs spotoptim lgbm & 29 & 1137.1 & 1391.9 & 19/29 &
0.019 \\
BABL2L vs spotoptim lgbm & 26 & 1156.0 & 1377.3 & 17/26 & 0.066 \\
chronos vs spotoptim lgbm & 35 & 1250.4 & 1369.1 & 17/35 & 0.238 \\
chronos vs optuna lgbm & 39 & 1345.8 & 1361.1 & 20/39 & 0.901 \\
BABL2L vs spotoptim xgb & 26 & 1156.0 & 1539.5 & 18/26 & 0.025 \\

\end{longtable}

\subsection{Error diagnostics}\label{sec-error-diagnostics}

\begin{figure}[H]

\centering{

\pandocbounded{\includegraphics[keepaspectratio]{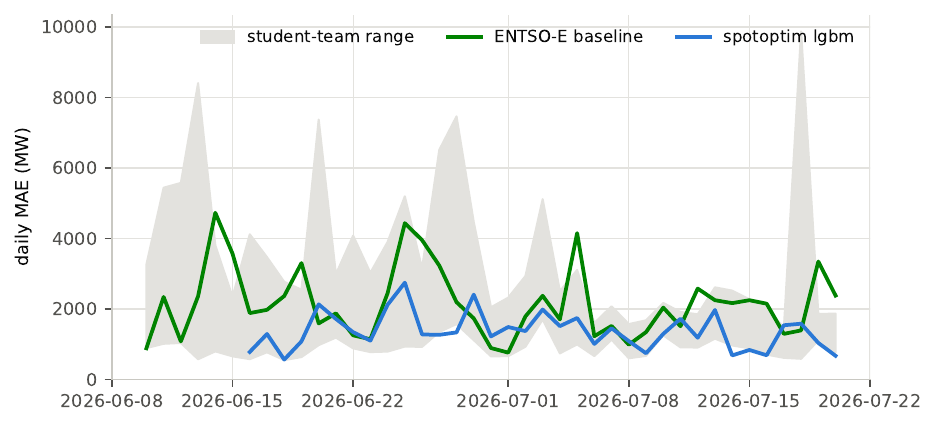}}

}

\caption{\label{fig-mae-over-time}Daily MAE over the live phase. The
grey band spans the minimum and maximum daily MAE of the student teams
with complete coverage. The spotoptim-lgbm forecaster entered the live
phase on 16 June.}

\end{figure}

Figure~\ref{fig-mae-over-time} traces the daily MAE of the
spotoptim-lgbm forecaster and the ENTSO-E baseline across the live
phase, against the band spanned by the ten full-coverage student teams.
The ENTSO-E baseline's opening week stands out: from 10 to 15 June it
over-forecast the load by 2264 MW on average. The spotoptim-lgbm
forecaster's trace begins on 16 June, so this episode lies outside its
scored period.

\begin{figure}[H]

\centering{

\pandocbounded{\includegraphics[keepaspectratio]{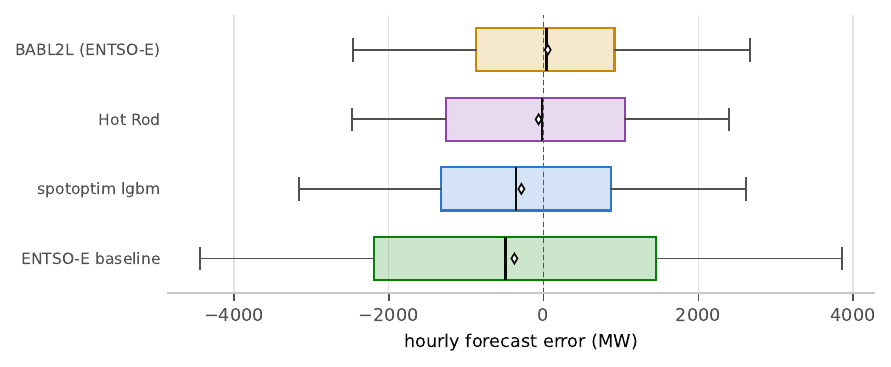}}

}

\caption{\label{fig-error-box}Distribution of the hourly forecast errors
(forecast minus actual) of the four entries of
Table~\ref{tbl-error-stats}, over the 696 hours scored for all of them.
The box spans the interquartile range with the median as a solid rule,
the whiskers reach the 5\% and 95\% quantiles reported in the table, and
the open diamond marks the mean. Points beyond the whiskers are omitted.
Extremes are shown in the table. The dashed line marks a zero error,
i.e., a box lying left of it indicates an under-forecast.}

\end{figure}

Table~\ref{tbl-error-stats} and Figure~\ref{fig-error-box} summarise the
distribution of the hourly errors of four entries over the 696 hours of
the 29 target days scored for all of them: BABL2L (ENTSO-E) as the most
accurate entry overall, Hot Rod as the most accurate student team, the
spotoptim-lgbm forecaster, and the ENTSO-E baseline. The window is
shorter than the 840-hour sample used below, because BABL2L (ENTSO-E)
entered the live phase on 22 June.

Three features are of interest. First, the standard deviation ranks the
four entries exactly as the mean MAE does, and the gap is dominated by
the ENTSO-E baseline, whose standard deviation of 2503 MW is 70\% above
the 1476 MW of BABL2L (ENTSO-E) and whose central 90\% range spans 8307
MW against 5142 MW. Second, the three model entries are close to
unbiased on this window, with mean errors between −286 and 54 MW,
whereas the ENTSO-E baseline sits at −377 MW with a median of −493 MW,
so it under-forecasts in the majority of hours once the opening week is
excluded. The third feature is a caution rather than a result. The two
leading entries change places depending on which part of the
distribution is read: Hot Rod has the narrower central 90\% range (4886
against 5142 MW), while BABL2L (ENTSO-E) has the tighter interquartile
box (1785 against 2309 MW) and the shorter lower tail (−4452 against
−5178 MW).

The bias depends on the sample. Over the full 41 days the ENTSO-E
baseline's mean bias is \(-38.4\) MW, the near-zero value discussed in
Section~\ref{sec-data}, but over the 35 days it shares with the
spotoptim-lgbm forecaster it is \(-433\) MW. The difference is the
opening-week over-forecast episode.

\begin{longtable}[]{@{}
  >{\raggedright\arraybackslash}p{(\linewidth - 8\tabcolsep) * \real{0.2247}}
  >{\raggedleft\arraybackslash}p{(\linewidth - 8\tabcolsep) * \real{0.2247}}
  >{\raggedleft\arraybackslash}p{(\linewidth - 8\tabcolsep) * \real{0.1236}}
  >{\raggedleft\arraybackslash}p{(\linewidth - 8\tabcolsep) * \real{0.2022}}
  >{\raggedleft\arraybackslash}p{(\linewidth - 8\tabcolsep) * \real{0.2247}}@{}}

\caption{\label{tbl-error-stats}Descriptive statistics of the hourly
forecast errors (forecast minus actual, in MW) over the 696 hours of the
29 target days scored for all four entries. Negative values are
under-forecasts. Columns run from the lowest to the highest mean
absolute error on this window.}

\tabularnewline

\toprule\noalign{}
\begin{minipage}[b]{\linewidth}\raggedright
Statistic
\end{minipage} & \begin{minipage}[b]{\linewidth}\raggedleft
BABL2L (ENTSO-E)
\end{minipage} & \begin{minipage}[b]{\linewidth}\raggedleft
Hot Rod
\end{minipage} & \begin{minipage}[b]{\linewidth}\raggedleft
spotoptim lgbm
\end{minipage} & \begin{minipage}[b]{\linewidth}\raggedleft
ENTSO-E baseline
\end{minipage} \\
\midrule\noalign{}
\endfirsthead

\toprule\noalign{}
\begin{minipage}[b]{\linewidth}\raggedright
Statistic
\end{minipage} & \begin{minipage}[b]{\linewidth}\raggedleft
BABL2L (ENTSO-E)
\end{minipage} & \begin{minipage}[b]{\linewidth}\raggedleft
Hot Rod
\end{minipage} & \begin{minipage}[b]{\linewidth}\raggedleft
spotoptim lgbm
\end{minipage} & \begin{minipage}[b]{\linewidth}\raggedleft
ENTSO-E baseline
\end{minipage} \\
\midrule\noalign{}
\endhead
\bottomrule\noalign{}
\endlastfoot
Mean & 54.2 & −64.5 & −286.5 & −376.9 \\
Median & 37.3 & −19.0 & −358.6 & −493.4 \\
5\% quantile & −2462.8 & −2477.6 & −3171.1 & −4442.7 \\
95\% quantile & 2679.1 & 2408.7 & 2619.9 & 3864.3 \\
Standard deviation & 1475.5 & 1537.4 & 1729.5 & 2503.2 \\
Minimum & −4451.7 & −5177.7 & −6828.3 & −6562.5 \\
Maximum & 4216.6 & 3949.3 & 3876.5 & 6572.2 \\

\end{longtable}

\begin{figure}[H]

\centering{

\pandocbounded{\includegraphics[keepaspectratio]{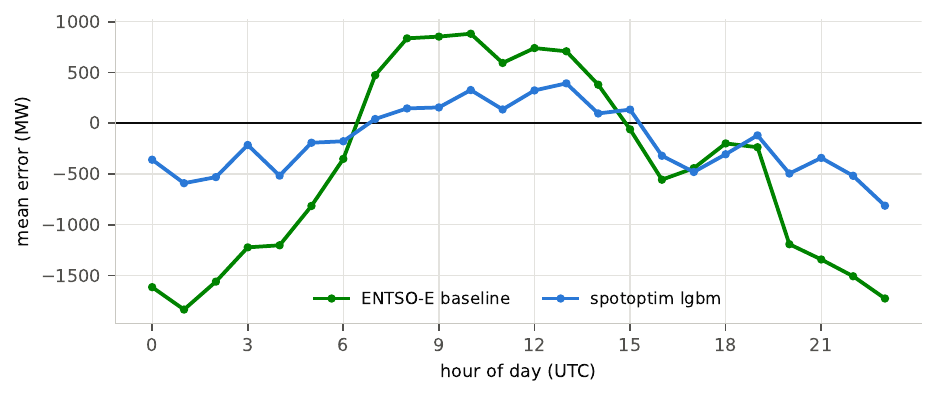}}

}

\caption{\label{fig-error-profile}Mean hourly forecast error by hour of
day (UTC) over the 35-day joint sample. Negative values are
under-forecasts.}

\end{figure}

The hour-of-day structure in Figure~\ref{fig-error-profile} mirrors the
pattern Möbius et al. (2025) report for 2016 to 2019: the ENTSO-E
baseline under-forecasts the night hours by 0.8 to 1.8 GW and
over-forecasts the morning ramp by up to 0.9 GW. For comparison, the
spotoptim-lgbm forecaster's mean error stays between \(-812\) and
\(+393\) MW at every hour. The night-time under-prediction is the
component a load-serving operator cares most about, and it is where the
margin of the spotoptim-lgbm forecaster over the ENTSO-E baseline is
widest.

\begin{figure}[H]

\centering{

\pandocbounded{\includegraphics[keepaspectratio]{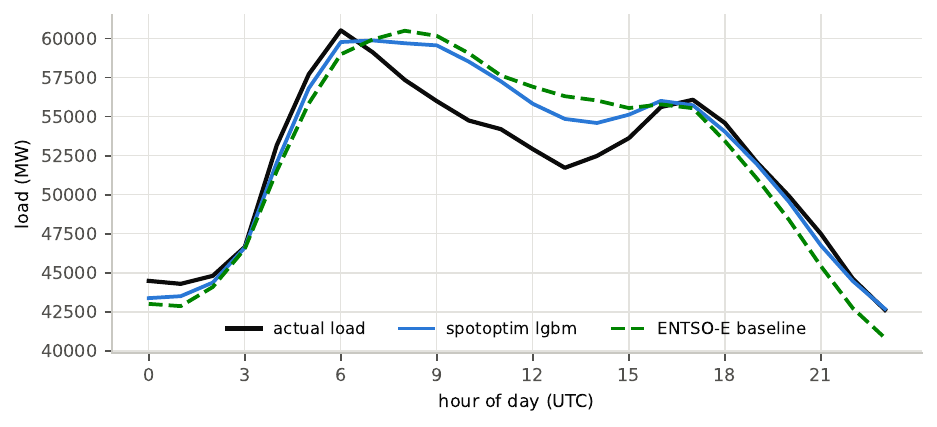}}

}

\caption{\label{fig-forecast-example}Day-ahead forecasts and realised
load on the example day, selected as the day whose daily MAE is closest
to the spotoptim-lgbm forecaster's median. Timestamps in UTC.}

\end{figure}

Figure~\ref{fig-forecast-example} shows one complete day-ahead forecast
against the realised load. To avoid judging by a favourable case
(Hewamalage et al. 2023), the day is chosen by a fixed rule: the target
day whose daily MAE lies closest to the spotoptim-lgbm forecaster's
median daily MAE of 1292 MW. That rule selects Friday 10 July 2026, with
an MAE of 1292 MW. The forecast tracks the night hours and the morning
ramp closely, overestimates the midday dip like the ENTSO-E baseline but
by a smaller margin, and re-converges on the actual load over the
evening decline, where the ENTSO-E baseline under-forecasts.

\section{Discussion}\label{sec-discussion}

\subsection{Reference pipeline}\label{reference-pipeline}

The TSFML-based spotoptim-lgbm forecaster reduced mean MAE against the
ENTSO-E baseline by 34.7\% on shared scored days. Error diagnostics show
that this improvement does not come from correcting a constant offset.
Over the completed live phase, the ENTSO-E baseline's mean bias of
\(-38.4\) MW is nearly two orders of magnitude smaller than its mean
absolute error, so subtracting the average error would change little.
The improvement comes from correcting the shape of the error.

The ENTSO-E baseline's mean error fluctuates from a night-time
under-forecast of up to 1.8 GW to a morning over-forecast of 0.9 GW
(Figure~\ref{fig-error-profile}), while the spotoptim-lgbm forecaster
holds every hour of the day inside the interval from \(-812\) to
\(+393\) MW and reduces the overall error distribution by roughly a
third (Table~\ref{tbl-error-stats}). A weaker version of the night-time
under-prediction reported by Möbius et al. (2025) for 2016 to 2019 is
therefore still present in the 2026 ENTSO-E baseline.

Although the spotoptim-lgbm forecaster corrects most of the remaining
structure, the sign of its residual bias matters from an operational
point of view. Its full-phase mean bias is \(-176\) MW with an
under-prediction rate of 56.0\%, a minor but constant under-forecast
that should be discussed with practitioners. By challenge design,
forecasters are tuned only on MAE and have no incentive to remove the
bias.

\subsection{Hyperparameter tuning}\label{hyperparameter-tuning}

The live-phase results give a clear answer about the tuners: neither was
superior. The spotoptim-lgbm forecaster and its Optuna-tuned sibling
were statistically indistinguishable on their 35 shared days
(\(p = 0.98\)).

The corresponding result for the two BABL2L variants (\(p = 0.80\))
suggests a common explanation. Within a daily tuning budget and on a
well-conditioned search space, both surrogate-model search and the
tree-structured Parzen estimator of Akiba et al. (2019) reach the same
accuracy plateau. The remaining day-to-day difference appears to be
driven more by the data than by the optimiser. How much of the final
accuracy depends on the covariate set rather than on the tuning budget
cannot be isolated under live conditions and remains unanswered.

\subsection{Data engineering}\label{data-engineering}

The value of the data-engineering layers is documented more indirectly.
The enriched calendar set entered production only after a paired-seed
benchmark showed a consistent cross-validated improvement
(Section~\ref{sec-dataset}). The wind covariates were dropped when their
archive coverage proved unreliable and then kept out because a second
paired-seed benchmark showed that restoring them after coverage
recovered was worse than the wind-free configuration
(Section~\ref{sec-dataset}).

\subsection{In-context models}\label{in-context-models}

The final board is topped by in-context models, namely BABL2L and
Chronos-2. Section~\ref{sec-foundation} shows that this result is not
only a coverage artefact. BABL2L (ENTSO-E) beat the recursive LightGBM
significantly on shared days (\(p = 0.019\)), and zero-shot Chronos-2
matched the daily-retuned Optuna variant (\(p = 0.90\)). Together with
the evidence in Hollmann et al. (2025) and Ansari et al. (2025), these
results mark in-context regression as a genuine challenger to
daily-refit gradient boosting on this task.

\subsection{Auditability and regulatory
framing}\label{auditability-and-regulatory-framing}

From the audit perspective of Section~\ref{sec-intro}, TSFML and
non-TSFML approaches are not interchangeable. The TSFML forecasters are
trained from scratch each day from committed data by a reviewed
deterministic engine, whereas Chronos-2 imports weights whose training
data and training procedure lie outside the operator's audit boundary.
Where auditability is a requirement, this gap influences the model
choice.

For the regulatory framing of Section~\ref{sec-intro}, the challenge
functioned as a rehearsal of the record-keeping obligations the AIA
(European Parliament and Council of the European Union 2024) attaches to
high-risk systems, although it is applied voluntarily to a system that
sits outside the safety perimeter. Determinism and reproducibility were
engineering constraints from the start, enforced by the reviewed subset
of the TSFML pipeline (Section~\ref{sec-recursive}) and the explicit
annotation of outliers, anomalies, and missing values
(Section~\ref{sec-outliers}). Auditability extends to the paper itself.
Results are computed at render time from a committed snapshot, and the
build aborts if they stop reproducing the published leaderboard. An
external reviewer can rerun the pipeline, rescore any day, and rederive
every number in Section~\ref{sec-results} from versioned artefacts.

\subsection{Team findings}\label{sec-team-findings}

The four subsections below are the teams' own readings of their results,
in the alphabetical order of Section~\ref{sec-teams}. Each discusses the
entry whose construction the corresponding methods subsection describes,
so the two can be read as a pair. The preceding paragraphs of this
section, and the field-wide comparisons of Section~\ref{sec-results}
they rest on, are computed once for the whole leaderboard and are not
restated per team.

\subsubsection{Team Fabinalii}\label{sec-findings-team-fabinalii}

Our final MAE of 1,672.7 MW across 41 scored days was 22\% below the
ENTSO-E baseline, although Section~\ref{sec-significance} shows that
this gap is not statistically resolved. That average hides two major
issues. Until late June, the main problem was over-forecasting on
Saturdays, especially on 20 June, when the MAE reached 7,371 MW. The
best entries scored between 926 and 1,100 MW on that day, so the day
itself was not an outlier. Our model was failing. That comparison, more
than our internal metrics, forced a redesign. The weekday-dependent lag
set worked: the morning ramp mismatch from Friday to Saturday, between
10,000 and 14,000 MW, matched our error window closely, and the weekday
backtests showed gains on Monday, Saturday, and Sunday. The longer
training window also helped, improving results by about 20\% relative to
the 90-day setting. Feature engineering on top of the unchanged core did
not help. Additional lags and a weekend interaction term made the model
unstable, bias correction revealed no reliable pattern, and a direct
multi-step strategy was consistently two to three times worse. Two
issues remain for future work. Our remaining bias of \(-387.5\) MW (UPR
60.8\%) is an under-forecast in the same direction as the ENTSO-E
baseline error. Public holidays also show the mismatch observed on
Saturdays, so they should be treated like weekends in the lag selection.

\subsubsection{Team Hot Rod}\label{sec-findings-team-hot-rod}

Hot Rod finished with a mean MAE of 1,227.8 MW over all 41 scored days,
the most accurate student entry and the most accurate entry with
complete coverage. It ranked third overall because the two
organiser-operated entries above it covered only 29 and 26 days. Its MAE
was 43.0\% below the ENTSO-E baseline's 2,155.6 MW. Its RMSE was 1,452.4
MW, MAPE 2.42\%, bias \(-90.1\) MW, under-prediction rate 51.6\%, and
MASE 0.81 (Section~\ref{sec-leaderboard}). In the full 41-day panel of
entries with at least 40 scored days, Hot Rod separates from the ENTSO-E
baseline after Holm correction (\(p<10^{-4}\),
Section~\ref{sec-panel-sensitivity}). This result supports the
operational pipeline as a whole. It does not show that Hot Rod is
intrinsically better than the neighbouring student methods: on the 35
days shared by all student entries, the differences between Hot Rod and
the five teams immediately following it are not statistically resolved
(adjusted \(p\geq0.15\), Section~\ref{sec-significance}).

Backtest and live score agree closely. The 364-day live-replica backtest
estimated a mean MAE of 1,220.2 MW for v7.9. The final 41-day
operational score was 1,227.8 MW, only 7.6 MW (0.6\%) higher. This
agreement is important operationally: the local harness provided a
useful performance scale before the live outcome was known and let the
team test candidate changes across a full historical year instead of
waiting days for each leaderboard result. It should not be read as an
unbiased point prediction for v7.11, because the live score combines
several deployed versions and a shorter summer evaluation period.

Our internal backtests explain why the model sequence changed, but they
should not be confused with the live leaderboard. On a 364-day
live-replica window from 1 July 2024 to 30 June 2025, v7.5 reduced mean
MAE from 1,633 MW for deployed v4.5 to 1,331 MW. The paired improvement
was 302 MW with a 95\% moving-block-bootstrap interval of {[}207, 407{]}
MW. The worst-ten-day mean fell from 5,572 to 4,153 MW, and the worst
rolling seven-day value from 3,830 to 2,621 MW. That comparison also
reversed our view of scalar bias correction. With otherwise identical
v4.5 forecasts, enabling the correction increased mean MAE by 85 MW and
the worst-day MAE by 1,052 MW, although it improved the worst rolling
seven-day window. A correction estimated from recent residuals was
therefore too volatile to retain.

On the same full-year design, v7.9 improved mean MAE by a further 110.9
MW relative to v7.5, with a 95\% block-bootstrap interval of {[}63,
168{]} MW. Its worst-ten-day mean improved from 4,153 to 3,842 MW and
its worst rolling seven-day value from 2,620.9 to 2,305.5 MW, while the
single worst day was essentially unchanged (5,095 versus 5,078 MW). The
component study was important: the publication-aware short lags improved
the mean but failed the tail gate alone, and pairs of the new components
still worsened the worst rolling window. The combination of short-lag
nowcasting, multi-NWP weather, and the regime offset passed. This is
evidence for the tested combination, not proof that every component is
beneficial in every period.

That qualification mattered on the next holdout. On a fresh 335-day
live-replica period from 1 July 2025 to 31 May 2026, evaluated with one
seed and a 24-hour publication lag, the trailing regime offset increased
mean MAE to 1,282.9 MW from 1,249.0 MW for the otherwise equivalent
offset-free model. It also increased the worst-ten-day mean from 3,449
to 3,769 MW and the worst rolling seven-day MAE from 2,195.8 to 2,399.4
MW. We therefore removed it in v7.10. Because this check used one seed,
it was treated as a directional tail gate rather than a final precision
estimate. The result is a practical warning against treating recency
correction as universally stabilising: even a settled-only feature can
amplify a persistent error when the load regime changes. The same
caution applies to feature attribution. Weekly lags dominate the model's
TreeSHAP values, but influence inside a fitted model is not evidence of
out-of-sample benefit and was not used as an adoption criterion.

The final solar-elevation change has deliberately weaker evidence. In a
165-day, one-seed backtest on the weather-scenario variant, it improved
daily MAE by 17.7 MW on average and won on 95 of 165 days, but the worst
rolling seven-day value was slightly worse (2,048 versus 1,999 MW). It
was deployed because it was deterministic, timestamp-only, and
introduced neither a new publication dependency nor a leakage path, not
because a large live effect had been established. Since v7.11 entered
late and on a different weather chassis from that experiment, the 41-day
team score cannot isolate the contribution of solar elevation.

Several proposed changes did not pass the same evaluation discipline,
including output-bias correction, day-ahead price, polynomial and
interaction features, wider bridge-day rules, school-holiday features,
wind-only additions, and feature selection based on importance. The
general lesson from the sequence is that long chronological backtests
and explicit tail criteria were more reliable than short-window gains or
plausible feature narratives. The remaining limitations are substantial.
The live phase covers only 41 summer days in one bidding zone, the
operational result combines several model versions, accuracy inherits
target-day NWP error and availability, first-published load values can
later be revised, and the system emits point forecasts without
calibrated uncertainty. Future work should extend publication-vintage,
multi-season comparisons, evaluate all candidate versions on identical
origins, and add calibrated probabilistic forecasts so that rare-regime
and tail risk can be expressed rather than hidden in a point estimate.

\subsubsection{Team Neura}\label{team-neura-1}

Team Neura's forecaster finished seventh on the final leaderboard with a
mean MAE of 1,404.8 MW over 41 scored days, compared with 2,155.6 MW for
the ENTSO-E baseline, an improvement of approximately 35\%. The
submitted model is clearly more accurate than the ENTSO-E baseline in
both absolute and percentage error, as MAE, RMSE, MAPE, and MASE all
favour it. However, it performs worse on two secondary measures: its
full-phase bias (−378.5 MW) is much larger than the baseline's −38.4 MW,
and its under-prediction rate (60.5\%) is farther from balance than the
baseline's 51.4\%. The paper notes that the baseline's small full-phase
bias partly reflects averaging over an early over-forecasting episode
and a later under-forecasting period
(Section~\ref{sec-error-diagnostics}), so the two are not fully
comparable in terms of the consistency of their bias. The MAPE reranking
also places the team slightly lower, suggesting that relative forecast
error remains larger during some lower-demand periods.

The team's own seven-day chronological backtest of the submitted
pipeline (13--19 July 2026) resulted in a mean MAE of 1,074.1 MW
compared with 2,347.5 MW for weekly persistence and a mean MAPE of
2.06\% compared with 4.51\%. The model outperformed weekly persistence
on all seven days by MAE. The same backtest showed a mean bias of −428.1
MW, consistent with the under-forecasting tendency observed over the
full live phase.

Several alternative approaches were evaluated during development using
their own independent seven-day chronological backtests but were not
retained. Because each ran on a different live data window, the results
below are only comparable to that experiment's own persistence baseline,
not to each other or to the submitted model's backtest above. Rolling
24-hour weather-window features increased mean MAE from approximately
1,385.5 MW to 1,559.7 MW and MAPE from 2.97\% to 3.22\% against the same
persistence baseline, supporting the decision not to retain these
additional features. Weather-calendar interaction features reached a
mean MAE of 1,843.6 MW against a 2,120.3 MW persistence baseline,
beating persistence on 5/7 days. Replacing LightGBM with CatBoost
resulted in a mean MAE of 2,351.3 MW against a 3,311.7 MW persistence
baseline, beating persistence on 6/7 days. A residual-anchor formulation
based on weekly persistence reached 1,695.6 MW MAE against a 2,851.1 MW
persistence baseline, beating persistence on 5/7 days. Adding ENTSO-E
day-ahead price, solar, and wind forecasts to this approach resulted in
1,750.2 MW MAE against a 2,827.9 MW persistence baseline, also beating
persistence on 5/7 days. Rolling load statistics and a bias-correction
step were also explored during development, but no reproducing backtest
results for them are available. None of these experiments provided
sufficiently consistent evidence to replace the submitted recursive
LightGBM configuration.

A direct-horizon approach, in which separate LightGBM models were
trained for individual forecast horizons rather than recursively
forecasting all 24 hours, was the most promising alternative. It
achieved a mean MAE of 1,085.4 MW against a 1,757.9 MW persistence
baseline and a MAPE of 1.99\%, but only four valid backtest days were
available, so the result is not directly comparable with the complete
seven-day evaluations above and requires further validation.

Overall, the evidence supports the submitted forecasting pipeline as a
whole but does not establish that individual design choices, such as the
four-city weather representation or the 168-hour lag window,
independently caused the improvement, since no controlled ablation study
isolated these components. As a next step, the team would evaluate all
alternatives over the same longer backtest period, investigate the
systematic negative bias (for example through a probabilistic or
quantile forecast in place of a pure point forecast, in line with
Section~\ref{sec-conclusion}), and test the promising direct-horizon
approach under identical validation and fault-masking conditions.

\subsubsection{Team Weather
Report}\label{sec-findings-team-weather-report}

\paragraph{Overall standing}\label{overall-standing}

Team Weather Report finished last among the eleven student teams, but
the official ranking considerably understates how the forecaster
actually performed. Ten of the team's 41 scored days were carried
forward from the previous submission (last observation carried forward,
LOCF) rather than generated afresh. Excluding those ten LOCF days, the
forecaster's mean MAE on its 31 genuine submissions falls to 1556.4 MW,
around 31.5\% below the officially scored 2271.5 MW.

The recalculation would place the team fourth among the eleven, behind
Das A Team and ahead of Team Gladiators, Team Fabinalii, and
Eigen-Squad. This is consistent with the student leaderboard's own
significance analysis: the error is too volatile, largely because of the
carried-forward days, to be distinguished confidently from any other
entry, from Hot Rod, the best-placed team, down to the daily naive
benchmark, the least sophisticated forecast in the challenge.

Against the ENTSO-E baseline, the leaderboard score, 2271.5 MW versus
the baseline's 2155.6 MW, loses. The recalculated score, 1556.4 MW
against the same baseline, wins, by roughly 28\%.

The 1556.4 MW figure comes from days the team actually submitted, not
from a comparison against other teams on the same grounds like
Section~\ref{sec-significance}'s tests require, so it should be read as
indicative rather than statistically confirmed.

\paragraph{Misleading bias score}\label{misleading-bias-score}

Team Weather Report finished with the season's smallest mean bias, +4.8
MW across the 41 scored days. However, that figure alone does not tell
the real story: mean absolute bias over the same 41 days is 1930 MW, two
orders of magnitude larger. The mean bias is therefore misleading. Its
near-zero average comes from positive and negative errors canceling out,
not from an accurate forecast.

Until day 38 (2026-07-17), the mean bias stood at -305 MW, with 14
positive days and 24 negative days, a persistent negative-bias tendency
that the final days erase. On day 39 (2026-07-18), a LOCF submission
scored +9872 MW, shifting the mean bias from -305 MW to -44 MW in a
single step. The final two days, +794 MW and +1109 MW, then closed the
gap to the reported +4.8 MW. The best-bias result therefore stems from
one missed day arriving at the right moment near the end, not from any
genuine gain in accuracy.

\paragraph{What did not work}\label{what-did-not-work}

The persistent negative bias reflects a systematic undershoot: the
forecaster consistently predicts too low on the days that matter most
for the sign of the daily total. The team's first response was
mathematical: quantile-loss training, which shifts the model's training
target towards higher predictions instead of the average, and post-hoc
additive bias correction, which adds a fixed offset after the fact.
Neither improved on the model now in use. Tested on the same seven-date
backtest, quantile-loss reached 971 MW, and post-hoc correction reached
1132 MW, both well above the production configuration's 750 MW.

The second attempt abandoned that route entirely and targeted the
problem from the input side instead: day-ahead temperature, meant to
anticipate the model's error rather than correct it afterwards.
Unusually hot days can push load 5,000-10,000 MW above typical levels,
but the connection between load and temperature depends less on the
absolute value than on how unusual it is for the time of year, since
30°C in July is unremarkable, while the same temperature in April would
signal a real anomaly, and a simple reading of absolute temperature
cannot tell the two apart. This required independent implementation
rather than a setting in the library, which made it the hardest feature
of the two attempts to build. Two custom functions handled it: one
computed day-ahead maximum and mean temperature and
cooling-degree-hours, capturing how hot the day would be. The other
compared that same day-ahead temperature against the three-year average
for that hour and month and fed the resulting anomaly into both training
and the forecast for tomorrow, capturing how unusual it would be. Tested
across eleven hot summer dates, the simpler baseline still won 8 of 11,
with a lower average MAE, 981 MW against 1030 MW. Neither route resolved
the systematic undershoot.

\paragraph{What worked: tuning}\label{what-worked-tuning}

Section~\ref{sec-wr-tuning} reports the two ends of this fix, the
original reference configuration and the final result with widened
bounds and a larger trial budget, reaching 750 MW. What it does not show
is the two intermediate combinations, bounds widened without more
trials, and more trials without wider bounds, which a follow-up ablation
on the same seven-date backtest tested directly.

Widening the bounds alone (0.5 to 0.3 and 0.001 to 0.0001) made things
worse, not better: average MAE moved from 1258 to 1341 MW. At this
smaller trial budget, only \texttt{bagging\_fraction} (0.512 → 0.317)
and \texttt{reg\_lambda} (0.0013 → 0.000139) moved at all, tracking the
lower edge as it shifted from {[}0.5, 1.0{]} to {[}0.3, 1.0{]} and
{[}0.001, 100{]} to {[}0.0001, 100{]}. Every other hyperparameter
matched the reference exactly. The wider bounds alone gave the search
nothing to work with.

Increasing the trial budget alone, without touching either bound,
reduced average MAE to 1019 MW. Every hyperparameter shifted to match
the final configuration Section~\ref{sec-wr-tuning} reports, including
\texttt{bagging\_fraction}, except \texttt{reg\_lambda}, which stayed
exactly like the starting point.

Widening the bounds and increasing the budget together reached the full
750 MW reported in Section~\ref{sec-wr-tuning}, but the two parameters
that motivated the widening tell different stories once the full data is
checked. \texttt{bagging\_fraction} moves to roughly two-thirds of its
range once the trial budget is large, whether the bounds are the
original {[}0.5, 1.0{]} or the widened {[}0.3, 1.0{]}.
\texttt{reg\_lambda} tells the opposite story: with both bounds and
budget widened, it still just tracks the bound the same way it did at
small budget, down to 0.000193 this time as the wider floor pulls it
further.

To see why that matters, it helps to know what each parameter actually
does: \texttt{bagging\_fraction} controls how much of the training data
each tree sees per round, and \texttt{reg\_lambda} keeps the model
cautious about patterns that do not show up often in the data. Once both
bounds and budget were combined, \texttt{bagging\_fraction} moved too,
but only by a few percent, far short of \texttt{reg\_lambda}'s
order-of-magnitude move. \texttt{reg\_lambda} dropping that far means
the model was freed to lean more into rarer patterns it saw in the data,
rather than getting penalised for it and pushed towards learning only
common patterns. That is why, on top of the improvements trial budget
alone already brought to the other hyperparameters and to MAE, the bound
widening's main real effect, on \texttt{reg\_lambda}, delivered a
further MAE improvement of its own.

The resulting tree structure is also worth a closer look. The numbers
Section~\ref{sec-wr-tuning} reports, fewer and shallower trees combined
with a higher learning rate, first show up once trial budget alone
increases. Concretely, each individual tree became smaller and simpler.
Fewer leaves means each tree has fewer distinct answers to give, and a
shallower depth means fewer questions it gets to ask before settling on
one, so each tree makes broader distinctions than before. The higher
learning rate is a similar trade-off in a different place: it takes
bigger steps towards an optimum and reaches an answer faster, but risks
overshooting past it. The risk in both cases did not disappear just
because the result turned out well. This configuration is simply the one
that, in practice, delivered the team's best result.

\paragraph{What worked: weather
covariates}\label{what-worked-weather-covariates}

The rich bundle described in Section~\ref{sec-covariate-selection}
reduced average MAE by roughly 29\%, a real improvement, but the four
feature groups were only ever tested together, never individually. That
means it reflects the bundle's combined effect, not evidence for any one
feature on its own. Any single group could be carrying most of the
improvement, contributing nothing, or even working against the others.
Bundling that many covariates at once also risks adding noise rather
than signal, a risk this comparison cannot rule out either way. The
tuning ablation result above showed exactly how each change contributed
to the final result. No equivalent test has been run here yet to isolate
what each of the four feature groups actually does on its own.

\paragraph{Next step}\label{next-step}

The clearest untested lead is a weekday/weekend-conditional bias
correction. Over five weeks of data, weekday bias averaged 786.8 MW
against 1791.8 MW on weekends, too small a sample to be conclusive on
its own, but a clear enough gap to be worth testing directly. The
post-hoc correction described above used one fixed offset for every hour
of every day. A version that adjusts by weekday versus weekend, or even
by hour of day, would be more targeted, and might succeed where the flat
version did not.

Day-ahead temperature is also worth another attempt, but it was already
the hardest feature the team built, reaching into the pipeline by hand
rather than flipping a setting. Whether there is still room to refine
how it is expressed mathematically or whether the team has already
reached a practical ceiling for what this feature can offer, is an open
question worth answering before drawing a final conclusion. A
confidence-weighted version, scaling how much the anomaly signal counts
based on how reliable that day's forecast is, rather than treating it as
fully trusted or not, is one direction that has not been tried.

Tuning is also worth extending, but not blindly. A much larger trial
budget, \texttt{t100/i40}, ten times the production setting, was tried
once already, and it backfired: average MAE rose to 1027 MW against the
production result's 750 MW, and bias more than doubled to -854 MW. This
larger budget had real room to explore and found different sets of
parameters. Through that it found lower error on the training folds by
leaning harder into underprediction. However, that trade-off did not
hold up on the actual test dates. Both the error and the bias came back
worse. Therefore, the next step is trying different budget and initial
search configurations and different bounds to look for an even better
tuning configuration.

The weather covariates are also worth revisiting. A per-feature
ablation, testing each of the four groups separately, would settle the
open question raised above and show which features to keep, which to
discard, and where there might be room to add new ones.

\subsection{Weaknesses and threats to
validity}\label{weaknesses-and-threats-to-validity}

Four threats to validity limit these findings. The evaluation covers 41
summer days in a single bidding zone, with three public holidays and no
winter peaks, cold spells, or other stress regimes, so the results say
nothing about the seasons in which load forecasting is hardest. Entries
joined on different dates. Although all cross-entry claims rest on
shared-day paired tests, the partial-coverage means on the board remain
sample-dependent in the sense of Hewamalage et al. (2023). The
carried-forward rule conflates submission discipline with model quality,
as Team Weather Report's last place shows. Finally, within each day the
recursive strategy feeds predicted lags into later horizon steps, so the
final hour of the horizon rests on 23 predicted values, and the error
profile of Figure~\ref{fig-error-profile} averages over this
accumulation.

\section{Conclusions}\label{sec-conclusion}

This paper documented a complete day-ahead load-forecasting system for
the aggregated German transmission-grid load and evaluated it in a
finalised 41-day live challenge. The system combines gap-aware data
preparation with explicit anomaly annotation, a leakage-clean set of
thirty exogenous covariates, a recursive multi-step LightGBM forecaster
implemented in the safety-scoped package \texttt{spotforecast2-safe},
and daily surrogate-model hyperparameter tuning with \texttt{spotoptim}
benchmarked against Optuna. Every design choice was made under the
determinism, reproducibility, and auditability constraints motivated in
Section~\ref{sec-intro}, and every result in the paper reproduces at
render time from a committed snapshot of the finalised leaderboard data.

The main findings are straightforward. The reference pipeline built on
\texttt{spotforecast2-safe}, that is, the spotoptim-lgbm forecaster,
beats the ENTSO-E baseline on shared days by 34.7\% (\(p = 0.0001\)).
In-context models also perform competitively. The final board is topped
by such models on partial coverage, although their provenance and
auditability still have to be resolved before they can be considered for
safety-critical deployment. At the same time, low-cost,
energy-efficient, and auditable local models such as BABL2L remain
competitive with large pretrained foundation models such as Chronos-2.
Finally, the choice between \texttt{spotoptim} and Optuna does not
matter within the daily budget used here, because the two variants are
statistically tied (\(p = 0.98\)).

Five directions follow from these results. The first is a multi-season
evaluation spanning winter load regimes, which the 41 summer days
reported here cannot supply. The challenge continues beyond the phase
evaluated in this paper, and its ongoing results are published at
\url{https://advm1.gm.fh-koeln.de/~bartz/sf2-forecast}. The second is
the specification of legal requirements for safety-critical machine
learning. The AIA and the NIS-2 Directive are in force, but their
operational meaning for day-ahead load forecasting remains under
discussion. The next step is to formalise the relevant requirement set
for safety-critical AI, in discussion with experts, for example in the
``AK Explainability, Transparency, and Safety (ExTraSafe)'' (Fachbereich
Künstliche Intelligenz der Gesellschaft für Informatik 2026), and with
regulators and operators. The third concerns in-context and time-series
foundation models. Their shared-day performance in this challenge
(Section~\ref{sec-foundation}) makes them the natural next benchmark,
provided the provenance and audit questions raised in
Section~\ref{sec-discussion} are answered for the safety-critical
setting. The fourth is richer covariates, since enhanced feature
engineering and selection, especially of weather covariates, is a
promising direction for further improvement. The fifth is probabilistic
output. The persistent mild under-forecast and its reserve-provisioning
cost recommend quantile or interval forecasts in place of a pure MAE
point forecast.

\subsection*{Data and code
availability}\label{data-and-code-availability}
\addcontentsline{toc}{subsection}{Data and code availability}

All results in this paper are computed at render time from the committed
snapshot of the finalised challenge data. The challenge implementation
is publicly available at
\url{https://bartzbeielstein.github.io/challenge-leaderboard/}. The
challenge continues in a slightly modified form at
\url{https://advm1.gm.fh-koeln.de/~bartz/sf2-forecast}. Its board scores
a missed day as missing, without carry-forward or jokers, so its scores
are not directly comparable with the 41-day results reported here. The
STLF pipeline is open source. The packages \texttt{spotforecast2-safe}
(Bartz-Beielstein 2026d; Bartz-Beielstein and Bartz 2026),
\texttt{spotforecast2} (Bartz-Beielstein 2026c), and \texttt{spotoptim}
(Bartz-Beielstein 2026b) are released under the AGPL-3.0-or-later
licence. Load data are taken from the ENTSO-E Transparency Platform
(ENTSO-E 2024), and weather covariates are downloaded from Open-Meteo.

\subsection*{Competing interests}\label{competing-interests}
\addcontentsline{toc}{subsection}{Competing interests}

The first author develops the open-source packages
\texttt{spotforecast2-safe}, \texttt{spotforecast2}, and
\texttt{spotoptim} used in this report. No financial competing interests
are declared.

\subsection*{Use of AI tools}\label{use-of-ai-tools}
\addcontentsline{toc}{subsection}{Use of AI tools}

Parts of this report were researched, drafted, and translated with the
support of artificial intelligence. This information is provided
voluntarily, as no legal requirement mandates its disclosure.

\subsection*{References}\label{references}
\addcontentsline{toc}{subsection}{References}

\protect\phantomsection\label{refs}

\section*{Appendix}\label{appendix}
\addcontentsline{toc}{section}{Appendix}

\appendix

\section{Software: Implementation Details}\label{sec-software}

The three data-preparation stages of Section~\ref{sec-outliers} are
implemented in the open-source package \texttt{spotforecast2-safe}. This
appendix shows where each stage is located in the package and how it is
called, using a demonstration data set that ships with the package so
that every example runs offline and reproduces exactly. Each section
pairs a minimal executable code example with a figure that isolates the
effect of one stage. The appendix is written as an introductory
tutorial, and Section~\ref{sec-outliers} provides the methodological
context.

\begin{tcolorbox}[enhanced jigsaw, arc=.35mm, bottomrule=.15mm, bottomtitle=1mm, breakable, colback=white, colbacktitle=quarto-callout-note-color!10!white, colframe=quarto-callout-note-color-frame, coltitle=black, left=2mm, leftrule=.75mm, opacityback=0, opacitybacktitle=0.6, rightrule=.15mm, title=\textcolor{quarto-callout-note-color}{\faInfo}\hspace{0.5em}{Software versions}, titlerule=0mm, toprule=.15mm, toptitle=1mm]

All code in this appendix was executed with \texttt{spotforecast2-safe}
version 28.0.0 (Bartz-Beielstein 2026d), which implements the three
data-preparation stages, and with \texttt{spotforecast2} version 10.9.1
(Bartz-Beielstein 2026c), which supplies the plotting style for the
figures. Both version numbers are read from the installed packages at
render time.

\end{tcolorbox}

\subsection{Where the code lives}\label{sec-software-map}

Table~\ref{tbl-software-map} maps each production stage to its
implementation in the multi-task pipeline and to the method that calls
it. The executable snippets below call selected low-level utilities
directly to isolate their behaviour. Those calls are demonstrations, not
additional production workflows. \texttt{prepare\_data} applies Stage 1
target-corruption checks to the native fifteen-minute data and then
resamples to hourly values. \texttt{detect\_outliers} runs Stage 2's
hard-bound removal and advisory Isolation Forest check, and
\texttt{impute} runs Stage 3 and produces sample weights for the
forecaster. A target-corruption \texttt{heal} policy also carries state
into Stage 3, which selects weighted interpolation for the repaired
target slots.

\begin{longtable}[]{@{}
  >{\raggedright\arraybackslash}p{(\linewidth - 4\tabcolsep) * \real{0.2200}}
  >{\raggedright\arraybackslash}p{(\linewidth - 4\tabcolsep) * \real{0.4200}}
  >{\raggedright\arraybackslash}p{(\linewidth - 4\tabcolsep) * \real{0.3600}}@{}}
\caption{Where the three data-preparation stages of
Section~\ref{sec-outliers} are implemented in the multi-task pipeline.
Module paths are relative to the installed
\protect\texttt{spotforecast2\_safe} package, and the last column names
the inherited base-class methods invoked by
\protect\texttt{MultiTask}.}\label{tbl-software-map}\tabularnewline
\toprule\noalign{}
\begin{minipage}[b]{\linewidth}\raggedright
Stage of Section~\ref{sec-outliers}
\end{minipage} & \begin{minipage}[b]{\linewidth}\raggedright
Multi-task implementation
\end{minipage} & \begin{minipage}[b]{\linewidth}\raggedright
Multi-task call site
\end{minipage} \\
\midrule\noalign{}
\endfirsthead
\toprule\noalign{}
\begin{minipage}[b]{\linewidth}\raggedright
Stage of Section~\ref{sec-outliers}
\end{minipage} & \begin{minipage}[b]{\linewidth}\raggedright
Multi-task implementation
\end{minipage} & \begin{minipage}[b]{\linewidth}\raggedright
Multi-task call site
\end{minipage} \\
\midrule\noalign{}
\endhead
\bottomrule\noalign{}
\endlastfoot
1. target-corruption flagging &
\texttt{apply\_target\_corruption\_policy} in
\texttt{preprocessing.target\_corruption}, with the detector
\texttt{detect\_target\_corruption} & \texttt{BaseTask.prepare\_data},
called by \texttt{MultiTask} \\
2. anomaly flagging & advisory \texttt{get\_outliers} in
\texttt{preprocessing.outlier}, hard-bound removal by
\texttt{manual\_outlier\_removal} & \texttt{BaseTask.detect\_outliers},
called by \texttt{MultiTask} \\
3. gap imputation & \texttt{apply\_imputation} in
\texttt{preprocessing.imputation}, which selects the configured fill and
weighting strategy & \texttt{BaseTask.impute}, called by
\texttt{MultiTask} \\
\end{longtable}

\subsection{The demonstration series}\label{sec-software-data}

All examples in this section run on a synthetic demonstration series
generated in place with \texttt{make\_synthetic\_load}, the
deterministic generator that \texttt{spotforecast2-safe} ships for
offline tutorials. The hourly series is the sum of a base level of 50, a
slight upward trend, a daily cycle peaking at noon, a weekday uplift,
and seeded Gaussian noise, so its values read as stylised gigawatts of
German load. A noise-free twin of the same series serves as the
published day-ahead reference, which makes the two columns differ by
exactly the noise term. Because the generator is hourly while the
production data arrives at a fifteen-minute cadence, both columns are
upsampled by time interpolation, and a small seeded sub-hourly noise
term is added to the actual column only, so that the intra-hour dynamics
that Stage 1 analyses are non-trivial. The frame carries the production
column names ``Actual Load'' and ``Forecasted Load'' on a tz-aware UTC
index. No external data source, platform access, or API key is required.

\protect\phantomsection\label{software-setup}
\begin{Shaded}
\begin{Highlighting}[]
\ImportTok{import}\NormalTok{ logging}
\ImportTok{import}\NormalTok{ numpy }\ImportTok{as}\NormalTok{ np}
\ImportTok{import}\NormalTok{ pandas }\ImportTok{as}\NormalTok{ pd}
\ImportTok{from}\NormalTok{ spotforecast2\_safe.data }\ImportTok{import}\NormalTok{ make\_synthetic\_load}

\CommentTok{\# Synthetic demonstration series in stylised GW: trend, daily peak at}
\CommentTok{\# noon, weekday uplift, and seeded noise. The noise{-}free twin serves as}
\CommentTok{\# the day{-}ahead reference column.}
\NormalTok{sw\_actual }\OperatorTok{=}\NormalTok{ make\_synthetic\_load(}\StringTok{"2025{-}01{-}06"}\NormalTok{, }\StringTok{"2025{-}01{-}27 23:00"}\NormalTok{, seed}\OperatorTok{=}\DecValTok{2026}\NormalTok{)}
\NormalTok{sw\_forecast }\OperatorTok{=}\NormalTok{ make\_synthetic\_load(}\StringTok{"2025{-}01{-}06"}\NormalTok{, }\StringTok{"2025{-}01{-}27 23:00"}\NormalTok{, noise\_std}\OperatorTok{=}\FloatTok{0.0}\NormalTok{)}

\CommentTok{\# Upsample to the 15{-}minute production cadence by time interpolation and}
\CommentTok{\# add a small seeded sub{-}hourly noise term to the actual column only.}
\NormalTok{sw\_idx }\OperatorTok{=}\NormalTok{ pd.date\_range(}
\NormalTok{    sw\_actual.index[}\DecValTok{0}\NormalTok{], sw\_actual.index[}\OperatorTok{{-}}\DecValTok{1}\NormalTok{], freq}\OperatorTok{=}\StringTok{"15min"}\NormalTok{, tz}\OperatorTok{=}\StringTok{"UTC"}
\NormalTok{)}
\NormalTok{sw\_rng }\OperatorTok{=}\NormalTok{ np.random.default\_rng(}\DecValTok{2026}\NormalTok{)}
\NormalTok{sw\_demo }\OperatorTok{=}\NormalTok{ pd.DataFrame(}
\NormalTok{    \{}
        \StringTok{"Actual Load"}\NormalTok{: sw\_actual.reindex(sw\_idx).interpolate(}\StringTok{"time"}\NormalTok{)}
        \OperatorTok{+}\NormalTok{ sw\_rng.normal(}\FloatTok{0.0}\NormalTok{, }\FloatTok{0.1}\NormalTok{, }\BuiltInTok{len}\NormalTok{(sw\_idx)),}
        \StringTok{"Forecasted Load"}\NormalTok{: sw\_forecast.reindex(sw\_idx).interpolate(}\StringTok{"time"}\NormalTok{),}
\NormalTok{    \}}
\NormalTok{)}
\end{Highlighting}
\end{Shaded}

Three weeks suffice to demonstrate every stage and keep execution fast.

\begin{verbatim}
2109 slots from 2025-01-06 00:00:00+00:00 to 2025-01-27 23:00:00+00:00
\end{verbatim}

\begin{figure}[H]

\centering{

\pandocbounded{\includegraphics[keepaspectratio]{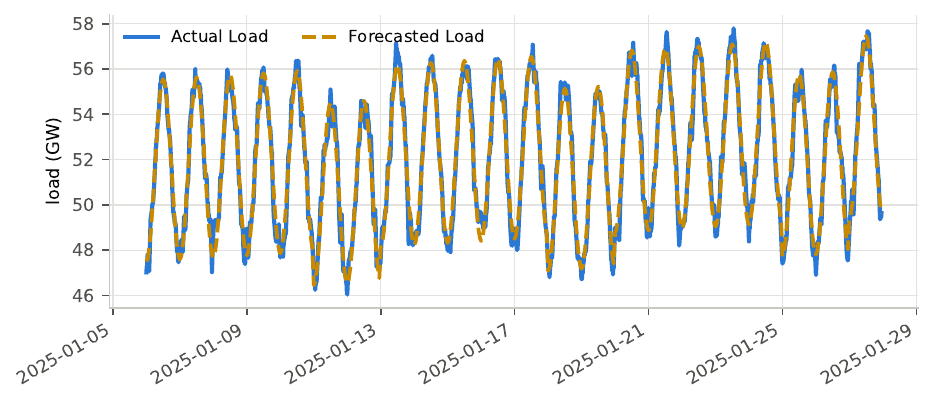}}

}

\caption{\label{fig-software-demo}The three-week synthetic demonstration
series used throughout this appendix, in stylised gigawatts. The blue
line is the Actual Load column, the sum of a base level, a slight trend,
a daily cycle peaking at noon, a weekday uplift, and seeded noise. The
dashed ochre line is its noise-free twin, the Forecasted Load column
that Stage 1 checks deviations against.}

\end{figure}

Because one unit reads as one gigawatt, the thresholds in the examples
below are the production values of the challenge lanes, given in MW,
divided by one thousand: an intra-hour range of 8000 MW, an adjacent
step of 6000 MW, and a deviation of 11000 MW below the reference.
Figure~\ref{fig-software-demo} shows the three weeks with both columns,
drawn with the forecast-overlay plotting tool of the
\texttt{spotforecast2} companion package. The daily cycle, the weekday
uplift, and the noise give the series the visual structure of real load
data, while the seeded generation keeps the appendix offline and
byte-reproducible. Because the series is clean by construction, each
example first injects a small, clearly marked artefact for its stage to
find.

\subsection{Stage 1: Target-corruption flagging and
healing}\label{sec-software-stage1}

Stage 1 of Section~\ref{sec-outliers} targets the corruption class that
a generic detector cannot see, namely sustained reporting dropouts of
the target series. In the multi-task pipeline,
\texttt{MultiTask.prepare\_data} calls
\texttt{apply\_target\_corruption\_policy} on the native fifteen-minute
data, which runs the detector \texttt{detect\_target\_corruption} and
then dispatches the configured policy. The example below calls that
utility directly to isolate its behaviour. It reconstructs the
characteristic geometry of such a dropout: the corrupted stretch ramps
in so gently that neither the intra-hour range rule nor the
adjacent-step rule fires. It sits far below the reference series and
only the one-sided deviation rule can discover it. Both ramps are
contained in hours that the deviation rule flags as well, which matters
for the figure below: the heal removes the episode in one piece, and no
partial ramp survives in the healed series.

\protect\phantomsection\label{software-stage1-inject}
\begin{Shaded}
\begin{Highlighting}[]
\CommentTok{\# Inject a sustained dropout below the reference: ramp down within one}
\CommentTok{\# hour, hold 11.6 GW below "Forecasted Load" for two hours, and ramp}
\CommentTok{\# back within one hour. The gentle ramp keeps every 15{-}minute step}
\CommentTok{\# below step\_mw and every intra{-}hour range below range\_mw, so only the}
\CommentTok{\# deviation rule can see the episode. Each ramp hour still contains two}
\CommentTok{\# consecutive slots deeper than deviation\_mw, so the deviation rule}
\CommentTok{\# flags the ramps together with the hold and the heal leaves no}
\CommentTok{\# partial ramp behind.}
\NormalTok{sw\_stage1 }\OperatorTok{=}\NormalTok{ sw\_demo.copy()}
\NormalTok{sw\_ref }\OperatorTok{=}\NormalTok{ sw\_stage1[}\StringTok{"Forecasted Load"}\NormalTok{]}
\NormalTok{sw\_ramp\_in }\OperatorTok{=}\NormalTok{ pd.date\_range(}\StringTok{"2025{-}01{-}25 04:00"}\NormalTok{, periods}\OperatorTok{=}\DecValTok{4}\NormalTok{, freq}\OperatorTok{=}\StringTok{"15min"}\NormalTok{, tz}\OperatorTok{=}\StringTok{"UTC"}\NormalTok{)}
\NormalTok{sw\_hold }\OperatorTok{=}\NormalTok{ pd.date\_range(}\StringTok{"2025{-}01{-}25 05:00"}\NormalTok{, periods}\OperatorTok{=}\DecValTok{8}\NormalTok{, freq}\OperatorTok{=}\StringTok{"15min"}\NormalTok{, tz}\OperatorTok{=}\StringTok{"UTC"}\NormalTok{)}
\NormalTok{sw\_ramp\_out }\OperatorTok{=}\NormalTok{ pd.date\_range(}\StringTok{"2025{-}01{-}25 07:00"}\NormalTok{, periods}\OperatorTok{=}\DecValTok{4}\NormalTok{, freq}\OperatorTok{=}\StringTok{"15min"}\NormalTok{, tz}\OperatorTok{=}\StringTok{"UTC"}\NormalTok{)}
\NormalTok{sw\_stage1.loc[sw\_ramp\_in, }\StringTok{"Actual Load"}\NormalTok{] }\OperatorTok{=}\NormalTok{ (}
\NormalTok{    sw\_ref.loc[sw\_ramp\_in] }\OperatorTok{{-}}\NormalTok{ np.array([}\FloatTok{4.8}\NormalTok{, }\FloatTok{8.8}\NormalTok{, }\FloatTok{11.2}\NormalTok{, }\FloatTok{11.6}\NormalTok{])}
\NormalTok{)}
\NormalTok{sw\_stage1.loc[sw\_hold, }\StringTok{"Actual Load"}\NormalTok{] }\OperatorTok{=}\NormalTok{ sw\_ref.loc[sw\_hold] }\OperatorTok{{-}} \FloatTok{11.6}
\NormalTok{sw\_stage1.loc[sw\_ramp\_out, }\StringTok{"Actual Load"}\NormalTok{] }\OperatorTok{=}\NormalTok{ (}
\NormalTok{    sw\_ref.loc[sw\_ramp\_out] }\OperatorTok{{-}}\NormalTok{ np.array([}\FloatTok{11.6}\NormalTok{, }\FloatTok{11.2}\NormalTok{, }\FloatTok{8.8}\NormalTok{, }\FloatTok{4.8}\NormalTok{])}
\NormalTok{)}
\end{Highlighting}
\end{Shaded}

With the dropout in place, a single call runs the detector and
dispatches the heal policy. The window of three days, the healing budget
of six hours, and the 48-hour anchor zone are the production values.
Only \texttt{deviation\_slots} is raised from one to two consecutive
slots, which makes the demonstration stricter than production.

\protect\phantomsection\label{software-stage1-run}
\begin{Shaded}
\begin{Highlighting}[]
\ImportTok{from}\NormalTok{ spotforecast2\_safe.preprocessing.target\_corruption }\ImportTok{import}\NormalTok{ (}
\NormalTok{    apply\_target\_corruption\_policy,}
\NormalTok{)}

\CommentTok{\# Thresholds are the production values of the challenge lanes, read in}
\CommentTok{\# GW: intra{-}hour range 8, adjacent step 6, and deviation vs reference 11.}
\CommentTok{\# deviation\_slots=2 is stricter than the production value of 1, so that}
\CommentTok{\# the episode below must be two slots deep to be flagged.}
\NormalTok{sw\_healed, sw\_report }\OperatorTok{=}\NormalTok{ apply\_target\_corruption\_policy(}
\NormalTok{    sw\_stage1,}
\NormalTok{    targets}\OperatorTok{=}\NormalTok{[}\StringTok{"Actual Load"}\NormalTok{],}
\NormalTok{    policy}\OperatorTok{=}\StringTok{"heal"}\NormalTok{,}
\NormalTok{    range\_mw}\OperatorTok{=}\FloatTok{8.0}\NormalTok{,}
\NormalTok{    step\_mw}\OperatorTok{=}\FloatTok{6.0}\NormalTok{,}
\NormalTok{    window\_days}\OperatorTok{=}\DecValTok{3}\NormalTok{,}
\NormalTok{    max\_heal\_hours}\OperatorTok{=}\DecValTok{6}\NormalTok{,}
\NormalTok{    anchor\_zone\_hours}\OperatorTok{=}\DecValTok{48}\NormalTok{,}
\NormalTok{    cutoff}\OperatorTok{=}\NormalTok{sw\_stage1.index[}\OperatorTok{{-}}\DecValTok{1}\NormalTok{],}
\NormalTok{    logger}\OperatorTok{=}\NormalTok{logging.getLogger(}\StringTok{"software{-}demo"}\NormalTok{),}
\NormalTok{    deviation\_mw}\OperatorTok{=}\FloatTok{11.0}\NormalTok{,}
\NormalTok{    deviation\_ref}\OperatorTok{=}\StringTok{"Forecasted Load"}\NormalTok{,}
\NormalTok{    deviation\_slots}\OperatorTok{=}\DecValTok{2}\NormalTok{,}
\NormalTok{)}
\BuiltInTok{print}\NormalTok{(sw\_report.action, sw\_report.n\_flagged\_hours, sw\_report.spans)}
\end{Highlighting}
\end{Shaded}

\begin{verbatim}
heal 4 [('2025-01-25T04:00:00+00:00', '2025-01-25T07:00:00+00:00')]
\end{verbatim}

\begin{figure}[H]

\centering{

\pandocbounded{\includegraphics[keepaspectratio]{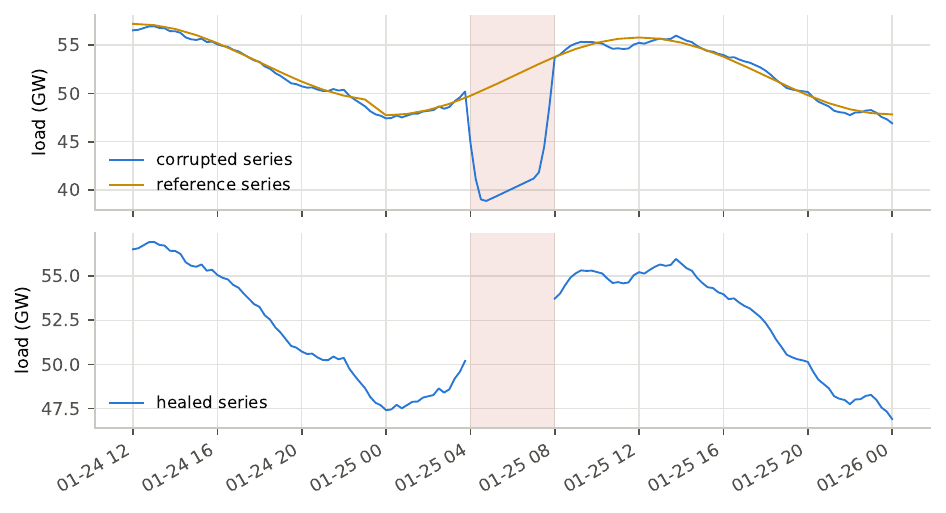}}

}

\caption{\label{fig-software-stage1}Stage 1 on the demonstration slice,
zoomed to the injected episode. Top: the corrupted series (blue) drops
below the reference series (ochre) gently enough to evade the range and
step rules, and the deviation rule flags the shaded hours, ramps
included. Bottom: the heal policy sets every slot of the flagged hours
to missing, so the healed series stops at the episode boundary and
resumes cleanly after it, with no partial ramp left behind. Closing the
hole is the job of Stage 3, not of this stage.}

\end{figure}%

The returned \texttt{TargetCorruptionReport} is the audit artefact of
the stage. It records whether the detector fired, how many hours were
flagged, the contiguous spans, and the action taken, which is what a
reviewer needs to reconstruct the decision. In the multi-task pipeline,
the healed hours from \texttt{MultiTask.prepare\_data} reach
\texttt{MultiTask.impute} as ordinary gaps. Had the episode exceeded the
healing budget of \texttt{max\_heal\_hours} or touched the 48-hour
anchor zone before the forecast origin, the policy would have refused to
heal, following the flag-and-refuse principle of the data-governance
rules of Bartz-Beielstein and Bartz (2026).

\subsection{\texorpdfstring{Stage 2: Anomaly flagging with the
\texttt{mark\_outliers}
utility}{Stage 2: Anomaly flagging with the mark\_outliers utility}}\label{sec-software-stage2}

Stage 2 of Section~\ref{sec-outliers} runs after \texttt{prepare\_data}
has resampled the series to hourly values. The multi-task pipeline keeps
its Isolation Forest advisory: \texttt{MultiTask.detect\_outliers} calls
\texttt{get\_outliers} and logs the suspicious values without removing
them. The example below instead demonstrates the separate cleaning
utility \texttt{mark\_outliers}, which fits the forest independently per
column and returns two new objects: a cleaned frame in which every
flagged slot is \texttt{NaN}, and the boolean mask of the flags. Since
release 28.0.0 of \texttt{spotforecast2-safe} the function is pure. The
frame it receives is never modified, so the spiked series
\texttt{sw\_stage2} remains the before state without any defensive copy,
and the optional \texttt{columns} argument selects the columns to fit,
which is why the frame no longer has to be restricted before the call.

\protect\phantomsection\label{software-stage2-run}
\begin{Shaded}
\begin{Highlighting}[]
\ImportTok{from}\NormalTok{ spotforecast2\_safe.preprocessing.outlier }\ImportTok{import}\NormalTok{ mark\_outliers}

\CommentTok{\# The demonstration series is clean, so inject three artificial spikes}
\CommentTok{\# of +8 GW at fixed positions. The forest sees values only, so the}
\CommentTok{\# spikes sit in daytime hours, where +8 GW leaves the value range that}
\CommentTok{\# the daily cycle spans.}
\NormalTok{sw\_stage2 }\OperatorTok{=}\NormalTok{ sw\_demo[[}\StringTok{"Actual Load"}\NormalTok{]].copy()}
\NormalTok{sw\_spikes }\OperatorTok{=}\NormalTok{ pd.to\_datetime(}
\NormalTok{    [}\StringTok{"2025{-}01{-}10 10:00"}\NormalTok{, }\StringTok{"2025{-}01{-}15 12:30"}\NormalTok{, }\StringTok{"2025{-}01{-}21 14:45"}\NormalTok{]}
\NormalTok{).tz\_localize(}\StringTok{"UTC"}\NormalTok{)}
\NormalTok{sw\_stage2.loc[sw\_spikes, }\StringTok{"Actual Load"}\NormalTok{] }\OperatorTok{+=} \FloatTok{8.0}

\NormalTok{sw\_flagged, sw\_labels }\OperatorTok{=}\NormalTok{ mark\_outliers(}
\NormalTok{    sw\_stage2, contamination}\OperatorTok{=}\FloatTok{0.005}\NormalTok{, random\_state}\OperatorTok{=}\DecValTok{1234}
\NormalTok{)}
\NormalTok{sw\_nan\_idx }\OperatorTok{=}\NormalTok{ sw\_stage2.index[sw\_labels[}\StringTok{"Actual Load"}\NormalTok{]]}
\BuiltInTok{print}\NormalTok{(}\SpecialStringTok{f"Slots flagged and set to NaN: }\SpecialCharTok{\{}\BuiltInTok{len}\NormalTok{(sw\_nan\_idx)}\SpecialCharTok{\}}\SpecialStringTok{"}\NormalTok{)}
\end{Highlighting}
\end{Shaded}

\begin{verbatim}
Slots flagged and set to NaN: 10
\end{verbatim}

\begin{figure}[H]

\centering{

\pandocbounded{\includegraphics[keepaspectratio]{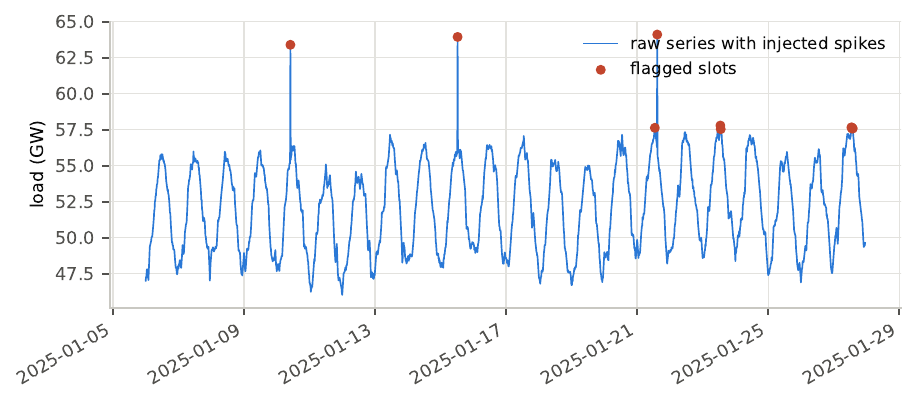}}

}

\caption{\label{fig-software-stage2}Stage 2 on the demonstration slice.
The blue line shows the raw series with the three injected spikes, and
the brick markers show the slots that the Isolation Forest flags and
sets to missing. With a contamination of 0.005 the forest flags the
three spikes together with a handful of genuine extremes of the series.}

\end{figure}%

The \texttt{contamination} parameter is a flag budget and not a
threshold. The forest flags approximately that fraction of the points
regardless of how extreme they are, which is why the run above marks a
handful of genuine extremes in addition to the three injected spikes.
The flagged slots become \texttt{NaN} instead of being corrected. This
causes Stage 3 to treat them exactly like native gaps. The fixed
\texttt{random\_state} makes this utility demonstration repeatable
across runs.

\subsection{\texorpdfstring{Stage 3: Gap imputation and sample weights
with the \texttt{get\_missing\_weights}
utility}{Stage 3: Gap imputation and sample weights with the get\_missing\_weights utility}}\label{sec-software-stage3}

Stage 3 closes every gap that the first two stages left behind. In the
multi-task pipeline, \texttt{MultiTask.impute} delegates to
\texttt{apply\_imputation}, which selects the configured fill and
weighting strategy and stores the resulting weight function for the
forecaster. The example below calls the lower-level
\texttt{get\_missing\_weights} utility directly: it fills missing slots
by forward and backward filling and returns, next to the filled frame, a
weight series that is zero inside a gap and for a trailing window after
it and one everywhere else. The example injects a six-hour gap into the
demonstration slice.

\protect\phantomsection\label{software-stage3-run}
\begin{Shaded}
\begin{Highlighting}[]
\ImportTok{from}\NormalTok{ spotforecast2\_safe.preprocessing.imputation }\ImportTok{import}\NormalTok{ get\_missing\_weights}

\CommentTok{\# Inject a six{-}hour gap (24 slots at the 15{-}minute cadence).}
\NormalTok{sw\_stage3 }\OperatorTok{=}\NormalTok{ sw\_demo.copy()}
\NormalTok{sw\_gap }\OperatorTok{=}\NormalTok{ pd.date\_range(}
    \StringTok{"2025{-}01{-}17 10:00"}\NormalTok{, }\StringTok{"2025{-}01{-}17 16:00"}\NormalTok{, freq}\OperatorTok{=}\StringTok{"15min"}\NormalTok{, tz}\OperatorTok{=}\StringTok{"UTC"}\NormalTok{,}
\NormalTok{    inclusive}\OperatorTok{=}\StringTok{"left"}\NormalTok{,}
\NormalTok{)}
\NormalTok{sw\_stage3.loc[sw\_gap, }\StringTok{"Actual Load"}\NormalTok{] }\OperatorTok{=}\NormalTok{ np.nan}

\CommentTok{\# window\_size counts rows: 96 slots of 15 minutes give a 24{-}hour}
\CommentTok{\# zero{-}weight zone after the gap.}
\NormalTok{sw\_filled, sw\_weights }\OperatorTok{=}\NormalTok{ get\_missing\_weights(sw\_stage3, window\_size}\OperatorTok{=}\DecValTok{96}\NormalTok{)}
\BuiltInTok{print}\NormalTok{(}\SpecialStringTok{f"NaN slots remaining: }\SpecialCharTok{\{}\NormalTok{sw\_filled}\SpecialCharTok{.}\NormalTok{isna()}\SpecialCharTok{.}\BuiltInTok{sum}\NormalTok{()}\SpecialCharTok{.}\BuiltInTok{sum}\NormalTok{()}\SpecialCharTok{\}}\SpecialStringTok{"}\NormalTok{)}
\BuiltInTok{print}\NormalTok{(}\SpecialStringTok{f"Zero{-}weight slots: }\SpecialCharTok{\{}\BuiltInTok{int}\NormalTok{((sw\_weights }\OperatorTok{==} \DecValTok{0}\NormalTok{).}\BuiltInTok{sum}\NormalTok{())}\SpecialCharTok{\}}\SpecialStringTok{ of }\SpecialCharTok{\{}\BuiltInTok{len}\NormalTok{(sw\_weights)}\SpecialCharTok{\}}\SpecialStringTok{"}\NormalTok{)}
\end{Highlighting}
\end{Shaded}

\begin{verbatim}
NaN slots remaining: 0
Zero-weight slots: 120 of 2109
\end{verbatim}

\begin{figure}[H]

\centering{

\pandocbounded{\includegraphics[keepaspectratio]{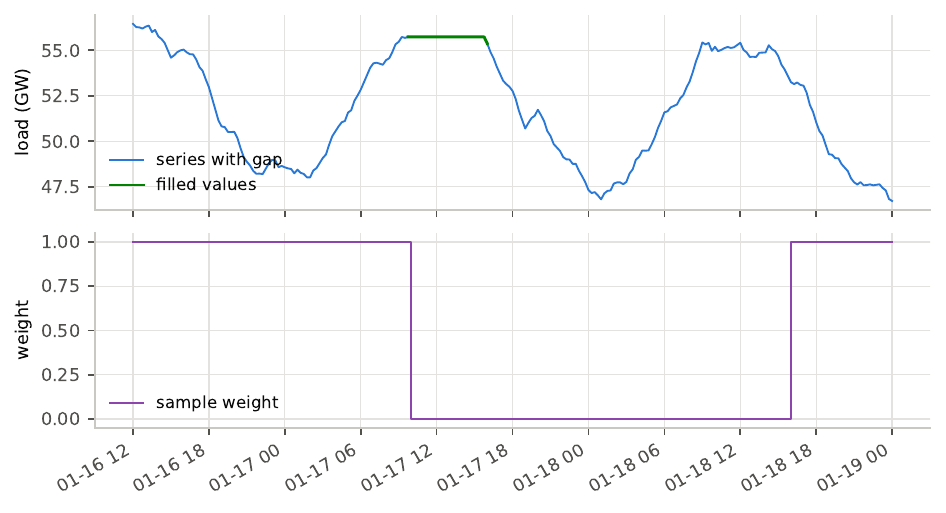}}

}

\caption{\label{fig-software-stage3}Stage 3 on the demonstration slice,
zoomed to the injected six-hour gap. Top: the series with the gap (blue)
and the values that the fill inserts to bridge it (green). Bottom: the
companion weight series (violet) drops to zero inside the gap and stays
there for a further 24 hours, so the fit ignores both the filled values
and the samples whose lag features would ingest them.}

\end{figure}%

The filled values keep the frame gap-free, which the feature
construction of Section~\ref{sec-covariates} requires, while the zero
weights remove them from the fit. The zero-weight zone extends one
\texttt{window\_size} beyond the gap because the lag features of those
samples would otherwise ingest filled values. In the multi-task
pipeline, \texttt{MultiTask.impute} passes the output of
\texttt{apply\_imputation} to the recursive forecaster of
Section~\ref{sec-recursive} through its stored weight function. When
Stage 1 heals a flagged stretch, \texttt{prepare\_data} records that
policy decision and \texttt{impute} temporarily selects
\texttt{weighted\_interp}: it bridges a gap by drawing a straight line
between the last observed value before it and the first one after it,
applies the same zero weights, and falls back to the nearest observed
hours only where no such bracketing exists. For ordinary configured
gaps, the default forward fill freezes the last value before the gap,
and Figure~\ref{fig-software-stage3} shows the resulting step where the
fill rejoins the series. The refusal to fabricate values sits one stage
earlier and at the series edge: the corruption policy of Stage 1 aborts
by default, and trailing slots without an observed target are truncated
rather than filled, as Section~\ref{sec-outliers} describes.

\section{Notation}\label{sec-notation}

Table~\ref{tbl-notation} summarises the notation used throughout the
paper. Our notation follows the symbol conventions of Hyndman and
Athanasopoulos (2021).

\begin{longtable}[]{@{}
  >{\raggedright\arraybackslash}p{(\linewidth - 2\tabcolsep) * \real{0.2222}}
  >{\raggedright\arraybackslash}p{(\linewidth - 2\tabcolsep) * \real{0.7778}}@{}}
\caption{Notation used throughout the
paper.}\label{tbl-notation}\tabularnewline
\toprule\noalign{}
\begin{minipage}[b]{\linewidth}\raggedright
Symbol
\end{minipage} & \begin{minipage}[b]{\linewidth}\raggedright
Meaning
\end{minipage} \\
\midrule\noalign{}
\endfirsthead
\toprule\noalign{}
\begin{minipage}[b]{\linewidth}\raggedright
Symbol
\end{minipage} & \begin{minipage}[b]{\linewidth}\raggedright
Meaning
\end{minipage} \\
\midrule\noalign{}
\endhead
\bottomrule\noalign{}
\endlastfoot
\(y_t\) & observed (actual) load at hour \(t\), in megawatts (MW) \\
\(\hat{y}_t\) & point forecast of \(y_t\) \\
\(\hat{y}_{T+h\mid T}\) & \(h\)-step-ahead forecast for time \(T+h\)
made at forecast origin \(T\) \\
\(\tilde{y}_s\) & substituted series, equal to the observed load for
\(s \le T\) and to the running forecast for \(s > T\) \\
\(T\) & forecast origin, equivalently the training-sample size \\
\(h\) & forecast lead time (step index), \(h = 1,\dots,H\) \\
\(H\) & forecast horizon, here \(H = 24\) \\
\(m\) & seasonal period, here \(m = 24\) (one day) \\
\(\mathcal{L}\) & set of autoregressive lags, default
\(\mathcal{L} = \{1, 2, 24\}\), tuned over the pool of
Table~\ref{tbl-searchspace} \\
\(w\) & history window per training row,
\(w = \max(\max \mathcal{L},\, 72)\) hours, where 72 is the rolling-mean
window of Section~\ref{sec-dataset} \\
\(\mathbf{x}_t\) & vector of exogenous covariates available at hour
\(t\) \\
\(\mathbf{z}\) & stacked input vector of the regression function (lags
and covariates) \\
\(g(\,\cdot\,;\boldsymbol{\theta})\) & LightGBM regression function with
hyperparameters \(\boldsymbol{\theta}\) \\
\(K\) & number of boosting iterations (trees) \\
\(f_k\) & \(k\)-th regression tree \\
\(\ell\) & training loss (squared error) \\
\(\Omega\) & tree-complexity penalty \\
\(\boldsymbol{\theta},\ \boldsymbol{\theta}^{*}\) & hyperparameter
vector and its optimum \\
\(\Theta\) & hyperparameter search space \\
\(e_t\) & forecast error, \(e_t = y_t - \hat{y}_t\) \\
\end{longtable}

\section{Definitions}\label{sec-definitions}

\begin{definition}[Mean absolute
error]\protect\hypertarget{def-mae}{}\label{def-mae}

The MAE is defined as

\begin{equation}\protect\phantomsection\label{eq-mae}{
\operatorname{MAE} = \frac{1}{H}\sum_{h=1}^{H}\left| y_{T+h} - \hat{y}_{T+h\mid T} \right|,
}\end{equation}

the average over the day of the absolute hourly forecast errors
\(e_{T+h}=y_{T+h}-\hat{y}_{T+h\mid T}\). The public leaderboard ranks
teams by the MAE of Equation~\ref{eq-mae} averaged across all scored
days. Lower MAE values indicate better forecasts.

\end{definition}

\begin{definition}[RMSE]\protect\hypertarget{def-rmse}{}\label{def-rmse}

The RMSE,

\begin{equation}\protect\phantomsection\label{eq-rmse}{
\operatorname{RMSE} = \sqrt{\frac{1}{H}\sum_{h=1}^{H}\left(y_{T+h}-\hat{y}_{T+h\mid T}\right)^{2}},
}\end{equation}

squares the hourly errors before averaging and therefore weights large
deviations more heavily than Equation~\ref{eq-mae} does.

\end{definition}

\begin{definition}[MAPE]\protect\hypertarget{def-mape}{}\label{def-mape}

The MAPE,

\begin{equation}\protect\phantomsection\label{eq-mape}{
\operatorname{MAPE} = \frac{100\%}{H}\sum_{h=1}^{H}\left|\frac{y_{T+h}-\hat{y}_{T+h\mid T}}{y_{T+h}}\right|,
}\end{equation}

expresses the same absolute errors relative to the actual load. This
normalisation makes accuracy comparable across days of differing demand
level.

\end{definition}

\begin{definition}[Bias]\protect\hypertarget{def-bias}{}\label{def-bias}

The mean bias is the signed counterpart of Equation~\ref{eq-mae},

\begin{equation}\protect\phantomsection\label{eq-bias}{
\operatorname{Bias} = \frac{1}{H}\sum_{h=1}^{H}\left(\hat{y}_{T+h\mid T}-y_{T+h}\right),
}\end{equation}

defined as forecast minus actual, so that a negative value signals a
systematic under-forecast and a positive value a systematic
over-forecast.

\end{definition}

\begin{definition}[UPR]\protect\hypertarget{def-upr}{}\label{def-upr}

The UPR reports the share of hours in the day on which the forecast
falls below the actual load,

\begin{equation}\protect\phantomsection\label{eq-upr}{
\operatorname{UPR} = \frac{1}{H}\sum_{h=1}^{H}\mathbb{1}\!\left[\hat{y}_{T+h\mid T} < y_{T+h}\right],
}\end{equation}

with \(\mathbb{1}[\cdot]\) the indicator function, equal to one when its
argument is true and zero otherwise.

\end{definition}

For comparability across series of different scale we additionally
report the MASE.

\begin{definition}[MASE]\protect\hypertarget{def-mase}{}\label{def-mase}

It divides the MAE by the average one-step naive forecast error of the
in-sample series,

\begin{equation}\protect\phantomsection\label{eq-mase}{
\operatorname{MASE} = \operatorname{MAE} \Big/ \frac{1}{T-1}\sum_{t=2}^{T}\left|y_t - y_{t-1}\right|,
}\end{equation}

so that a value below one identifies a forecaster that beats the naive
one-step benchmark (Hyndman and Athanasopoulos 2021).

\end{definition}

\section{Original wording of the legal provisions
cited}\label{sec-legal-texts}

The regulatory argument of Section~\ref{sec-intro} rests on a small
number of legal provisions. This appendix reproduces them verbatim from
the cited versions so that the paper's characterisation can be checked
without consulting the original sources. The three Union instruments are
quoted from the English version of the Official Journal (European
Parliament and Council of the European Union 2024; European Parliament
and Council 2024b, 2024a), and the two German texts are quoted in their
official German wording (Bundesministerium des Innern 2026; Deutscher
Bundestag 2025). Quotations preserve the original numbering, and page
numbers refer to the cited documents.

\subsection{EU AI Act, Regulation (EU) 2024/1689}\label{sec-app-aiact}

\subsubsection{Recital 55}\label{sec-app-recital-55}

The recital (p.~15) carries the safety-component notion for critical
infrastructure on which the perimeter argument of
Section~\ref{sec-intro} rests:

\begin{quote}
\begin{enumerate}
\def\labelenumi{(\arabic{enumi})}
\setcounter{enumi}{54}
\tightlist
\item
  As regards the management and operation of critical infrastructure, it
  is appropriate to classify as high-risk the AI systems intended to be
  used as safety components in the management and operation of critical
  digital infrastructure as listed in point (8) of the Annex to
  Directive (EU) 2022/2557, road traffic and the supply of water, gas,
  heating and electricity, since their failure or malfunctioning may put
  at risk the life and health of persons at large scale and lead to
  appreciable disruptions in the ordinary conduct of social and economic
  activities. Safety components of critical infrastructure, including
  critical digital infrastructure, are systems used to directly protect
  the physical integrity of critical infrastructure or the health and
  safety of persons and property but which are not necessary in order
  for the system to function. The failure or malfunctioning of such
  components might directly lead to risks to the physical integrity of
  critical infrastructure and thus to risks to health and safety of
  persons and property. Components intended to be used solely for
  cybersecurity purposes should not qualify as safety components.
  Examples of safety components of such critical infrastructure may
  include systems for monitoring water pressure or fire alarm
  controlling systems in cloud computing centres.
\end{enumerate}
\end{quote}

\subsubsection{Article 6(2) and Annex III, point 2}\label{sec-app-art6}

Together the two provisions (pp.~53 and 127) form the classification
rule behind the statement that the AIA classifies an AI system used as a
safety component in the supply of electricity as high-risk:

\begin{quote}
2. In addition to the high-risk AI systems referred to in paragraph 1,
AI systems referred to in Annex III shall be considered to be high-risk.
\end{quote}

\begin{quote}
2. Critical infrastructure: AI systems intended to be used as safety
components in the management and operation of critical digital
infrastructure, road traffic, or in the supply of water, gas, heating or
electricity.
\end{quote}

\subsubsection{Article 15(1)}\label{sec-app-art15}

The paragraph (p.~61) states the accuracy, robustness, and cybersecurity
requirement:

\begin{quote}
1. High-risk AI systems shall be designed and developed in such a way
that they achieve an appropriate level of accuracy, robustness, and
cybersecurity, and that they perform consistently in those respects
throughout their lifecycle.
\end{quote}

\subsubsection{Article 12(1)}\label{sec-app-art12}

The paragraph (p.~59) states the record-keeping requirement:

\begin{quote}
1. High-risk AI systems shall technically allow for the automatic
recording of events (logs) over the lifetime of the system.
\end{quote}

\subsubsection{Article 11(1), first sentence}\label{sec-app-art11}

The sentence (p.~58) concerns the technical documentation:

\begin{quote}
1. The technical documentation of a high-risk AI system shall be drawn
up before that system is placed on the market or put into service and
shall be kept up-to date.
\end{quote}

\subsubsection{Article 10(1)}\label{sec-app-art10}

The paragraph (p.~57) frames data and data governance, the context of
the gap-aware preparation of Section~\ref{sec-outliers}:

\begin{quote}
1. High-risk AI systems which make use of techniques involving the
training of AI models with data shall be developed on the basis of
training, validation and testing data sets that meet the quality
criteria referred to in paragraphs 2 to 5 whenever such data sets are
used.
\end{quote}

\subsection{Kritisverordnung, Referentenentwurf of 26 May
2026}\label{sec-app-kritisv}

\subsubsection{§ 2 Abs. 1 and 2}\label{sec-app-kritisv-2}

The provision (p.~5) designates the supply of electricity as a critical
service and enumerates the areas it comprises, namely generation,
transmission, distribution, and trading:

\begin{quote}
§ 2 Kritische Dienstleistungen und kritische Anlagen im Sektor Energie
(1) Im Sektor Energie (§ 4 Absatz 1 Nummer 1 des KRITIS-Dachgesetzes)
sind kritische Dienstleistungen die Versorgung der Allgemeinheit 1. mit
Elektrizität (Stromversorgung); 2. mit Gas (Gasversorgung); 3. mit
Kraftstoff und Heizöl (Kraftstoff- und Heizölversorgung); 4. mit
Fernwärme und Fernkälte (Fernwärme und -kälteversorgung). (2) Die
Stromversorgung umfasst die folgenden Bereiche 1. Stromerzeugung, 2.
Stromübertragung, 3. Stromverteilung und 4. Stromhandel.
\end{quote}

\subsubsection{Anhang 1 Teil 3 Nummer 1.2.1 and Nummer
2.5}\label{sec-app-kritisv-anhang}

The annex row (p.~19) sets the transmission-network threshold, listing
the Anlagenkategorie, the Bemessungskriterium, and the Schwellenwert:

\begin{quote}
1.2. Stromübertragung 1.2.1 Übertragungsnetz Durch Letztverbraucher und
Weiterverteiler entnommene Jahresarbeit in GWh/Jahr 3 700
\end{quote}

The installation category is defined in Anhang 1 Nummer 2.5 (p.~13):

\begin{quote}
2.5 Übertragungsnetz ein Netz zur Übertragung im Sinne des § 3 Nummer
100 des Energiewirtschaftsgesetzes.
\end{quote}

\subsection{BSI-Gesetz (BSIG), BGBl. 2025 I Nr. 301}\label{sec-app-bsig}

\subsubsection{Title of the statute}\label{sec-app-bsig-title}

The title (p.~1) announces an information-security statute:

\begin{quote}
Gesetz über das Bundesamt für Sicherheit in der Informationstechnik und
über die Sicherheit in der Informationstechnik von Einrichtungen
(BSI-Gesetz -- BSIG)
\end{quote}

\subsubsection{§ 2 Nr. 22, 24, and 39}\label{sec-app-bsig-2}

The three definitions (p.~6) fix the notions of critical installation,
critical service, and security in information technology:

\begin{quote}
22. „kritische Anlage`` eine Anlage, die für die Erbringung einer
kritischen Dienstleistung erheblich ist; die kritischen Anlagen im Sinne
dieses Gesetzes werden durch die Rechtsverordnung nach § 56 Absatz 4
näher bestimmt;
\end{quote}

\begin{quote}
24. „kritische Dienstleistung`` eine Dienstleistung zur Versorgung der
Allgemeinheit in den Sektoren Energie, Transport und Verkehr,
Finanzwesen, Leistungen der Sozialversicherung sowie der Grundsicherung
für Arbeitsuchende, Gesundheitswesen, Wasser, Ernährung,
Informationstechnik und Telekommunikation, Weltraum oder
Siedlungsabfallentsorgung, deren Ausfall oder Beeinträchtigung zu
erheblichen Versorgungsengpässen oder zu Gefährdungen der öffentlichen
Sicherheit führen würde;
\end{quote}

\begin{quote}
39. „Sicherheit in der Informationstechnik`` die Einhaltung bestimmter
Sicherheitsstandards, die die Verfügbarkeit, Integrität oder
Vertraulichkeit von Informationen betreffen, durch
Sicherheitsvorkehrungen a) in informationstechnischen Systemen,
Komponenten oder Prozessen oder b) bei der Anwendung
informationstechnischer Systeme, Komponenten oder Prozesse;
\end{quote}

\subsubsection{§ 28 Abs. 8}\label{sec-app-bsig-28}

The provision (p.~21) defines the operator of critical installations:

\begin{quote}
\begin{enumerate}
\def\labelenumi{(\arabic{enumi})}
\setcounter{enumi}{7}
\tightlist
\item
  Ein Betreiber kritischer Anlagen ist eine natürliche oder juristische
  Person oder eine rechtlich unselbstständige Organisationseinheit einer
  Gebietskörperschaft, die unter Berücksichtigung der rechtlichen,
  wirtschaftlichen und tatsächlichen Umstände bestimmenden Einfluss auf
  eine oder mehrere kritische Anlagen ausübt. Abweichend von Satz 1 hat
  im Sektor Finanzwesen bestimmenden Einfluss auf eine Anlage, wer die
  tatsächliche Sachherrschaft ausübt. Die rechtlichen und
  wirtschaftlichen Umstände bleiben insoweit unberücksichtigt.
\end{enumerate}
\end{quote}

\subsubsection{§ 30 Abs. 1}\label{sec-app-bsig-30}

The provision (p.~22) states the core risk-management obligation:

\begin{quote}
\begin{enumerate}
\def\labelenumi{(\arabic{enumi})}
\tightlist
\item
  Besonders wichtige Einrichtungen und wichtige Einrichtungen sind
  verpflichtet, geeignete, verhältnismäßige und wirksame technische und
  organisatorische Maßnahmen, die in Absatz 2 konkretisiert werden, zu
  ergreifen, um Störungen der Verfügbarkeit, Integrität und
  Vertraulichkeit der informationstechnischen Systeme, Komponenten und
  Prozesse, die sie für die Erbringung ihrer Dienste nutzen, zu
  vermeiden und Auswirkungen von Sicherheitsvorfällen möglichst gering
  zu halten. Bei der Bewertung der Verhältnismäßigkeit der Maßnahmen
  nach Satz 1 sind das Ausmaß der Risikoexposition, die Größe der
  Einrichtung, die Umsetzungskosten und die Eintrittswahrscheinlichkeit
  und Schwere von Sicherheitsvorfällen sowie ihre gesellschaftlichen und
  wirtschaftlichen Auswirkungen zu berücksichtigen. Die Einhaltung der
  Verpflichtung nach Satz 1 ist durch die Einrichtungen zu
  dokumentieren.
\end{enumerate}
\end{quote}

\subsubsection{§ 31 Abs. 2}\label{sec-app-bsig-31}

The provision (p.~23) adds the attack-detection duty for operators of
critical installations:

\begin{quote}
\begin{enumerate}
\def\labelenumi{(\arabic{enumi})}
\setcounter{enumi}{1}
\tightlist
\item
  Betreiber kritischer Anlagen sind verpflichtet, für die
  informationstechnischen Systeme, Komponenten und Prozesse, die für die
  Funktionsfähigkeit der von ihnen betriebenen kritischen Anlagen
  maßgeblich sind, Systeme zur Angriffserkennung einzusetzen. Die
  eingesetzten Systeme zur Angriffserkennung müssen geeignete Parameter
  und Merkmale aus dem laufenden Betrieb kontinuierlich und automatisch
  erfassen und auswerten. Sie sollten dazu in der Lage sein, fortwährend
  Bedrohungen zu identifizieren und zu vermeiden sowie für eingetretene
  Störungen geeignete Beseitigungsmaßnahmen vorzusehen. Dabei soll der
  Stand der Technik eingehalten werden. Der hierfür erforderliche
  Aufwand soll nicht außer Verhältnis zu den Folgen eines Ausfalls oder
  einer Beeinträchtigung der betroffenen kritischen Anlage stehen.
\end{enumerate}
\end{quote}

\subsection{Cyber Resilience Act, Regulation (EU)
2024/2847}\label{sec-app-cra}

\subsubsection{Article 12(1)}\label{sec-app-cra-12}

The paragraph (p.~34) joins the two Union regulations, so that
conformity under the Cyber Resilience Act discharges the cybersecurity
requirement of the AIA:

\begin{quote}
1. Without prejudice to the requirements relating to accuracy and
robustness set out in Article 15 of Regulation (EU) 2024/1689, products
with digital elements which fall within the scope of this Regulation and
which are classified as high-risk AI systems pursuant to Article 6 of
that Regulation shall be deemed to comply with the cybersecurity
requirements set out in Article 15 of that Regulation where: (a) those
products fulfil the essential cybersecurity requirements set out in Part
I of Annex I; (b) the processes put in place by the manufacturer comply
with the essential cybersecurity requirements set out in Part II of
Annex I; and (c) the achievement of the level of cybersecurity
protection required under Article 15 of Regulation (EU) 2024/1689 is
demonstrated in the EU declaration of conformity issued under this
Regulation.
\end{quote}

\subsubsection{Article 3, points (22) and (48), and Recital
18}\label{sec-app-cra-oss}

The obligations of the Regulation attach to making a product available
on the market, and the definition (p.~30) confines that notion to supply
in the course of a commercial activity:

\begin{quote}
\begin{enumerate}
\def\labelenumi{(\arabic{enumi})}
\setcounter{enumi}{21}
\tightlist
\item
  `making available on the market' means the supply of a product with
  digital elements for distribution or use on the Union market in the
  course of a commercial activity, whether in return for payment or free
  of charge;
\end{enumerate}
\end{quote}

The companion definition (p.~31) fixes the notion of free and
open-source software:

\begin{quote}
\begin{enumerate}
\def\labelenumi{(\arabic{enumi})}
\setcounter{enumi}{47}
\tightlist
\item
  `free and open-source software' means software the source code of
  which is openly shared and which is made available under a free and
  open-source licence which provides for all rights to make it freely
  accessible, usable, modifiable and redistributable;
\end{enumerate}
\end{quote}

Recital 18 (pp.~4 and 5) draws the consequence for software that is not
monetised, and states the condition under which a component supplied for
integration nonetheless counts. The two sentences are quoted in the
order in which they appear:

\begin{quote}
In relation to economic operators that fall within the scope of this
Regulation, only free and open-source software made available on the
market, and therefore supplied for distribution or use in the course of
a commercial activity, should fall within the scope of this Regulation.
\end{quote}

\begin{quote}
Furthermore, the supply of products with digital elements qualifying as
free and open-source software components intended for integration by
other manufacturers into their own products with digital elements should
be considered to be making available on the market only if the component
is monetised by its original manufacturer.
\end{quote}

\subsubsection{Article 71(2)}\label{sec-app-cra-71}

The paragraph (p.~67) sets the dates from which the Regulation applies:

\begin{quote}
This Regulation shall apply from 11 December 2027.

However, Article 14 shall apply from 11 September 2026 and Chapter IV
(Articles 35 to 51) shall apply from 11 June 2026.
\end{quote}

\subsection{Product Liability Directive, Directive (EU)
2024/2853}\label{sec-app-pld}

\subsubsection{Article 2(1) and (2)}\label{sec-app-pld-2}

The two paragraphs (p.~11) fix the temporal scope of the Directive and
exclude free and open-source software supplied outside a commercial
activity:

\begin{quote}
1. This Directive shall apply to products placed on the market or put
into service after 9 December 2026.
\end{quote}

\begin{quote}
2. This Directive does not apply to free and open-source software that
is developed or supplied outside the course of a commercial activity.
\end{quote}

\subsubsection{Article 4, point (1)}\label{sec-app-pld-4}

The definition (p.~12) makes software a product in its own right, which
is what brings a forecasting system inside a strict-liability regime:

\begin{quote}
\begin{enumerate}
\def\labelenumi{(\arabic{enumi})}
\tightlist
\item
  `product' means all movables, even if integrated into, or
  inter-connected with, another movable or an immovable; it includes
  electricity, digital manufacturing files, raw materials and software;
\end{enumerate}
\end{quote}

\subsubsection{Article 7(2), points (c) and (f)}\label{sec-app-pld-7}

Article 7(1) defines a defective product as one that does not provide
the safety a person is entitled to expect. Two of the circumstances that
Article 7(2) requires the assessment to take into account (p.~14) reach
a machine-learning system directly:

\begin{quote}
\begin{enumerate}
\def\labelenumi{(\alph{enumi})}
\setcounter{enumi}{2}
\tightlist
\item
  the effect on the product of any ability to continue to learn or
  acquire new features after it is placed on the market or put into
  service;
\end{enumerate}
\end{quote}

\begin{quote}
\begin{enumerate}
\def\labelenumi{(\alph{enumi})}
\setcounter{enumi}{5}
\tightlist
\item
  relevant product safety requirements, including safety-relevant
  cybersecurity requirements;
\end{enumerate}
\end{quote}

\section{The challenge data bundle}\label{sec-app-data}

The results of Section~\ref{sec-results} are computed from a frozen data
bundle that ships with the manuscript source in its \texttt{data/}
directory and that this appendix documents. The bundle is the final
state of the challenge, as scored on 21 July 2026 after the last ENTSO-E
data revisions, and consists of two files: an hourly forecast matrix and
an entry register. Every number, table, and figure of
Section~\ref{sec-results} is recomputed from these two files at render
time, and the public leaderboard they reproduce is available at
\url{https://bartzbeielstein.github.io/challenge-leaderboard/}.

\subsection{\texorpdfstring{The forecast matrix
\texttt{results\_plain.parquet}}{The forecast matrix results\_plain.parquet}}\label{sec-app-data-matrix}

The matrix holds 1344 hourly rows on a UTC axis, from 2026-05-26 00:00
to 2026-07-20 23:00, and 21 columns (Table~\ref{tbl-app-data-matrix}).

\begin{longtable}[]{@{}
  >{\raggedright\arraybackslash}p{(\linewidth - 2\tabcolsep) * \real{0.3333}}
  >{\raggedright\arraybackslash}p{(\linewidth - 2\tabcolsep) * \real{0.6667}}@{}}
\caption{Columns of the forecast matrix
\protect\texttt{results\_plain.parquet}.}\label{tbl-app-data-matrix}\tabularnewline
\toprule\noalign{}
\begin{minipage}[b]{\linewidth}\raggedright
Column
\end{minipage} & \begin{minipage}[b]{\linewidth}\raggedright
Meaning
\end{minipage} \\
\midrule\noalign{}
\endfirsthead
\toprule\noalign{}
\begin{minipage}[b]{\linewidth}\raggedright
Column
\end{minipage} & \begin{minipage}[b]{\linewidth}\raggedright
Meaning
\end{minipage} \\
\midrule\noalign{}
\endhead
\bottomrule\noalign{}
\endlastfoot
index \texttt{timestamp\_utc} & hour beginning, UTC, no gaps \\
\texttt{actual\_load} & realised load in MW, the ground truth \\
\texttt{entsoe} & official ENTSO-E day-ahead forecast in MW, the ENTSO-E
baseline \\
19 further columns & one per scored identity, in MW \\
\end{longtable}

Forecast cells hold each submission exactly as the leaderboard scored
it. A day that a participant missed was scored by carrying their last
submission forward, and the carried value is what appears here. Hours
outside an identity's scored days are missing.

The 41 target days of the live phase run from 10 June to 20 July 2026.
The fifteen preceding days carry the actual load and the ENTSO-E
baseline but no forecasts. They are present because the MASE scaling
factor of Section~\ref{sec-metrics} and the 168-hour seasonal-naive
benchmark both reach back before the first target day. Coverage per
identity is derivable from the matrix and is deliberately not duplicated
in the entry register.

\subsection{\texorpdfstring{The entry register
\texttt{entries.csv}}{The entry register entries.csv}}\label{sec-app-data-entries}

The register holds one row per scored identity, 22 rows in total
(Table~\ref{tbl-app-data-entries}).

\begin{longtable}[]{@{}
  >{\raggedright\arraybackslash}p{(\linewidth - 2\tabcolsep) * \real{0.3333}}
  >{\raggedright\arraybackslash}p{(\linewidth - 2\tabcolsep) * \real{0.6667}}@{}}
\caption{Columns of the entry register
\protect\texttt{entries.csv}.}\label{tbl-app-data-entries}\tabularnewline
\toprule\noalign{}
\begin{minipage}[b]{\linewidth}\raggedright
Column
\end{minipage} & \begin{minipage}[b]{\linewidth}\raggedright
Meaning
\end{minipage} \\
\midrule\noalign{}
\endfirsthead
\toprule\noalign{}
\begin{minipage}[b]{\linewidth}\raggedright
Column
\end{minipage} & \begin{minipage}[b]{\linewidth}\raggedright
Meaning
\end{minipage} \\
\midrule\noalign{}
\endhead
\bottomrule\noalign{}
\endlastfoot
\texttt{team\_id} & matches a column of the matrix, or names a derived
benchmark \\
\texttt{display\_name} & name as printed in this paper \\
\texttt{kind} & \texttt{student}, \texttt{reference}, \texttt{baseline},
or \texttt{benchmark} \\
\texttt{group} & cohort of a student team, \texttt{ING} or \texttt{AIT},
otherwise empty \\
\texttt{joker} & date of the one-time joker substitution, if the team
used it \\
\texttt{n\_locf} & number of carried-forward days among the scored
days \\
\end{longtable}

The \texttt{benchmark} rows are the two seasonal-naive comparators. They
have no column in the matrix because they are computed from the actual
load at render time.

\subsection{Rebuilding the bundle}\label{sec-app-data-rebuild}

The bundle is rebuilt from the frozen leaderboard repository by

\begin{verbatim}
cd bart26o && uv run python data/make_results_plain.py
\end{verbatim}

The builder writes nothing unless five checks pass: (i) the hourly
coverage matches the scored-day counts, (ii) the rows preceding the live
phase carry no forecasts, (iii) every scored team-day reproduces its
frozen per-day value on all of MAE, RMSE, MAPE, bias, and UPR, (iv)
every column's mean MAE reproduces the published board scores in
\texttt{site\_scores.json}, and (v) the entry metadata covers exactly
the matrix.

The file \texttt{site\_scores.json} is retained beside the bundle purely
as that verification gate. It is a copy of the published board and is
not read by the manuscript for any value it reports.

\subsection{Provenance and licence}\label{sec-app-data-licence}

The series \texttt{actual\_load} and \texttt{entsoe} originate from the
ENTSO-E Transparency Platform (ENTSO-E 2024) and are subject to its
terms of use. The forecast columns are the participants' own
submissions, contributed for this challenge.

\end{document}